\documentclass[twoside,english]{elsarticle}
\usepackage[latin9]{inputenc}
\usepackage{amsthm}
\usepackage{hyperref}

\makeatletter
\theoremstyle{plain}

\journal{Elsevier}

\usepackage{lineno}

\usepackage{amssymb}
\usepackage{amsmath}
\usepackage{har2nat}
\usepackage[margin=3cm]{geometry}
\usepackage{float}
\usepackage{algpseudocode}
\usepackage{graphicx}
\usepackage{xcolor}

\biboptions{authoryear}

\makeatother

\usepackage{babel}
\providecommand{\theoremname}{Theorem}

\begin{document}

\begin{frontmatter}{}

\title{A Multilayered Block Network Model to Forecast Large Dynamic Transportation Graphs: an Application to US Air Transport\tnoteref{t1,t2}}

\author[liu]{Hector~Rodriguez-Deniz\corref{cor1}}

\ead{hector.rodriguez@liu.se}

\author[liu,su]{Mattias~Villani}

\ead{mattias.villani@stat.su.se}

\author[edi]{Augusto~Voltes-Dorta}

\ead{augusto.voltes-dorta@ed-ac.uk}

\cortext[cor1]{Corresponding author}

\address[liu]{Division of Statistics and Machine Learning, Link\"oping University,
Sweden}

\address[su]{Department of Statistics, Stockholm University, Sweden}

\address[edi]{Management Science and Business Economics Group, University of Edinburgh, United Kingdom}
\begin{abstract}
Dynamic transportation networks have been analyzed for years by means of static graph-based indicators in order to study the temporal evolution of relevant network components, and to reveal complex dependencies that would not be easily detected by a direct inspection of the data. This paper presents a state-of-the-art probabilistic latent network model to forecast multilayer dynamic graphs that are increasingly common in transportation and proposes a community-based extension to reduce the computational burden. Flexible time series analysis is obtained by modeling the probability of edges between vertices through latent Gaussian processes. The models and Bayesian inference are illustrated on a sample of 10-year data from four major airlines within the US air transportation system. Results show how the estimated latent parameters from the models are related to the airline's connectivity dynamics, and their ability to project the multilayer graph into the future for out-of-sample full network forecasts, while stochastic blockmodeling allows for the identification of relevant communities. Reliable network predictions would allow policy-makers to better understand the dynamics of the transport system, and help in their planning on e.g. route development, or the deployment of new regulations.

\end{abstract}

\end{frontmatter}{}


\section{Introduction and related work}
\label{sec:intro}

We live in a highly interconnected world, and networks have become
an integral part of our life, from telecommunications and social media
to transportation systems and the Internet of Things. Further technological
advances and the advent of automation, which may enable the autonomous
operation of actors within the network, are likely to push the scale
and sophistication of network systems up to new levels in the near
future. This increasing complexity has permeated into science in a
natural way, and the use of network modeling has become widespread
in disciplines as diverse as Sociology, Neuroscience or Transportation
\citep{jasny2009complex,Barabasi2016networks}, propelled by the availability
of data and computing power. Transportation science has been for many
decades an active field aiming for the development of models and policies
that ensure the efficiency, safety and social acceptability of transportation
systems, while limiting costs and environmental impact. In recent years there has been a growing number of research
directions that reveal the need for appropriate methods to address
the complexity imposed by network problems. Network resilience analysis
against e.g. natural disasters or terrorist attacks, structural evolution
of network systems, and network-wide traffic forecasting, are examples
of these new directions. The interest in modeling and understanding
transportation networks is not merely academic. The maritime shipping
network processes over 80\% of the world trade, whereas travel and
tourism industries, which sustain 10\% of the global GDP, rely on
the air transport network (\citealp{UNCTAD2018,WTTC2018}). Also,
public transport networks are a key element in the infrastructure
of large urban areas, where the bulk of the economic activity is concentrated
in most developed countries. 

Here we focus on multilayer dynamic networks and use latent variable models for analyzing complex graph data that are increasingly available in transportation and related fields. Multilayer networks add a new dimension to the network representation through the definition of layers, and allow for the modeling of complex systems that would be difficult to represent using regular, "flat" graphs. Multilayer networks can be defined and structured in various ways depending on the problem and the specific role of the layers \citep{kivela2014multilayer}. In transportation, a multilayer graph could represent how different airlines interact with the underlying airport network, or how different transport modes (e.g. bus and metro networks) operate simultaneously within a public transportation system. Dynamic networks, on the other hand, add a temporal dimension to the problem by assuming that the interaction between the elements in the network change over time, which is common in transport systems, as discussed below. A latent network model \citep{kolaczyk2017topics,crane2018probabilistic} is a probabilistic model that uses unobserved features to characterize different properties and processes within a graph. These models are flexible and allow for e.g. link prediction and community detection, and are also relatively easy to estimate, see Section \ref{sec:statnetworks} for further details and relevant literature.

Dynamic transportation networks have been studied for years, both
from short-term and long-term perspectives. Studies featuring a long-term
approach usually aim to analyze the structural dynamics of the transportation
system in order to assess the temporal evolution of relevant network
components in terms of months or years, and to reveal complex dependencies
and patterns that would not be easily detected by a direct inspection
of the data. A graph-based analysis through the use of measures such
as e.g. the node degree or betweenness (\citealp{guimera2005worldwide})
became the \textit{de facto} methodological approach, which has been
also used to study the dynamics of shipping and airport networks \citep{ducruet2012worldwide,wang2014evolution},
or airline de-hubbing \citep{rodriguez2013classifying}, among many
others. On the other hand, short-term network dynamic problems usually
deal with time spans of minutes or hours, and focus on modeling specific
elements within the network (e.g. link congestion) rather than adopting
a structural approach. A good example of a short-term network problem
in transportation is urban traffic forecasting. This is a time-series
prediction problem that has been traditionally addressed with statistical
and machine learning models \citep{vlahogianni2014short}, and where
a explicit graph-based representation of the network is not strictly
necessary. We believe there is an opportunity for a methodological
advance in studies involving the long-term analysis of transportation
networks by using state-of-the-art statistical models for dynamic
and multilayer graph data, therefore moving from the current descriptive,
indicator-based approach to an inferential one. Moreover, reliable graph 
forecasting would allow for the definition of benchmarking scenarios in 
problems where origin-destination pairs are usually assumed
 fixed. Examples of those applications would be airline schedule optimization 
\citep{cadarso2017trc}, cargo assignment in shipping networks \citep{wang2016trc} 
and hub-location \citep{alkaabneh2019trc}, to name a few. 

In spite of the above, there have been some recent contributions to model-based inference in transportation graphs. 
A representative example is the paper from \citet{kotegawa2010trc}, which addresses the problem of 
route (link) prediction in air transportation networks. Their work is motivated by the air traffic forecasts from the
US Federal Aviation Administration (FAA), which does not consider network dynamics in their predictions.
The authors tackle the graph forecasting problem by training a model that uses topological characteristics 
of the airports as covariates, and yield the probability of city-pairs being connected by new routes in the future. Three
competing models (logistics regression, neural networks, and a preferential attachment algorithm) were tested, with the 
artificial neural network being the top performer. The authors stress that given the competitive nature of the airline industry
 and its implications on the network's structure (e.g. dehubbing),  reliable network forecasts would allow policy-makers to 
better understand the dynamics of the system, and help in their planning on e.g. infrastructure development, or the deployment of new regulations.
Similarly, \citet{de2016route} perform an econometric analysis to study the probability of market closures for European low-cost
carriers. They implement a logistic regression model with route-based covariates (e.g. distance, offered seats) that could help identify
the factors that make markets more likely to be canceled, although their approach lack the network and dynamic perspective.

Working with large graphs can be computationally tough, and transportation networks are no exception to that. A practical way 
to circumvent this issue that has been recently adopted in the literature is to reduce the scale of the
 problem by partitioning the graph. In graph clustering, also known as community detection (see e.g. \citealp{fortunato2010community}) 
or stochastic blockmodeling (\citealp{nowicki2001estimation, airoldi2008mixed}), the objective is to find groups of highly interconnected
 elements within the network, which may reveal structures such as e.g. social cliques or spatial patterns. In a public transport context, \citet{yap2019trc} used a two-pronged strategy to reduce the size
of their transfer synchronization problem. First they performed spatial clustering to isolate relevant transfer locations, and then used graph-based community detection to determine which line bundles within the selected hubs to synchronize. \citet{tian2020trc} greatly reduced
the computational cost in a large-scale rebalancing problem by clustering their bike-sharing network into five management areas. 
Still, graph clustering has not been exclusively used to deal with the computational bottleneck.
\citet{yildirimoglu2018trc} use modularity-based community detection to find demand patterns in an urban multilayer
 setting (bus, passenger and car networks), thus allowing for a demand analysis at different spatial resolutions that may 
be helpful for planning and operation, whereas \citet{olmos2020trc} identified relevant demand clusters for a better
design of a network of bicycle paths. One aspect that is common among the previous studies is that the clustering is framed as a separate spatial problem, without explicitly considering the dynamics of the network. 
	
The aim of this paper is threefold: i) present a state-of-the-art latent network model to forecast multilayer dynamic graphs that are increasingly common in transportation, which  have potential applications to the long-term analysis of transportation networks, ii) propose a community-based extension of the model to reduce the computational burden by jointly considering the temporal and spatial dimensions of the network, and iii) demonstrate their applicability to a real multilayer transportation network in two case studies with US airline data. 

The rest of the manuscript is organized as follows. Section \ref{sec:statnetworks} briefly introduce
relevant work on statistical models for network data. The methodological
framework and the community-based model are detailed in Section 3, whereas in Section 
4 a set of validation experiments are performed. Section 5 presents applications to real 
data from an airline network. The last section 
summarizes the paper and discuss limitations and possible directions for further research.

\section{Statistical models for graph data}\label{sec:statnetworks}

Statistical network analysis is a well-established field of research
(see e.g. \citealp{Kolaczyk:2009:SAN:1593430})
with origins dating back to the seminal work on random graphs by \citet{Erdos1959}.
Despite their fundamental contributions, the original mathematical
models, along with other recent models such as the ``small-worlds''
from \citet{watts1998collective} and the hub-and-spoke networks from
\citet{barabasi1999emergence}, are too limited for most applications.
Exponential Random Graph Models (ERGMs) were designed with this aim
in mind, initially with the $p_1$\textit{ }model from \citet{Holland1981},
and define an exponential family of distributions over a graph. However, model degeneracy
and intractability are still unresolved problems, which has represented a hurdle for a wider applicability outside
social networks. Interestingly, \citet{zhang2019trc} recently used ERGM models to learn
social networks effects that can be used to generate synthetic populations in 
agent-based transport simulators.

In contrast with the log-linear approach of the ERGMs, latent
network models (LNMs) define latent classes or features to 
capture the network complexity in a non-linear fashion.
The Stochastic Block Model (SBM - \citealp{holland1983stochastic})
is perhaps the most popular latent network model, and assumes a latent
community structure that drives the relationship patterns between
actors in the network. \citet{nowicki2001estimation} proposed a Bayesian
inference algorithm using Gibbs sampling whereas \citet{daudin2008mixture}
developed variational inference for the model. Current research on
SBM's is mainly aimed at mixed membership clustering (\citealp{airoldi2008mixed}),
extensions for weighted graphs (\citealp{aicher2014learning}), dynamic
and state-space modeling (\citealp{ishiguro2010dynamic,xu2014dynamic}),
and multi-layer networks (\citealp{han2015consistent,stanley2016clustering}). A
different approach to LNMs is to define
a latent space over the network nodes themselves (\citealp{hoff2002latent}).
In this case, the probability that two network elements are connected
can be defined in terms of a distance function, in such a way that
nodes neighboring in the unobserved latent space are more likely to
be connected. Latent space models are able to capture transitive dependencies
in a natural way (in contrast with SBM's) and are flexible enough
to incorporate dynamics while allowing for practical maximum likelihood
and Bayesian inference. More recently, \citet{durante2014nonparametric} introduce
exact Bayesian inference using P{\'o}lya-Gamma augmentation (\citealp{polson2013bayesian})
for a dynamic latent space network model driven by Gaussian Processes.
A natural extension of the previous model to a dynamic multilayer
setting is presented in \citet{durante2017bayesian}, although scalability
issues arise for large network data since the model introduces a Gaussian
process for each node in each layer.

\section{Methodology}
\label{sec:meth}
\subsection{Dynamic Multilayered Network Model}

We first describe the dynamic multilayered network model in \citet{durante2014nonparametric}
and \citet{durante2017bayesian} that serves as a starting point for
our community-based extension. We represent a network as a graph $G=(V,E)$ where
$V$ is the set of vertices (also called nodes) $i=1,\ldots,N$ and
$E$ a set of unweighted edges (also called links) between node pairs
$\{i,j\}$. The connectivity of the graph is summarized in the $N \times N$
adjacency matrix $A_{ij}$, where $A_{ij}=1$ if there is an edge
connecting vertices $i$ and $j$, and $A_{ij}=0$ otherwise. We assume
undirected edges and no self-loops, i.e. $A_{ij}=A_{ji}$ and $A_{ii}=0$.
Dynamic multilayer graphs have a graph per layer that evolves in time
and can be represented by adjacency matrices $A_{ij}^k(t)$ where
$A_{ij}^k(t)=1$ if vertices $i$ and $j$ are connected in layer
$k=1,\ldots,K$ at time $t=t_1,\ldots,t_T$.

The dynamic multilayer network model in \citet{durante2014nonparametric}
and \citet{durante2017bayesian} defines a Bayesian logistic regression for each element in the adjacency matrix $A_{ij}^{k}(t)$, i.e. for each possible edge in the graph at any layer $k$ and time $t$. The logit of the probability that any two vertices in the multilayer graph are connected depends on a model with three additive components featuring unobserved variables that encode connectivity patterns. Specifically, the model is of the form
\begin{align}
A_{ij}^{k}(t) & \sim\mathrm{Bernoulli}\left(\pi_{ij}^{k}(t)\right)\nonumber \\
\psi_{ij}^{k}(t)=\mathrm{Logit}\left(\pi_{ij}^{k}(t)\right) & =\mu(t)+\sum_{r=1}^{R}\bar{x}_{ir}(t)\bar{x}_{jr}(t)+\sum_{h=1}^{H}x_{ih}^{k}(t)x_{jh}^{k}(t),\label{eq:DuranteModel}
\end{align}
where the latent processes, $\mu(t)$, $\bar{x}_{ir}(t)$ and $x_{ih}^{k}(t)$,
are assumed to be smoothly evolving Gaussian processes with RBF kernel
functions (\citealp{rasmussen2006gaussian})\begin{subequations}\begin{align}
	\mu(t) & \sim \mathcal{GP}(0,k_{\mu})\\
	\bar{x}_{ir}(t) & \sim \mathcal{GP}(0,\tau_r^{-1}k_{\bar{x}})\\
	x^k_{ih}(t) & \sim \mathcal{GP}(0,\tau_h^{k^{-1}}k_{x}).
	\end{align}\end{subequations}The model is structured through a set of latent variables that capture
different effects within the network. The global time-varying intercept
$\mu(t)$ defines a baseline network density for all nodes in all
layers. The cross-layer effects $\bar{x}_i(t)$ enter as a bilinear
form \citep{hoff2005bilinear}, increasing the probability of a link
between vertices as their latent coordinates become aligned whereas
the within-layer $x^k_i(t)$ coordinates act in an identical manner
capturing those effects not shared across the different layers. Instead
of learning the dimensionality $R$ and $H$ of the latent coordinates
$\bar{x}_{ir}(t),r=1,\ldots,R$ and $x^k_{ih}(t),h=1,\ldots,H$, the
model uses  multiplicative inverse Gamma priors \citep{bhattacharya2011sparse}
to induce a shrinkage effect that becomes stronger for larger $r$
and $h$

\begin{equation}
\tau_r^{-1} = \prod_{u=1}^r\delta_u^{-1},\quad r=1,\ldots,R
\end{equation}

\begin{equation}
\delta_1 \sim \mathrm{Gamma}(a_1,1),\ \delta_{u>1} \sim \mathrm{Gamma}(a_2,1)
\end{equation}

\begin{equation}
\left(\tau_h^k\right)^{-1} = \prod_{v=1}^h\left(\delta_v^k\right)^{-1},\quad h=1,\ldots,H,\quad k=1,\ldots,K
\end{equation}

\begin{equation}
\delta_1^k \sim \mathrm{Gamma}(a_1,1),\ \delta_{v>1}^k \sim \mathrm{Gamma}(a_2,1).
\end{equation}

\citet{durante2017bayesian} prove that the model in Eq.\eqref{eq:DuranteModel}
is very flexible and can essentially model any matrix of edge probabilities
if $R$ and $H$ is large enough. The likelihood factorizes into a
set of Bernoulli logistic regressions which can be Gibbs sampled using
the P{\'o}lya-Gamma data augmentation in \citet{polson2013bayesian}. 
However, the number of Gaussian processes that needs to be learned
from data is $1+RN+HKN$, which makes computations and storage unmanageable
for anything except small networks with few layers and small number
of nodes. Moreover, the model is completely unstructured and is therefore
massively overparametrized when the data follow some structure, e.g.
some sort of community clustering. In the next section we propose
a SBM extension of the model with a dramatically reduced number of
Gaussian processes and develop a Gibbs sampling algorithm for inference 
using the P{\'o}lya-Gamma data augmentation
trick. The model imposes a community structure and is
therefore less general than \citet{durante2017bayesian}, but benefits
from a reduction in the number of Gaussian processes and scales much
better to larger networks.

\subsection{Dynamic Multilayered Block Network Model}

To impose a community structure we assume that each vertex in the
network belongs to a stochastic block (\citealp{nowicki2001estimation})
or cluster $b\in\{1,\ldots,B\}$ with prior probability $p(z_i=b)=\eta_b$,
where $z$ is the vector of block assignments, indicating to which
block each vertex $i$ belongs, and $\eta\sim \mathrm{Dirichlet}(\alpha_1,\ldots,\alpha_B)$. This results in logistic regressions that model the interactions between groups of vertices, i.e. the probability of the existence of an edge between any two vertices in the network now depends on which block/cluster they belong.  We propose the following block model extension of \citet{durante2017bayesian} 

\begin{align}
z_{i} & \sim\mathrm{Categorical}(\eta_{1},...,\eta_{B})\nonumber \\
A_{ij}^{k}(t)&\vert\left(z_{i}=p,z_{j}=q\right)  \sim\mathrm{Bernoulli}\left(\pi_{pq}^{k}(t)\right)\\
\psi_{pq}^{k}(t)=\mathrm{Logit}\left(\pi_{pq}^{k}(t)\right) & =\begin{cases}
\mu(t)+\sum_{r=1}^{R}\bar{x}_{pr}(t)\bar{x}_{qr}(t)+\sum_{h=1}^{H}x_{ph}^{k}(t)x_{qh}^{k}(t) & \text{if }p\neq q\\
\mu_{p}^{k}(t)+\sum_{r=1}^{R}\bar{x}_{pr}(t) & \text{if }p=q
\end{cases}\label{eq:OurModel}
\end{align}

\begin{figure}[h] 
	\centering 
	\includegraphics[trim={0 -20 0 -20},width=1\textwidth]{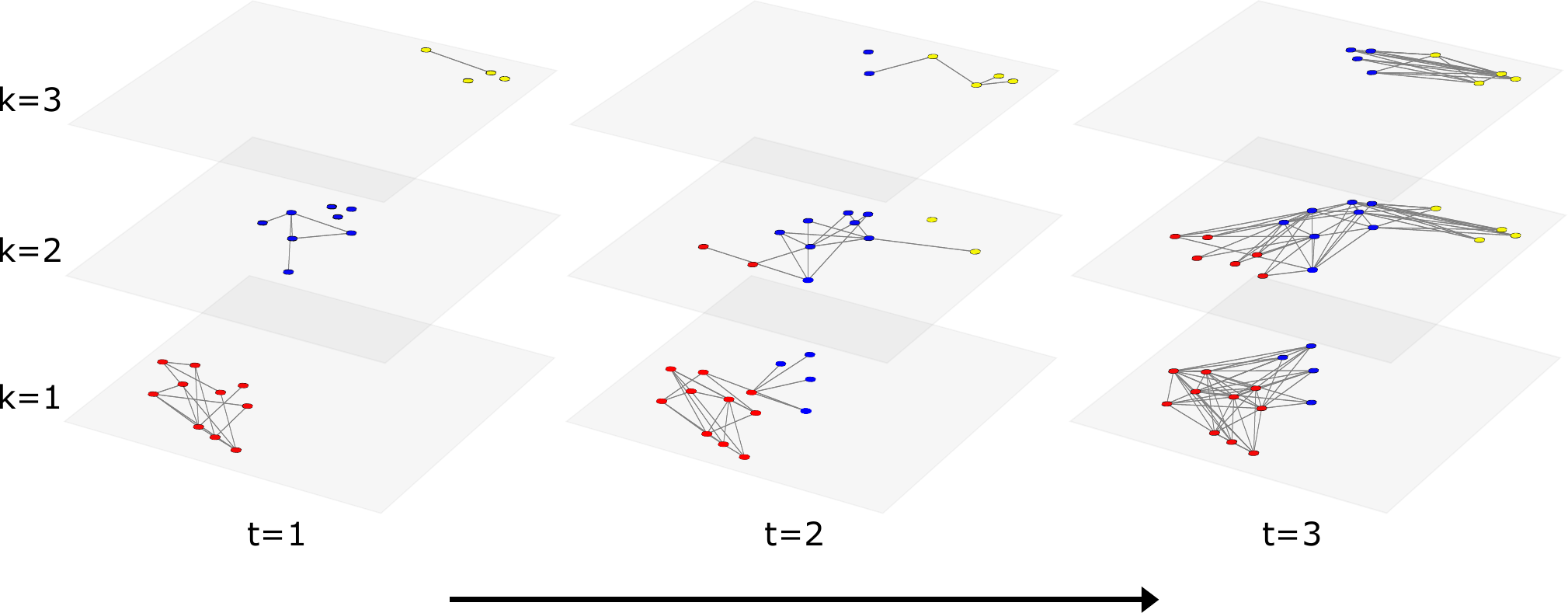} 
	\caption{Example of a dynamic multilayer network with three stochastic blocks (red, blue and yellow), layers and time points.} 
	\label{fig:multilayer}
\end{figure}

The link probabilities $\pi_{pq}^{k}(t)$ for $p\neq q$ are of the
same form as in \citet{durante2017bayesian}, but here defined over
blocks, for a given block assignment. The within-block link probabilities for
$p=q$ are modeled separately with a dynamic intercept per block and
layer $\mu_p^k(t)$, and a second term with the sum of cross-layer
coordinates of the corresponding block that allows for some block-wise
leveraging between the two logits. See Figure \ref{fig:multilayer}
for a graphical representation.

\subsection{A Scalable Gibbs Sampler for Bayesian Inference}

The complete-data likelihood for the proposed model is
\begin{align*}
p(A_{ij}^{k}(t)\lvert\psi_{pq}^{k}(t)) & =\prod_{t=1}^{T}\prod_{k=1}^{K}\prod_{i=2}^{N}\prod_{j=1}^{i-1}\frac{\exp(\psi_{z_{i}z_{j}}^{k}(t))^{A_{ij}^{k}(t)}}{1+\exp(\psi_{z_{i}z_{j}}^{k}(t))}=\prod_{t=1}^{T}\prod_{k=1}^{K}\prod_{p=1}^{B}\prod_{q=1}^{p}\frac{\exp(\psi_{pq}^{k}(t))^{y_{pq}^{k}(t)}}{\left[1+\exp(\psi_{pq}^{k}(t))\right]^{n_{pq}^{k}(t)}},
\end{align*}
where $\psi_{pq}^k(t)=\mathrm{Logit}(\pi_{pq}^k(t))$, and $n_{pq}^k(t)$
and $y_{pq}^k(t)$ are the number of possible and actual edges in
$A_{ij}^k(t)$ between blocks $p$ and $q$, respectively. The block
model induces a set of within-block summations over edges that structures
the likelihood into $TK$ explicit Binomial components instead of
the Bernoulli components in \citet{durante2017bayesian}. This likelihood
allows for exact Bayesian inference using Gibbs sampling with the
P{\'o}lya-Gamma data augmentation for Binomial logistic regression in
\citet{polson2013bayesian}, while automatically reducing the model
size for large network problems as the estimation will be over $B(B+1)/2$
blocks instead of $N(N-1)/2$ vertices, where $B\ll N$.

In \ref{sec:pgtrick} we briefly review the P{\'o}lya-Gamma data augmentation
method (\citealp{polson2013bayesian}), which provides a tractable,
efficient way to perform Bayesian inference on models with binomial
likelihoods, whereas \ref{sec:gibbssampler} gives a detailed description of
a Gibbs sampler algorithm to sample from the joint posterior of all
model parameters. The sampler combines the multilayer network model
from \citet{durante2017bayesian}, modified to our specific structure
of the block link probabilities in Eq.\eqref{eq:OurModel}, with updating
steps for the latent block allocations $\mathbf{z}$ and block probabilities
$\boldsymbol{\eta}$ following \citet{nowicki2001estimation}. 

\begin{figure}[h]
	\centering  
	\begin{tabular}{cc} 
		\includegraphics[trim={0 25 20 50},width=0.5\textwidth]{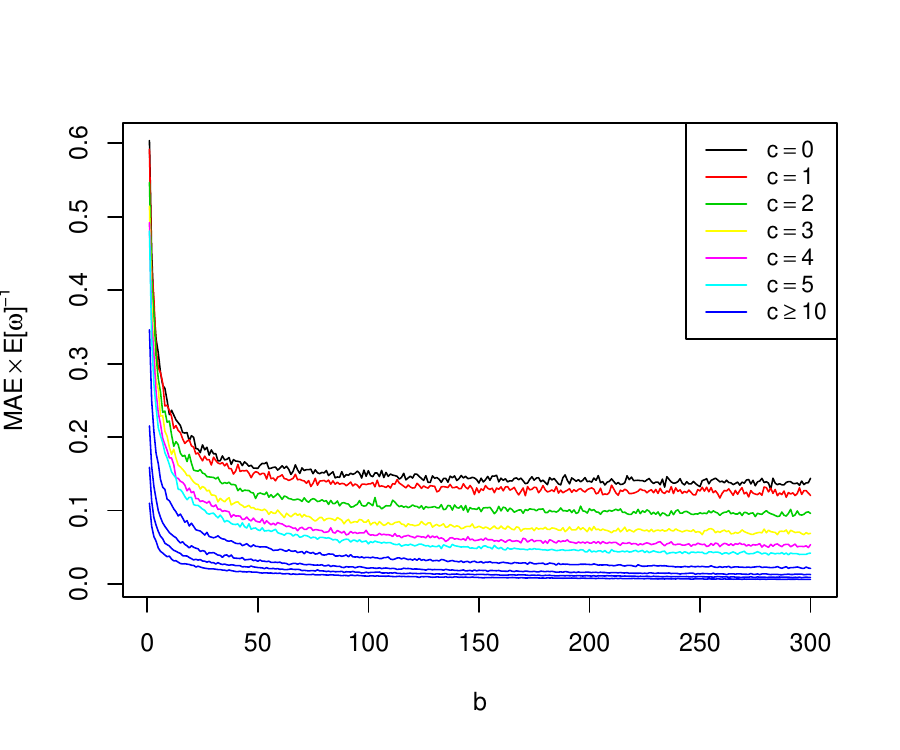} & 
		\includegraphics[trim={0 25 20 50},width=0.5\textwidth]{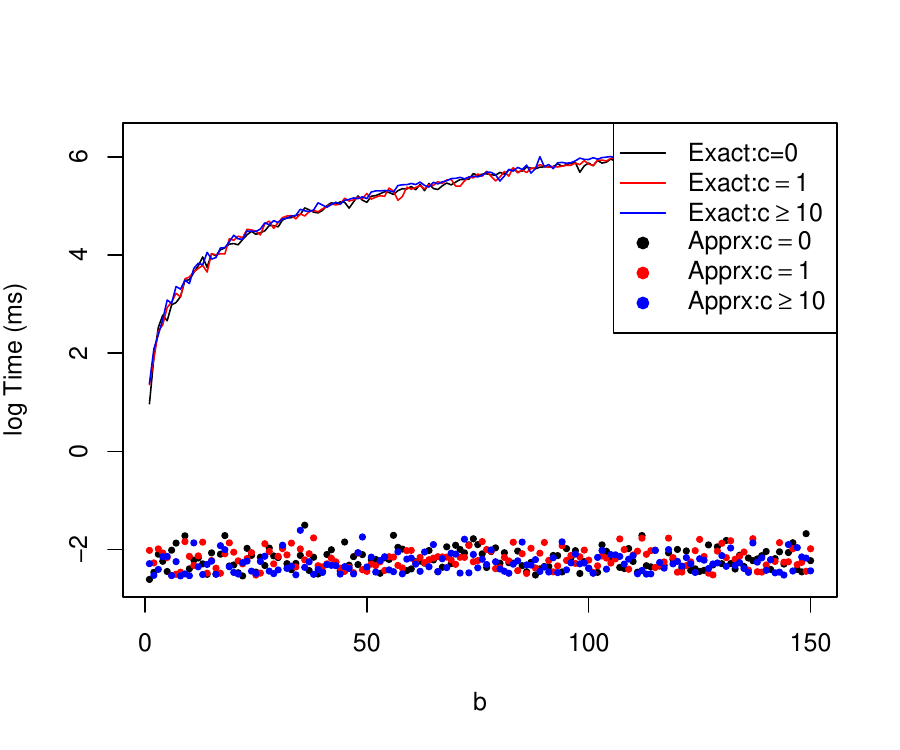} 
	\end{tabular} 
	\caption{Simulation Errors and Times for the P{\'o}lya-Gamma Approximation} 
	\label{fig:pgsimulation}
\end{figure}

The Gibbs sampler in \citet{durante2017bayesian} involves $TKN(N-1)/2$
draws from the $\mathrm{PG(1,c)}$ distribution. \ref{sec:gibbssampler} shows
that this step in our algorithm includes $TKB(B+1)/2$ updating step
for the P{\'o}lya-Gamma variables $\omega_{pq}^{k}(t)\sim\mathrm{PG}(b,c)$,
where $b=n_{pq}^{k}(t)$. Hence, although the number of draws is dramatically
smaller for our algorithm, each draw tends to be more costly since
the time to simulate from $\mathrm{PG}(b,c)$ increase in $b$. To
speed up computations we follow up on the suggestion mentioned in
\citet{windle2014sampling} and develop a fast normal approximation
via moment-matching; see \ref{sec:pgapprox}. Figure \ref{fig:pgsimulation} (left) show the mean
absolute error between the normal approximation and the sampling methods
from \citep{devroye2009exact,polson2013bayesian}, relative to the
theoretical mean. We see that for values of $b\ge 50$  the deviation
from the theoretical mean is less than 20\% in the worst case where
$c<10$ and negligible for $c\ge 10$. Simulation times are independent
from $b$ when sampling from the approximation (right). Therefore,
we use our normal approximation to sample P{\'o}lya-Gamma variables for
$b\ge 100$ , and the standard exact methods otherwise.

The update of the block assignments in the last step of the Gibbs
sampler implies the computation of the posterior of the latent assignment
probabilities for each node in the network (see step 10 in \ref{sec:gibbssampler}). A na\"\i ve implementation will result in $\mathcal{O}(NKTB^2)$
time, which could be prohibitive for large $N$, specially if we update
all $z_i$ sequentially at each MCMC iteration. We recommend an annealed
random-scan sampling that starts updating the entire network and exponentially
decreases the number of vertices being updated to a small fraction
as the estimation progress. This would allow the Gibbs sampler to
initially explore a large space of possible clusterings at a higher
computational cost, and then concentrate the estimation effort on
the parameters defining the block dynamics while still allowing
for some refinement in the assignments.

\section{Simulation study}
\label{sec:sim}

In this section we use synthetic data to compare the fitting capability
and estimation time of the DMN model in \citet{durante2017bayesian}
with our DMBN model with block structure. We simulate multilayer networks
with sizes ranging from $N=\{32,64,128,256,512\}$ and different levels
of granularity: $B=\{5,15,45\}$. We also simulate networks from the
DMN model without block structure. All networks have the same number
of $K=4$ layers and $T=12$ time points, and are generated from a
dynamic six-dimensional latent space, i.e. $R=H=6$, with common smoothness
$l_{\mu}=l_{\mu_p}=l_{\bar{x}}=l_{x}=0.05$ over all components. Each
latent coordinate are simulated from three predefined types of connectivity
patterns: i) smoothed constant, ii) smoothed seasonal connectivity
and iii) smoothed linear trend. Table \ref{tab:sim_param_config} shows the parameter configuration
that is used by both models during the simulations, which were performed
on a cluster from the Swedish National Supercomputer Center (NSC-SNIC).

\begin{table}
	\centering 
	\begin{tabular}{lcccccccccc}
		\hline 
		Model & $B$ & $R$ & $H$ & $l_{\mu}$ & $l_{\mu_p}$ & $l_{\bar{x}}$ & $l_{x}$ & $a_1$ & $a_2$ & MCMC Samples\tabularnewline
		\hline 
		DMN & - & 6 & 6 & 0.05 & - & 0.05 & 0.05 & 2 & 2 & 5,000 (20\% burnin)\tabularnewline
		\hline 
		DMBN & 10 & 6 & 6 & 0.05 & 0.05 & 0.05 & 0.05 & 2 & 2 & 5,000 (20\% burnin)\tabularnewline
		\hline 
	\end{tabular}	
	\caption{Parameter configuration of the two models for the experiment}
	\label{tab:sim_param_config}
\end{table}

\begin{figure}[ht]
	\centering  
	\begin{tabular}{p{0.1cm}ccccc}
		&(a) $B=5$&(b) $B=15$&(c) $B=45$ &(d) No blocks\vspace{0.2cm}
		\\  \rotatebox[origin=l]{90}{\hspace{1.3cm}True}&
		\includegraphics[trim={60 65 60 65},width=.22\linewidth]{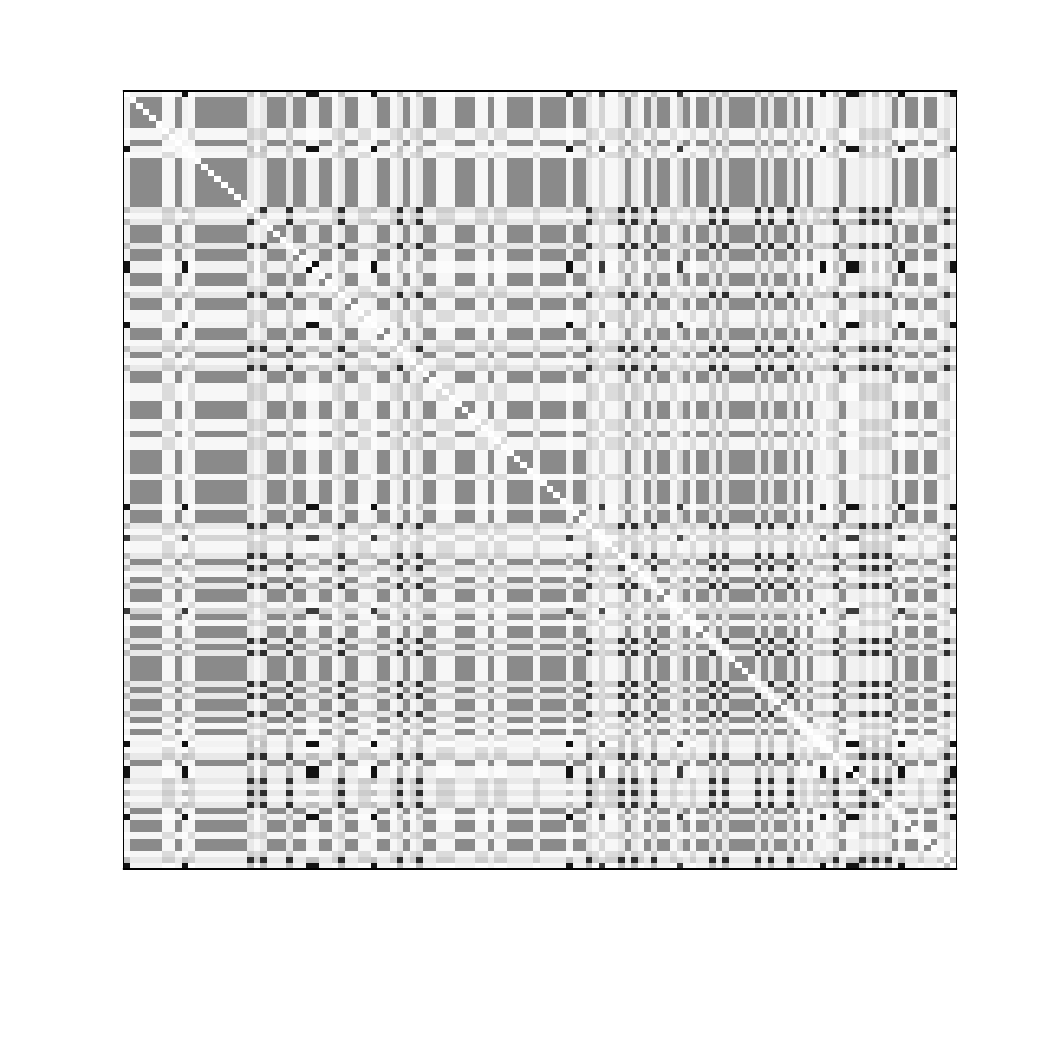} &   \includegraphics[trim={60 65 60 65},width=.22\linewidth]{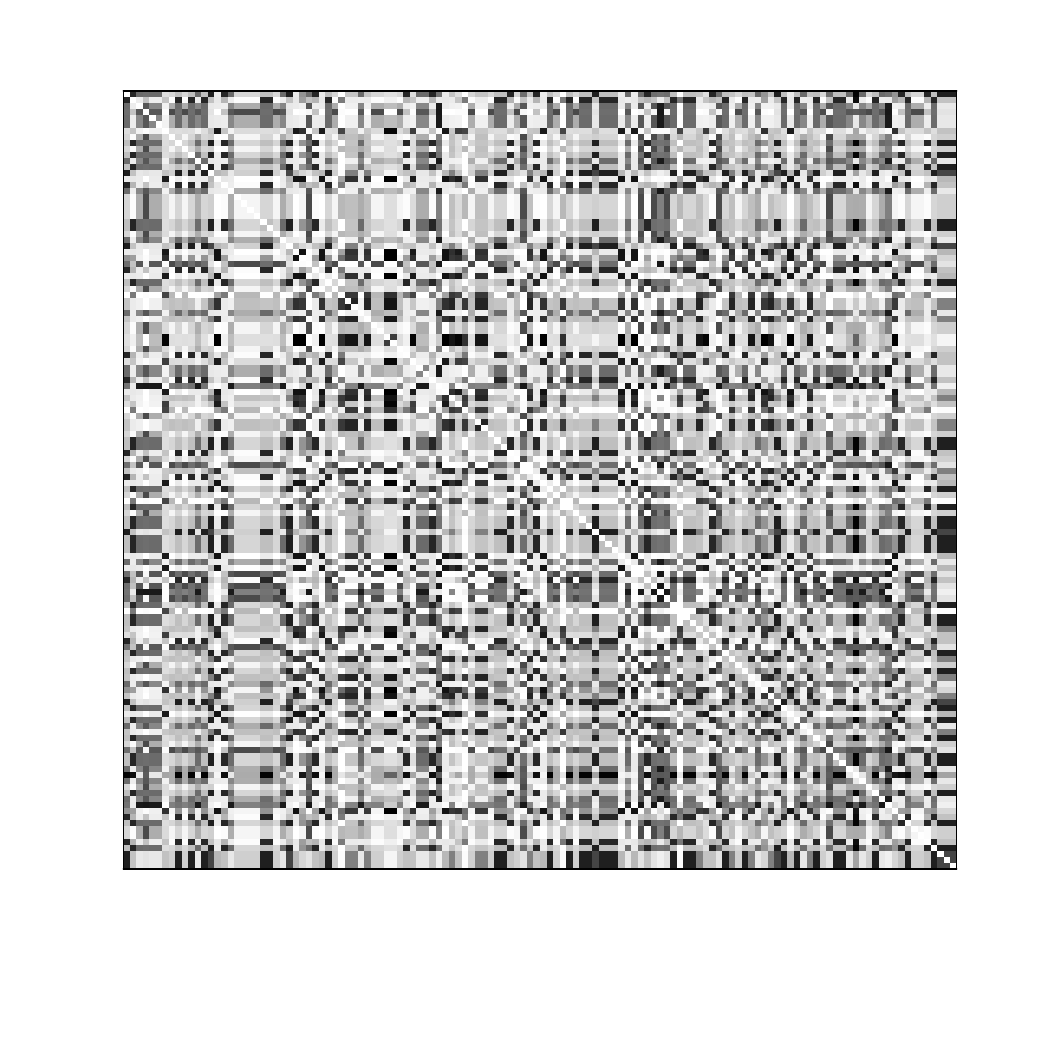} & \includegraphics[trim={60 65 60 65},width=.22\linewidth]{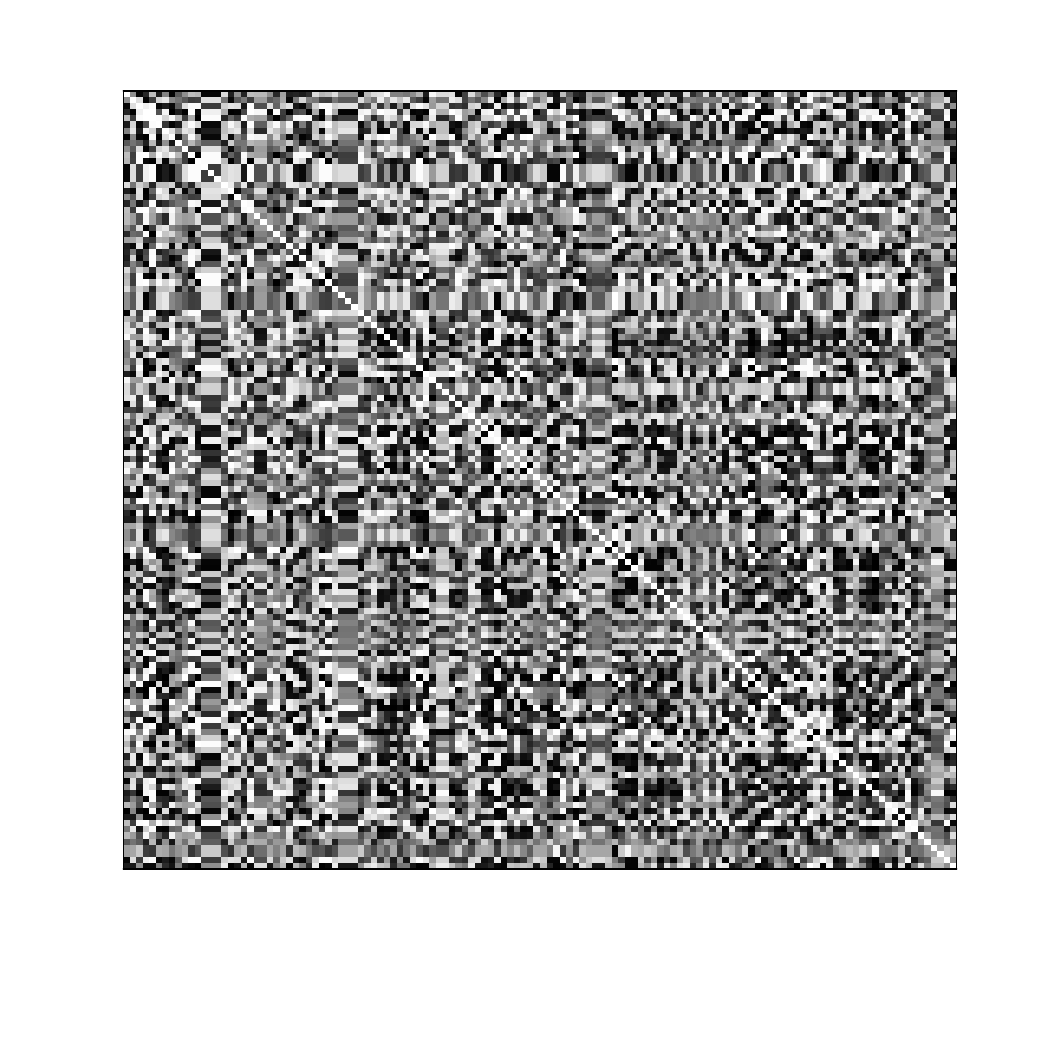} &  \includegraphics[trim={60 65 60 65},width=.22\linewidth]{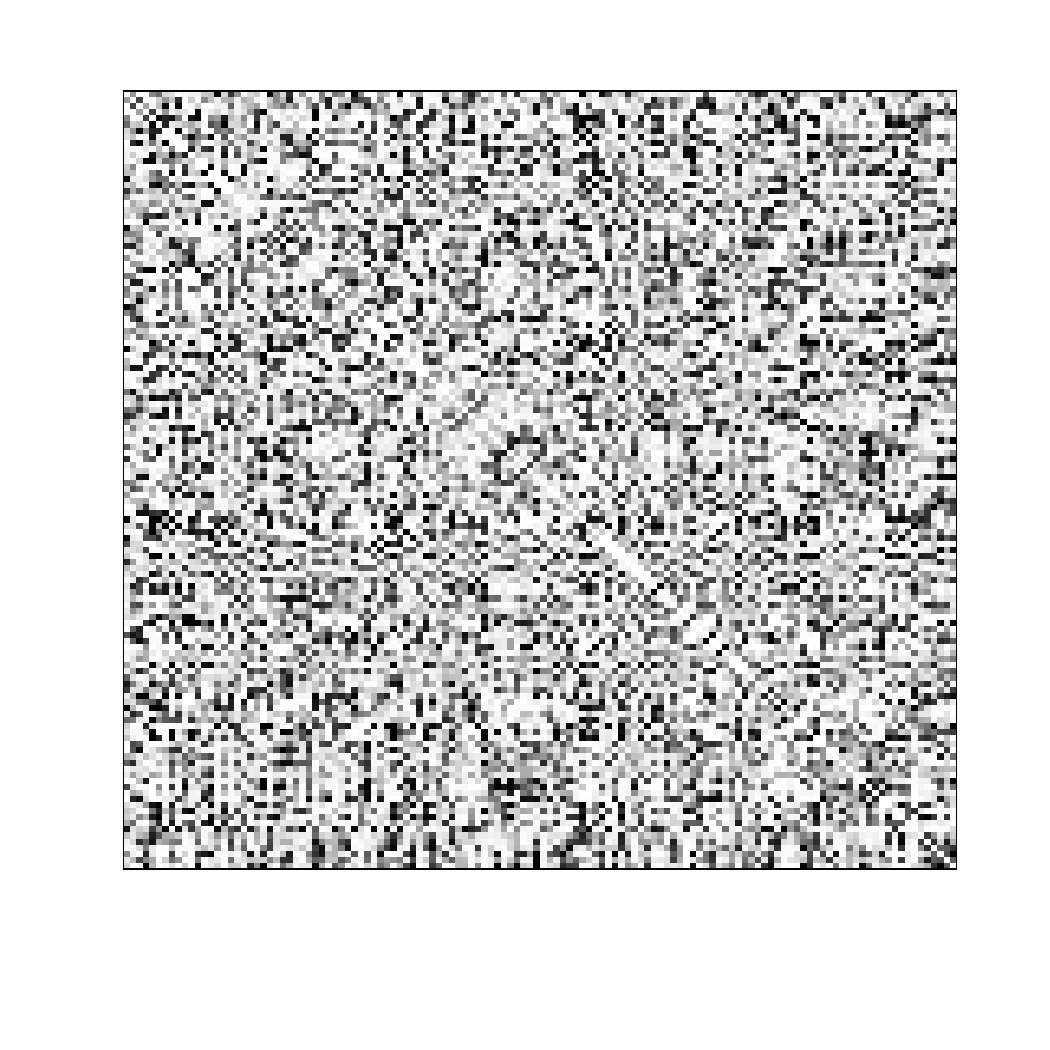} 
		\\
		\rotatebox[origin=l]{90}{\hspace{0.9cm}Estimated}&
		\includegraphics[trim={60 65 60 65},width=.22\linewidth]{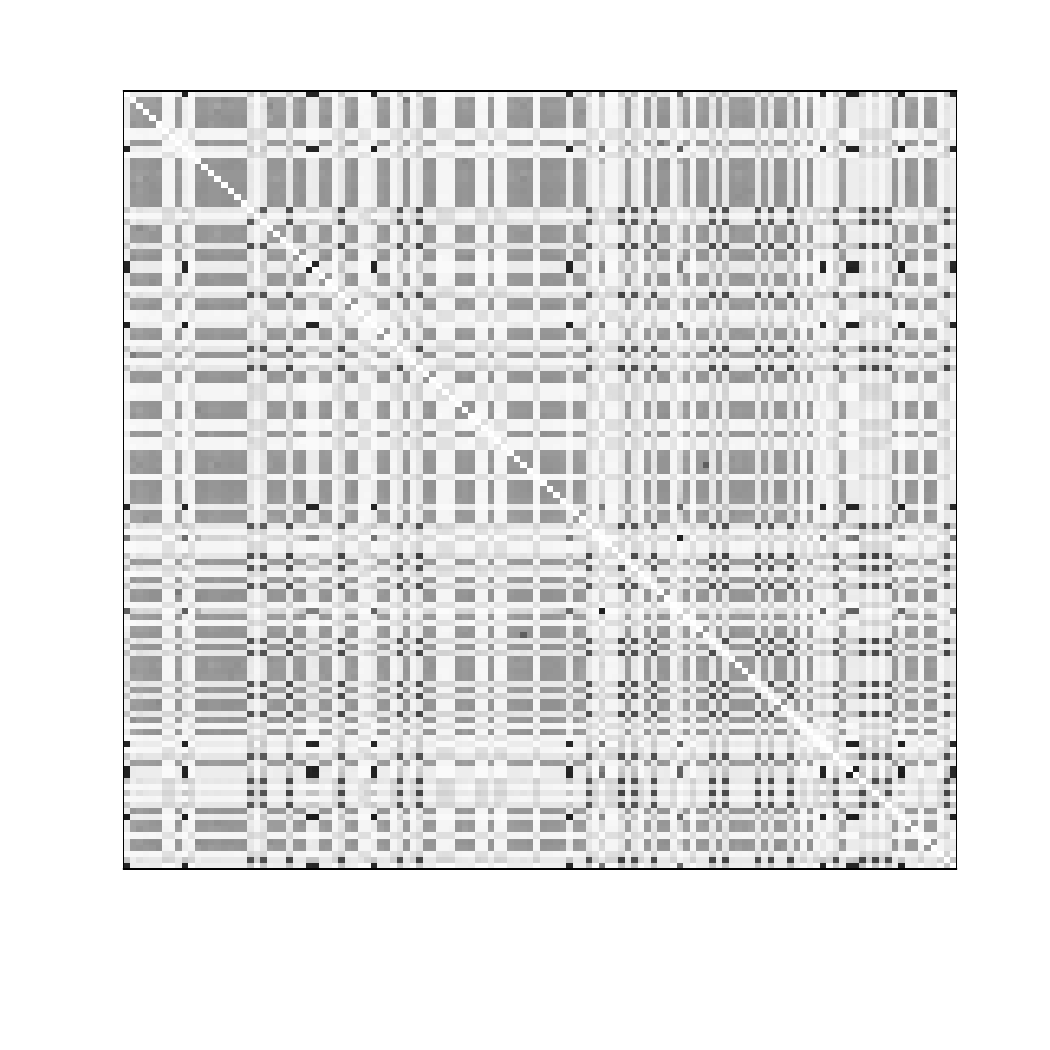} &   \includegraphics[trim={60 65 60 65},width=.22\linewidth]{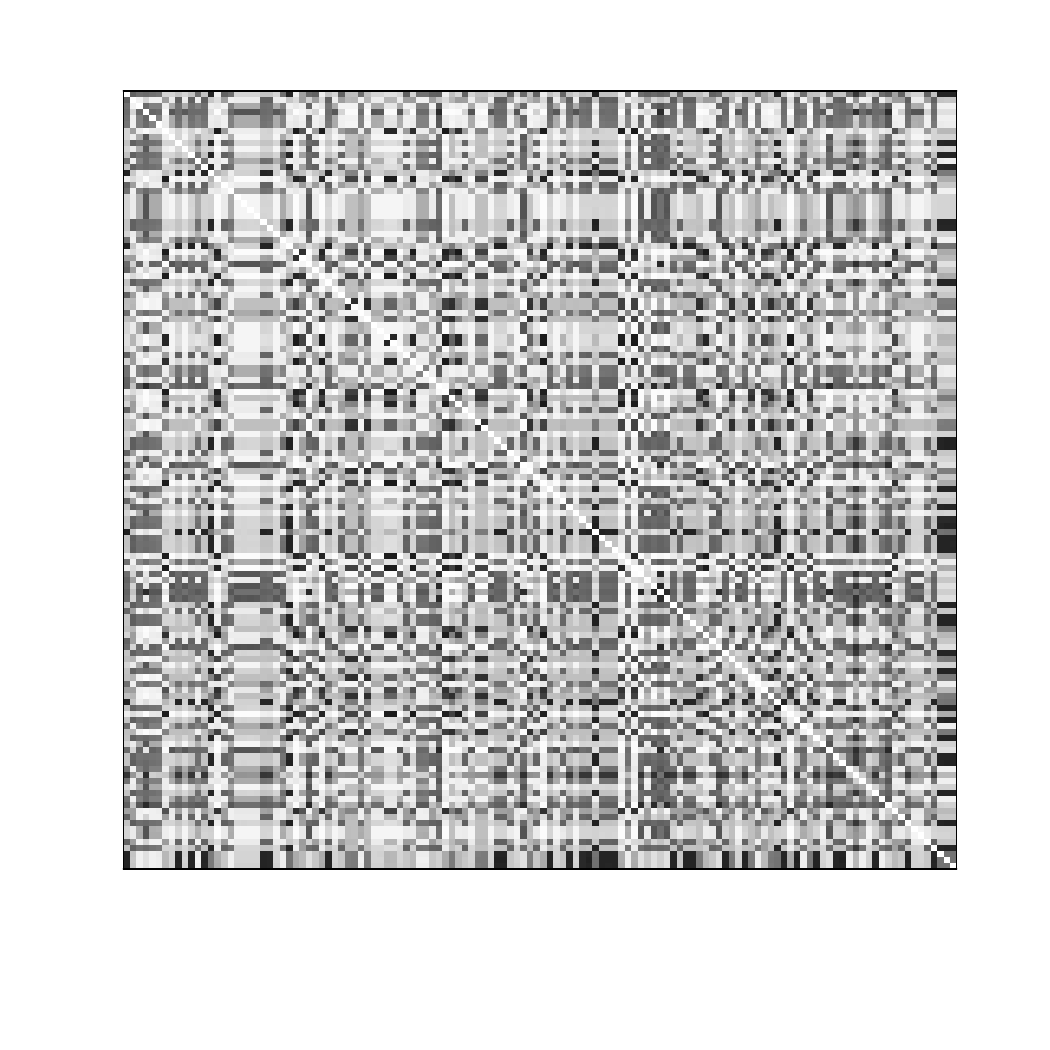} &  \includegraphics[trim={60 65 60 65},width=.22\linewidth]{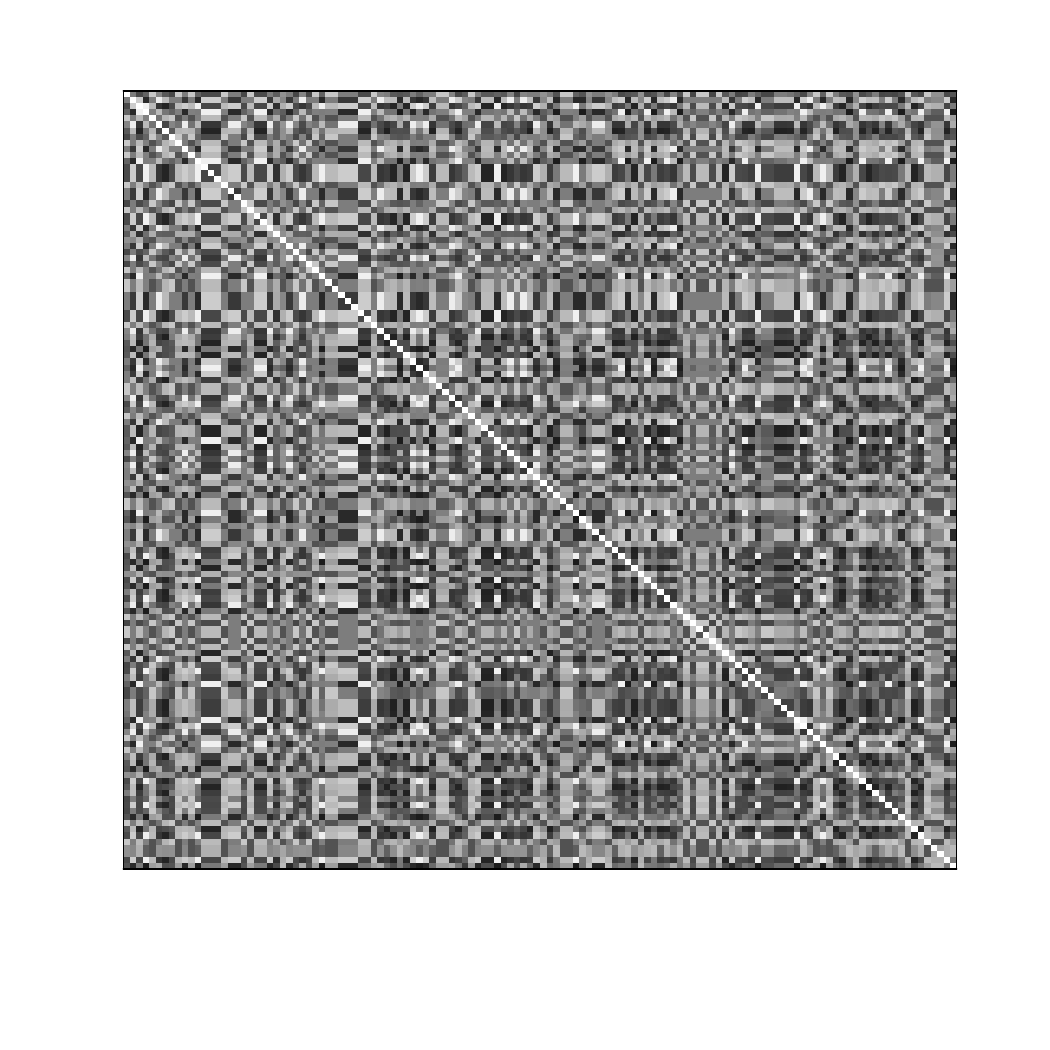} & \includegraphics[trim={60 65 60 65},width=.22\linewidth]{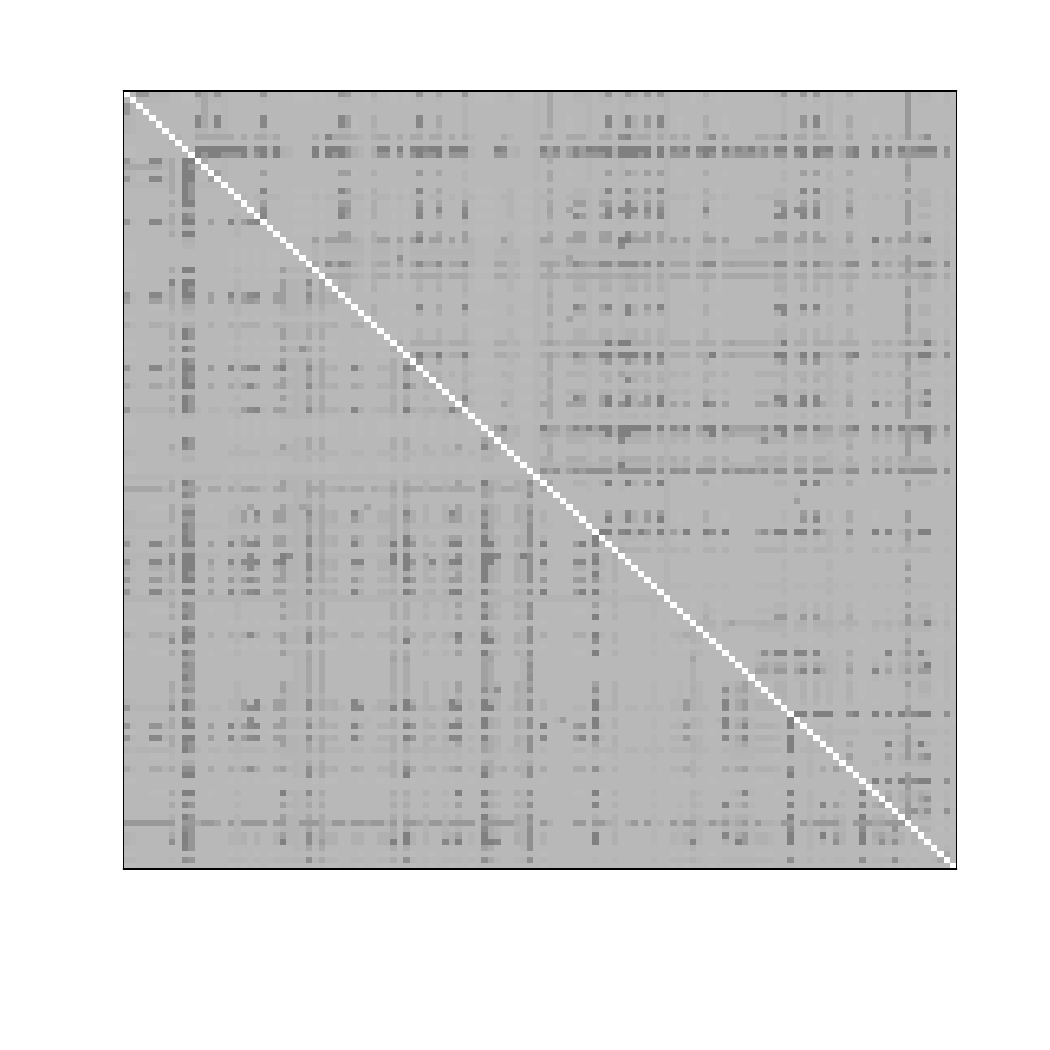} 
		
	\end{tabular} 
	\caption{Illustrating the fitting capability of the DMBN with $B=10$ blocks on data simulated from four multilayer networks with 5, 15, 45 and no blocks (DMN), as indicated in each column. All networks have $N=128$ edges, $K=4$ layers,  and $T=12$ time points. The top row displays the true link probabilities and the bottom row their estimated counterparts. The gray scale represent link probabilities going from $p(A_{ij})=0$ (white) to $p(A_{ij})=1$ (black).}
	\label{fig:sim_results}
\end{figure}

\textcolor{black}{We estimated the models specified above to every
	simulated networks ten times with random initialization of block assignments
	and latent coordinates. Estimation times and performance metrics were
	averaged accordingly. Relative mean absolute errors (MAE) for estimating
	the true link probabilities are presented in Table \ref{tab:sim_mae}. The results
	illustrate how the DMBN model takes advantage of the community structures
	to recover the link probabilities, effect that is more pronounced
	as the network size increases. For larger networks ($N=\{128,256\}$)
	and clear block structure ($B=5$) the DMBN model outperforms the
	DMN with relative MAE ratios of 3.65 and 4.20 respectively. On the
	other hand, the performance of the DMBN model decreases with granularity
	regardless of the network size. Absolute MAE's for both models are
	between 0.01 and 0.08. Note that for the DMN model the number of latent
	coordinates grows rapidly with $N$, and for networks of size $N=512$
	the estimation times exceeded the limit from the computing infrastructure,
	thus relative measures are not shown.  The fact that the block-based
	model is not able to fully recover all individual link probabilities
	at full granularity is an expected outcome since this model tries
	to summarize the dynamic of groups of links using a very limited set
	of parameters. As the granularity increases towards the worst-case-scenario
	of a multilayer network where the dynamics of each link is generated
	by its own stochastic process and the block structure vanishes, the
	DMBN is expected to be less effective to fit the data and outperformed
	by the DMN. }

Figure \ref{fig:sim_results} shows the true and estimated probabilities from the DMBN
on four dynamic multilayer networks with $N=128$ nodes and different
block structure. For each network ($B=5,15,45$, and no block structure)
the figure shows the probabilities from a randomly selected graph
out of the entire set of $TK=48$ graphs. The four images in the top
row show the true probabilities, which appear clearly structured in
(a-c) compared to the full-granularity graph in (d). In the bottom
row we see how the DMBN is able to almost perfectly recover all probabilities
when $B=5$ and $B=15$, is doing a decent job when $B=45$ and, as
expected, struggles to fit the DMN model without any block structure. 

Table \ref{tab:sim_times} presents the relative running times (originally in minutes)
for all simulations. As expected, the capacity of the DMN model to
capture network link dynamics at full granularity comes at the cost
of time. The estimation of the DMN model is significantly slower compared
to the DMBN, with running times ranging from twice ($N=32$) to more
than thirty times slower to those from the DMBN for the network with
$N=256$ nodes. For the DMBN the absolute estimation times are below
one hour in most cases, and only grow noticeably when the network
size is above $N=256$, hence demonstrating the scalability of the
model when $B\ll N$. This assumption may hold true for many real
networks, such as social or transportation networks, where community
structure naturally arises. In the next section we present a case
study using real data, and evaluate the classification performance
of the proposed model to predict markets within the US airport system,
a classic example of a hub-and-spoke network. 

\begin{table}
	\centering 
	\begin{tabular}{cccccc}
		\hline 
		& \multicolumn{5}{c}{Network Size }\tabularnewline
		\hline 
		True $B$ & 32 & 64 & 128 & 256 & 512\tabularnewline
		\hline 
		5 & 1.49 & 2.30 & 3.65 & 4.20 & -\tabularnewline
		15 & 0.92 & 1.05 & 1.25 & 1.12 & -\tabularnewline
		45 & 0.93 & 0.94 & 0.87 & 0.79 & -\tabularnewline
		No blocks & 1.13 & 1.11 & 0.92 & 0.73 & -\tabularnewline
		\hline 
	\end{tabular}\caption{Relative Mean Absolute Error (MAE) ($\mathrm{MAE}_\mathrm{DMN}/\mathrm{MAE}_\mathrm{DMBN}$) for recovering
		the true probabilities from simulated data. The DMBN model is estimated
		with $B=10$ blocks.}
	\label{tab:sim_mae}
\end{table}

\begin{table}
	\centering 
	\begin{tabular}{cccccc}
		\hline 
		& \multicolumn{5}{c}{Network Size }\tabularnewline
		\hline 
		True $B$ & 32 & 64 & 128 & 256 & 512\tabularnewline
		\hline 
		5 & 2.15 & 5.93 & 16.80 & 31.52 & -\tabularnewline
		15 & 2.19 & 5.86 & 16.38 & 32.07 & -\tabularnewline
		45 & 2.14 & 5.76 & 16.32 & 32.12 & -\tabularnewline
		No blocks & 2.11 & 5.57 & 16.24 & 31.16 & -\tabularnewline
		\hline 
	\end{tabular}\caption{Relative computing time (minutes) of the DMN model compared to the
		DMBN model with $B=10$ blocks.}
	\label{tab:sim_times}
\end{table}

\section{Application to the US air transport network}
Modeling complex transportation systems as dynamic multilayer graphs
\citep{kivela2014multilayer} has been recently attempted for e.g.
air transportation \citep{cardillo2013emergence}, public transport
\citep{gallotti2015multilayer} or maritime networks \citep{ducruet2017multilayer}.
The majority of these contributions focus on the visual inspection
of the graphs, or the temporal and multilayer analysis of the networks
by means of static, layer-wise topological measures, with no use of
statistical or machine learning models. In this case study we
apply the model from \citet{durante2017bayesian} and our community-based extension introduced in Section \ref{sec:meth} to a real airline network
with airports as nodes and airlines as layers. The dynamic and multilayer
dimensions of the network are modeled jointly in a probabilistic fashion,
and the stochastic block structure allows for interesting model-based clustering of airports. The data description is presented first, followed by two different study scenarios. In the first scenario we fit the DMN model to a small subnetwork of two airlines: Delta and Southwest. The objective is to investigate how the model estimates reflect changes in network structure due to the merger of Southwest and AirTran after 2011, and other events happened during the sample period. The second study features a full-scale forecasting application to a dynamic multilayer transport network with 80 airports and four airlines over a 10-year period. We compare both models in terms of classification performance and estimation times, and also investigate the resulting airport clusters from the DMBN. Table \ref{tab:cs_description} summarizes the experimental setup and network data utilized in each case study. All experiments were implemented using R Open 3.4.2 (MKL support), and executed on an Intel i7 Dual-Core PC with 16 GB of RAM running Windows 10.

\subsection{Data source and network description}
\begin{figure}[h] 
	\centering 
	\includegraphics[trim={0 0 0 0},width=0.65\textwidth]{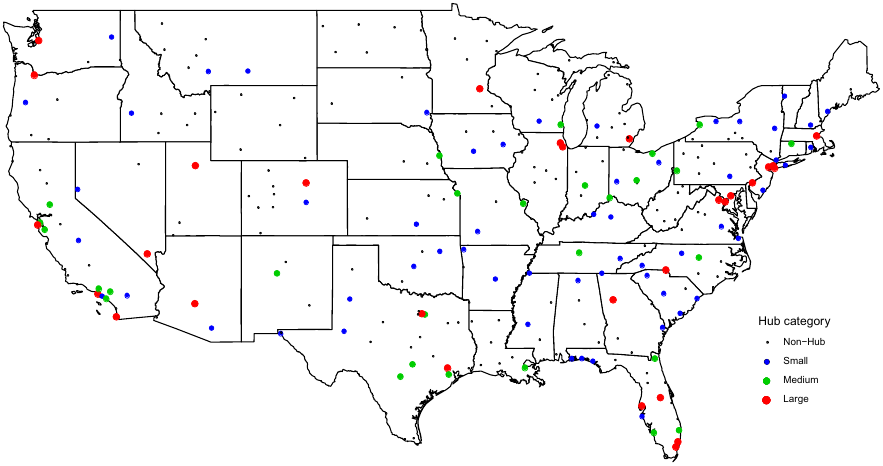} 
	\caption{Mainland airports from the US air transport network, and their classification according to the Federal Aviation Administration \citep{FAA2019}.} 
	\label{fig:cs_usairports}
\end{figure}

\begin{figure}[ht]
	\centering 
	\includegraphics[trim={0 0 0 -10},width=0.7\textwidth]{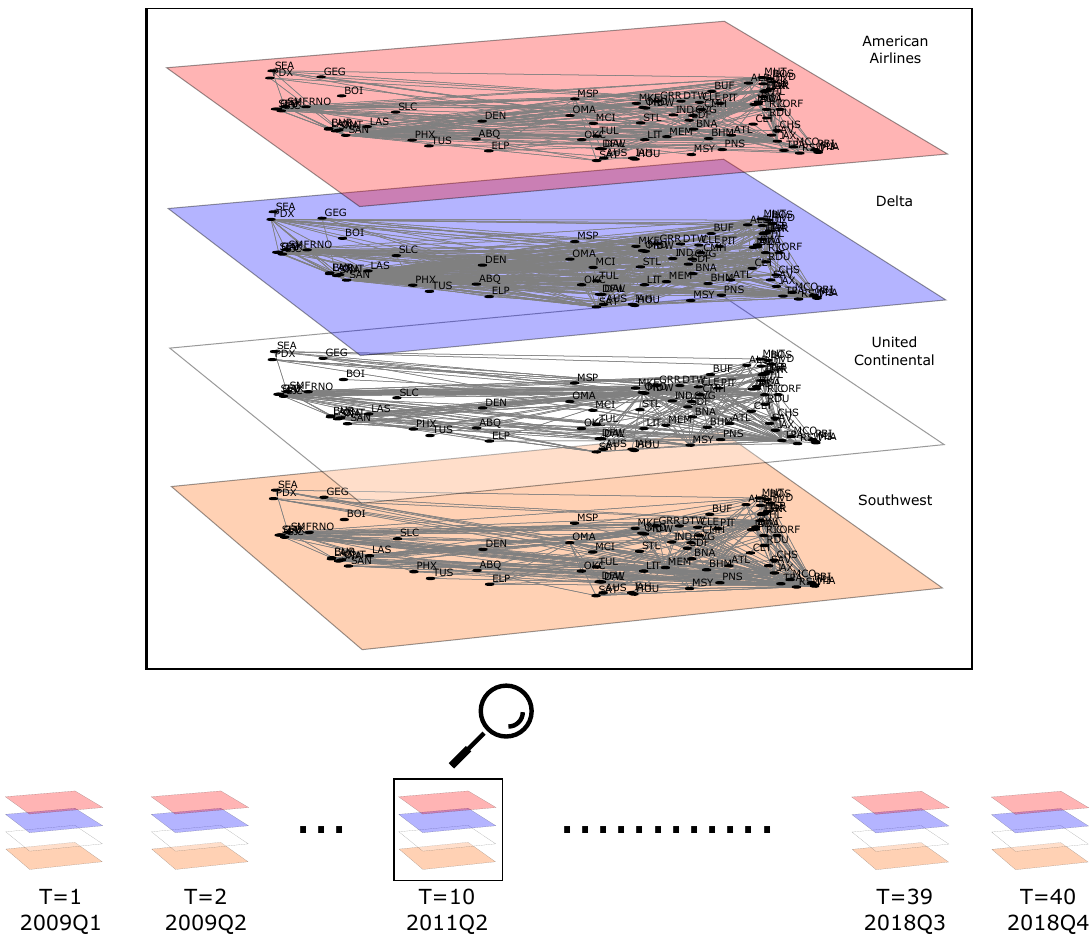}  
	\caption{Dynamic multi-layered air transport network created from quarterly airline ticket data (2009-2018), and a snapshot at $T=10$ (2011Q2).}
	\label{fig:cs_multilayer} 
\end{figure}

We collected publically available airline ticket data from the Airline
Origin and Destination Survey (\citealp{BTS2019a}), from which we
create the air transportation graphs. Figure \ref{fig:cs_usairports} shows the mainland
US airports available from the survey along with their hub classification
\citep{FAA2019}; we selected the $N=80$ airports with the largest number of (departing) flights, over a period of ten years (2009-2018). These airports concentrate around 95\% of the total network traffic.
The survey provides quarterly data so the number of time steps will
be $T=40$. To create a multilayer network, we generate separate graphs
corresponding to $K=4$ major airlines: American Airlines, Delta,
United/Continental and Southwest. Our choice of airlines is not random: after the mergers of Delta/Northwest (2009), United/Continental (2010), Southwest/AirTran (2011) and American Airlines/US Airways (2013), the resulting "big four" became the dominant airlines in the system. A graphical representation of the
multilayer structure at the second quarter of 2011 is presented in
Figure \ref{fig:cs_multilayer}, and relevant network statistics in Table \ref{tab:cs_descriptive_stats}. We generated
the multilayer graph using the library Pymnet \citep{kivela2019pymnet}.
All indicators but the number of flights are based on the corresponding
unweighted, undirected graphs. We see similar network characteristics
among the full-service carriers, whereas Southwest stands out with
a higher edge density and less centralized degree distribution, which
agrees with the tendency of low-cost airlines to relax the hub-and-spoke
model by developing a significant number of point-to-point markets \cite{doganis2019}.
Note also that there is only around 25\% percent of edge-overlap between
the different airlines, which justifies the use multilayer models.

\begin{table}[h]
	\centering 
	\begin{tabular}{ccccccc}
		\hline 
		Layer & Nodes & Edges & Density & ASPL & Degree & Flights\tabularnewline
		\hline 
		1 - AA & 80 & 579 & 0.092 & 1.948 & 14.486 & 206,078\tabularnewline
		2 - DL & 80 & 563 & 0.089 & 1.957 & 14.070 & 170,825\tabularnewline
		3 - UA & 80 & 557 & 0.088 & 1.975 & 13.912 & 125,464\tabularnewline
		4 - WN & 80 & 1,047 & 0.166 & 1.803 & 26.163 & 284,437\tabularnewline
		\hline 
		Combined & 80 & 2,190 & 0.346 & 1.669 & 54.756 & 789,163\tabularnewline
		\hline 
	\end{tabular}\caption{Quarterly-averaged network statistics for the selected sample (2009-2018).
		AA: American Airlines, DL: Delta Airlines, UA: United Continental,
		WN: Southwest, ASPL: Average shortest path length. }
	\label{tab:cs_descriptive_stats}
\end{table}

\begingroup
\setlength{\tabcolsep}{4pt} 
\begin{table}
	\centering
	\begin{tabular}{ |l|ccccccccccc| } 
		\hline
		Description	 & \multicolumn{11}{|c|}{Selected graphs by quarter and year}\\
		\hline
		\textbf{Case study 1: model analysis} & &09 &10 &11 &12 &13 &14 &15 &16 &17 &18\\
		Only DMN model	 &Q1 &$\times$ &$\times$ &$\times$ &$\times$ &$\times$ &$\times$ &$\times$ &$\times$ &$\times$ &$\times$\\
		Only Delta/Southwest (K=2)	 &Q2 & & & & & & & & & &\\
		Subnetwork of N=40 airports	 &Q3 & & & & & & & & & &\\
			 &Q4 & & & & & & & & & &\\
		\hline
		\textbf{Case study 2: forecasting}	 & &09 &10 &11 &12 &13 &14 &15 &16 &17 &18\\
		Both DMN/DMBN models	 &Q1 &$\times$ &$\times$ &$\times$ &$\times$ &$\times$ &$\times$ &$\times$ &$\times$ &$\times$ &$\circ$\\
		All airlines (K=4)	 &Q2 &$\times$ &$\times$ &$\times$ &$\times$ &$\times$ &$\times$ &$\times$ &$\times$ &$\times$ &$\circ$\\
		All airports (N=80)	 &Q3 &$\times$ &$\times$ &$\times$ &$\times$ &$\times$ &$\times$ &$\times$ &$\times$ &$\times$ &$\circ$\\
		 	 &Q4 &$\times$ &$\times$ &$\times$ &$\times$ &$\times$ &$\times$ &$\times$ &$\times$ &$\times$ &$\circ$\\
		\hline
	\end{tabular}
	\caption{General description of the case studies and the network data used. Fit/training graphs are marked with a cross ($\times$), and test data with ($\circ$).}	
	\label{tab:cs_description} 
\end{table}
\endgroup

\subsection{Merger of Southwest and AirTran (2011)}
When Southwest acquired AirTran in 2011, it was the first merger between low-cost carriers in US history. Prior to the merge, Southwest 
carried almost 90 million passengers in 2010, more than triple than AirTran, and operated point-to-point markets, in contrast to the hub-and-spoke structure of AirTran's network. The operation was appealing for Southwest, which not only had the opportunity to take over the network of its 
competitor, but also to enter Atlanta, which was at the time the only major US city without a strong presence of Southwest. This 
would help Southwest solidify its position as a dominant domestic low-cost carrier while adding some new international routes, mainly in the Caribbean, to its network. In this study, we investigate changes in the network graphs of Southwest and Delta, the dominant airline at Atlanta, as a consequence of the merger, and other potential causes between 2009-2018, using the dynamic multilayer model of \citet{durante2017bayesian}. 

To that end, we fit the DMN model to a selected subset of the full multilayer graph shown in Figure \ref{fig:cs_multilayer}, featuring $N=40$ airports from Delta and Southwest ($K=2$) networks, and keeping only the first quarter of every year ($T=10$), see Table \ref{tab:cs_description}. Note that even though we will estimate $N(N-1)KT/2=15,600$ logits, this scenario is rather conservative compared to the half-million required for the full problem, hence the need for the community-based extension. We draw $5,000$ posterior samples using similar hyperparameter values as for the simulations in Section \ref{sec:sim}, i.e. $R=H=6$, with common smoothness $l_{\mu}=l_{\bar{x}}=l_{x}=0.05$ over all components. Due to the bilinear forms in Eq. \eqref{eq:DuranteModel} the latent coordinates $\bar{x}(t)$ and $x^k(t)$ are not identified and their estimates should not be interpreted directly. We follow \citet{hoff2005bilinear} and calculate the posterior means of $\bar{x}(t)^T\bar{x}(t)$ for each $t$ instead, which are identified, and then obtain estimators $\hat{\bar{x}}(t)$ by truncating the spectral decompositions at the $R$-th largest eigenvalue. The within-layer coordinates $x^k(t)$ are recovered using the same procedure, for every layer and time point. Since the estimated coordinates contain information about the connectivity profile of every airport $i$, it is straightforward to define cross and within layer vertex connectivity scores (VCS) as the euclidean norm of the coordinates, i.e.

\begin{subequations}\begin{align}
	\text{CL-VCS}_i(t)=\lVert \hat{\bar{x}}_i(t)\rVert\\
	\text{WL-VCS}^k_i(t)=\lVert \hat{x}^k_i(t)\rVert
\end{align}\end{subequations}

\begin{figure}[h] 
	\centering 
	\begin{tabular}{cc}   
		\includegraphics[trim={20 30 20 30},width=.43\linewidth]{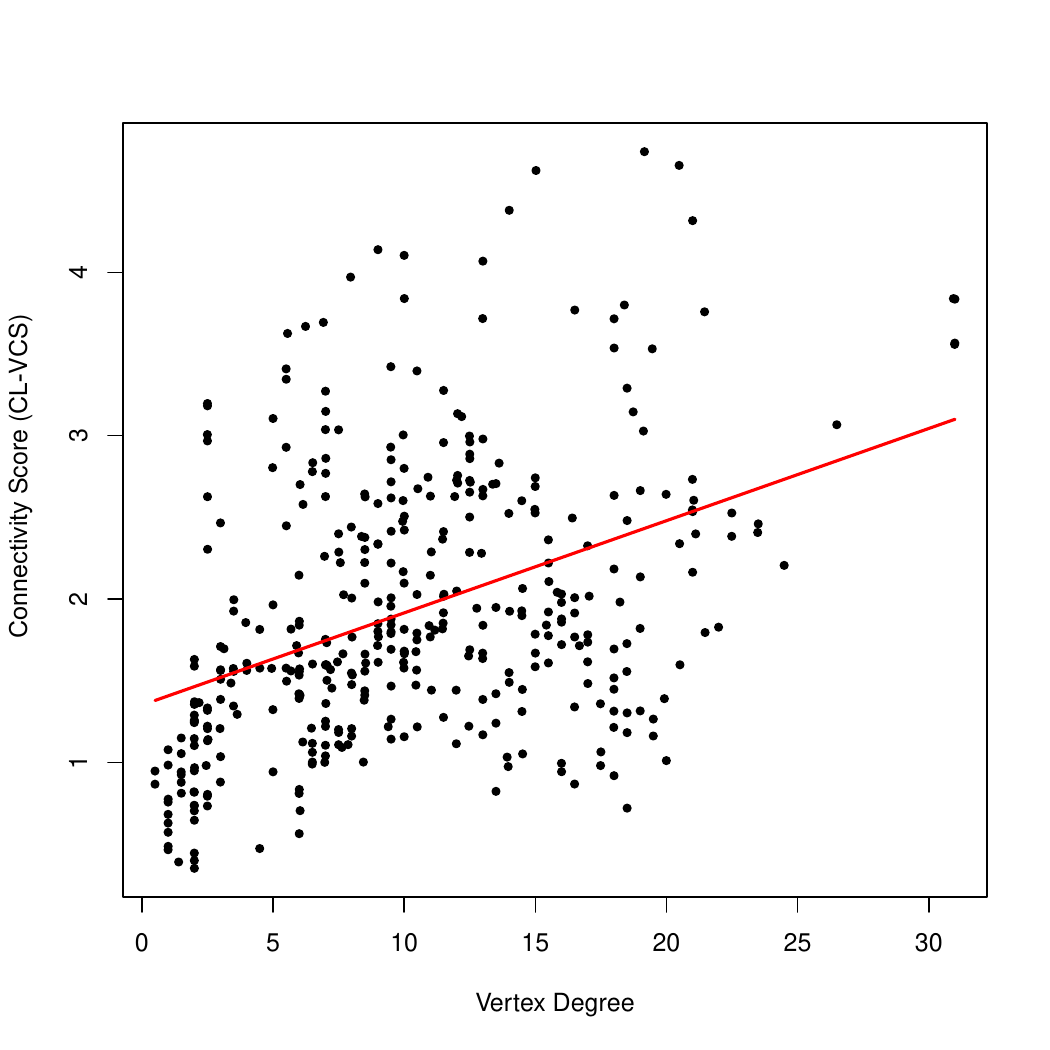} &   \includegraphics[trim={20 30 20 30},width=.43\linewidth]{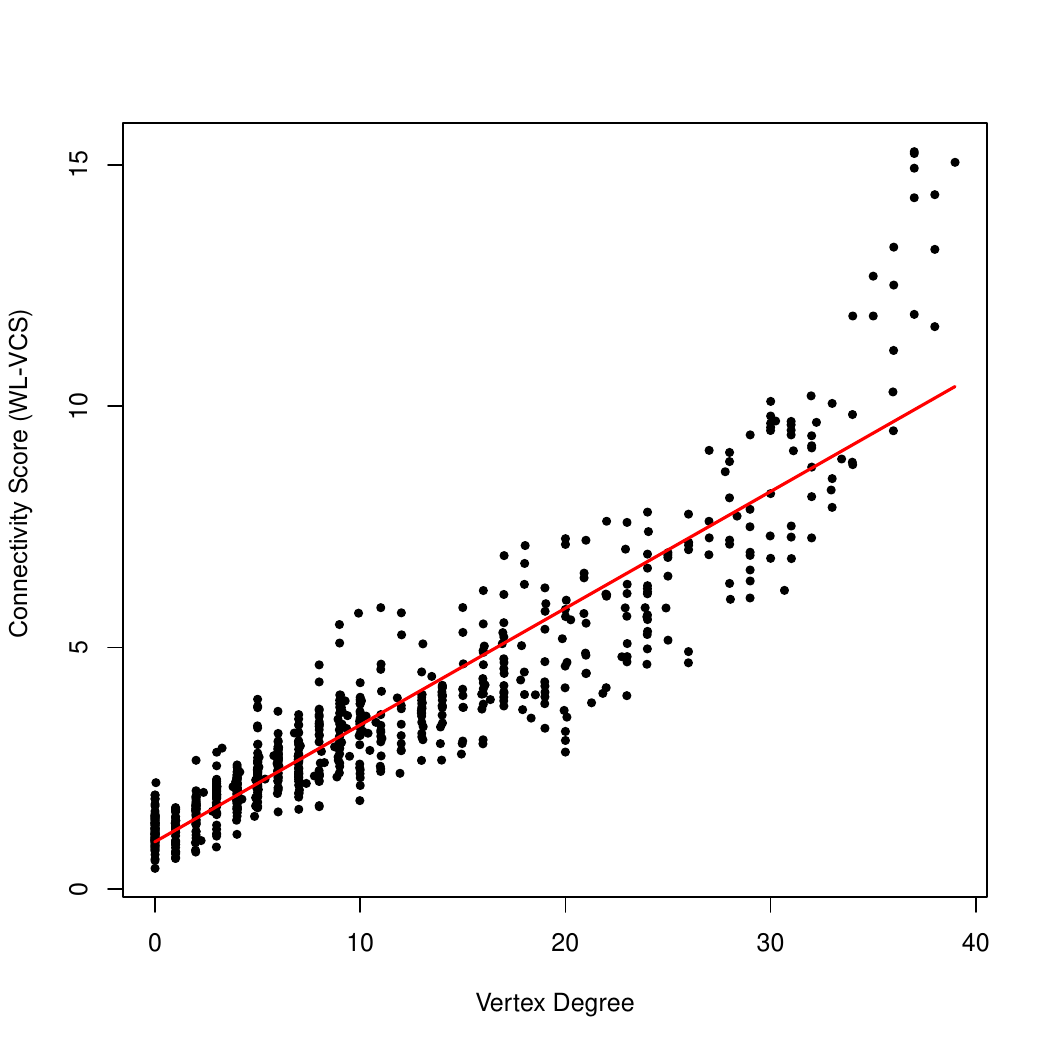}
	\end{tabular} 
	\caption{Estimated Vertex Connectivity Scores (VCS) for all airports, time points and layers, against the original airport degrees. Left: cross-layer scores. Right: within-layer scores (both Southwest and Delta).}
	\label{fig:cs_vps}
\end{figure}

Figure \ref{fig:cs_vps} present both scores for all airports in the sample as a function of the number of connections (i.e. degree). There seems to be a strong linear correlation between the within-layer scores and the degree of an airport, which is not present in the cross-layer estimates. This is probably related to our choice of mixing a legacy (hub-and-spoke) and a low-cost (point-to-point) carrier in the experiment. If airlines are significantly different in their structure and routes, their connectivity patterns will be absorbed by the layer effects, and the interpretation of the cross-layer connectivity scores may not be possible by direct inspection. We compared the scores against more elaborated centrality indicators such as betweenness and closeness, with similar results. The fact that the latent coordinates, which can be projected to future time steps, are related to topological properties from the underlying graph that determine the efficiency of air transportation in terms of e.g. passenger-kilometers \cite{kotegawa2014trc}, is an interesting outcome from the model.

\begin{figure}[h] 
	\centering 
	\begin{tabular}{cc}   
		\includegraphics[trim={20 30 20 30},width=.43\linewidth]{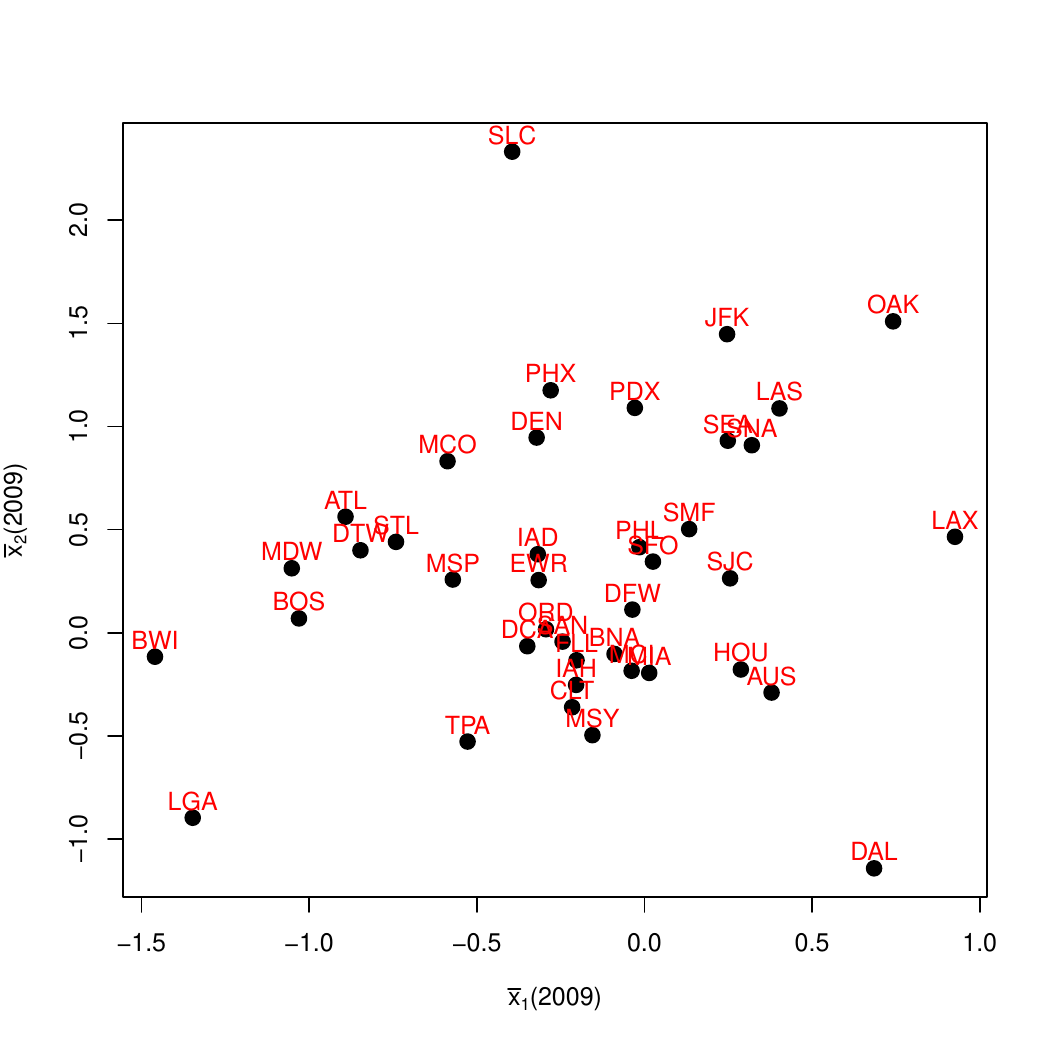} &   \includegraphics[trim={20 30 20 30},width=.43\linewidth]{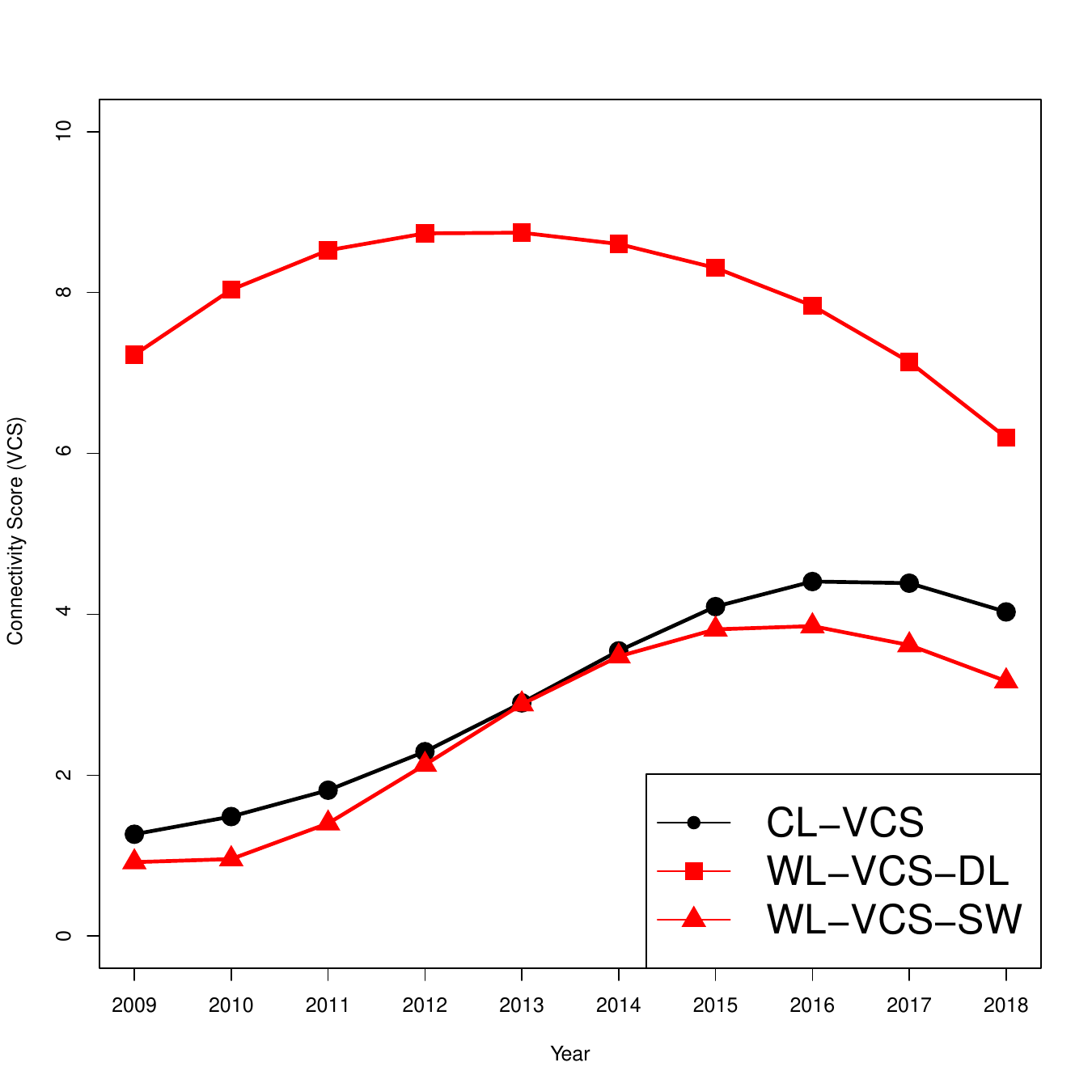}
	\end{tabular} 
	\caption{Left: estimated cross-layer coordinates (two largest components) for all airports in 2009. Right: Vertex Connectivity Scores (VCS) for Atlanta-Hartsfield. See \ref{sec:iatacodes} for the IATA airport codes.}
	\label{fig:cs_vps_atl}
\end{figure}

In Figure \ref{fig:cs_vps_atl} (left) we look directly at the two largest components from the estimated cross-layer coordinates for all airports in 2009. Even though there is no apparent clustering structure, there exists a general pattern based on airline dominance at the airports, and the fact that we left American Airlines and Continental out of the experiment. Airports that are close to the origin are mostly American (e.g. Miami, Dallas Fort-Worth), US Airways (e.g. Philadelphia) or United/Continental (e.g. Houston, Chicago O'Hare)  hubs, whereas the most outlying airports are hubs for Delta (e.g. Atlanta, Salt Lake City) or focus cities for Southwest (e.g. Midway), with the exception of Los Angeles and Boston. In this case, a visual inspection of the coordinates for a specific year and quarter is sufficient, but a proper cluster analysis would probably be the best choice for their interpretation over time. Back to the vertex scores, Figure \ref{fig:cs_vps_atl} (right) presents the dynamic evolution of the cross-layer and within-layer VCS for Atlanta-Hartsfield. We see how the within-layer score for Southwest grows steadily after the merger with AirTran in 2011 and the subsequent expansion at Atlanta. As a consequence, the cross-layer score is pushed up as both airlines hold now an important presence in that vertex, while the score for Delta decreases slightly though remaining high. Similarly, once the cross-layer score is significantly increased, Southwest's score starts to diminish as well, which could be an indication on how the model tries to leverage the different components of the logit during estimation.

\begin{table}[h]
	\centering 
	\begin{tabular}{ccccccc}
		\hline 
		Rank & CL-2009 & CL-2018 & DL-2009 & DL-2018 & SW-2009 & SW-2018\tabularnewline
		\hline 
		1	& SLC	& $\uparrow$ATL$\uparrow$	& ATL	& ATL	& MDW	& MDW\tabularnewline
		2	& LAX	& LAX 	& SLC	& SLC	& DEN	& $\uparrow$DAL$\uparrow$\tabularnewline
		3	& DAL	& LGA	& JFK	& MSP	& LAS	& DEN\tabularnewline
		4	& OAK	& $\uparrow$MSP$\uparrow$	& LGA	& DTW	& PHX	& STL\tabularnewline
		5	& LGA	& SLC	& DTW	& JFK	& HOU	& PHX\tabularnewline
		6	& MSY	& $\uparrow$MCO$\uparrow$	& MSP	& LAX	& AUS	& BNA\tabularnewline
		7	& BWI	& DAL	& BOS	& LGA	& MCI	& LAS\tabularnewline
		8	& JFK	& LAS	& LAX	& $\uparrow$SEA$\uparrow$	& BWI	& AUS\tabularnewline
		9	& DTW	& BWI	& MCO	& BOS	& STL	& HOU\tabularnewline
		10	& LAS	& OAK	& TPA	& $\uparrow$LAS$\uparrow$	& BNA	& BWI\tabularnewline
		\hline 
	\end{tabular}\caption{Top 10 airports according to estimated vertex connectivity scores 2009-2018. CL: cross-layer VCS, DL: within-layer VCS for Delta, SW: within-layer VCS for Southwest. Up-arrows $\uparrow$X$\uparrow$ indicate that airport X climbed the ranks up to the top-10 between 2009-2018.}
	\label{tab:cs_top10}
\end{table}
\begin{figure}[h] 
	\centering 
	\begin{tabular}{ccc}   
		\includegraphics[trim={20 30 20 30},width=.3\linewidth]{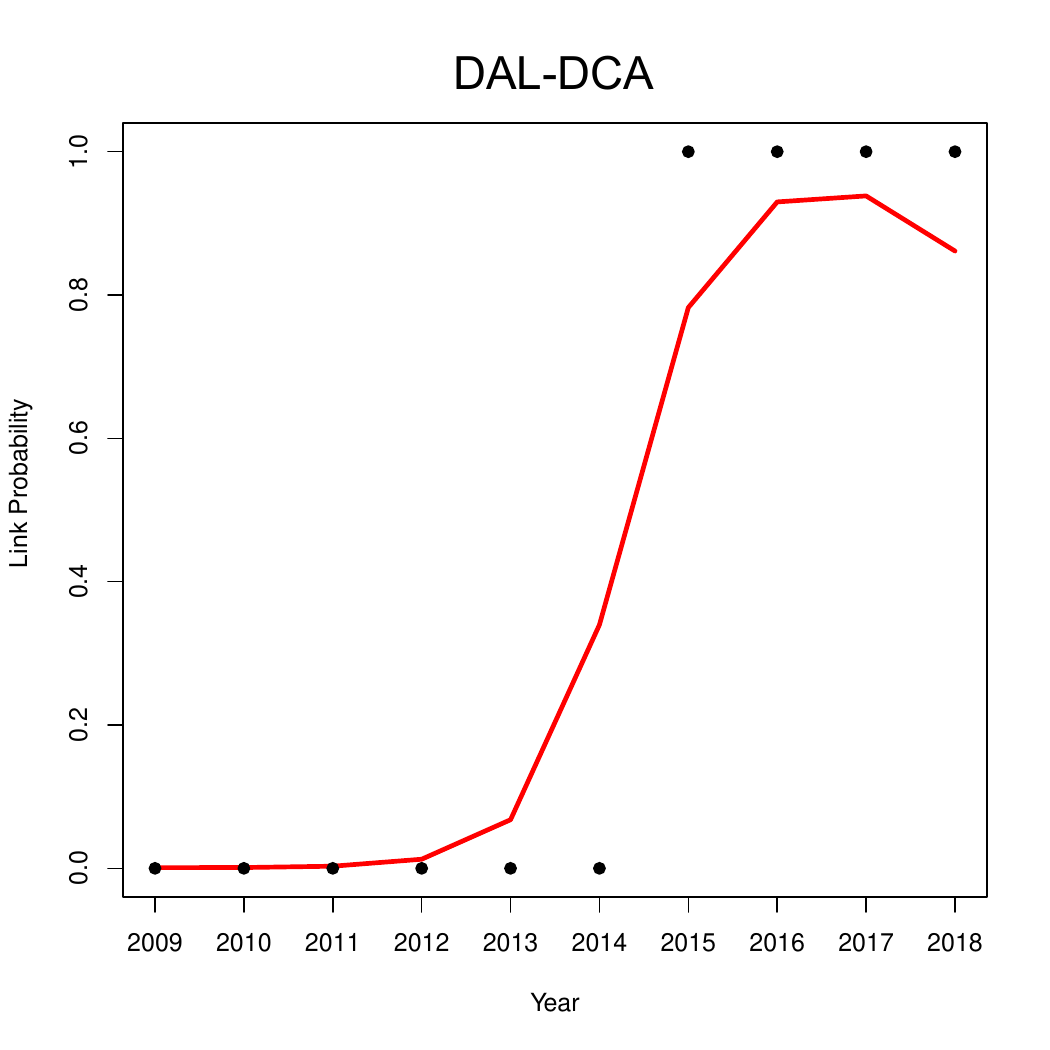} &   \includegraphics[trim={20 30 20 30},width=.3\linewidth]{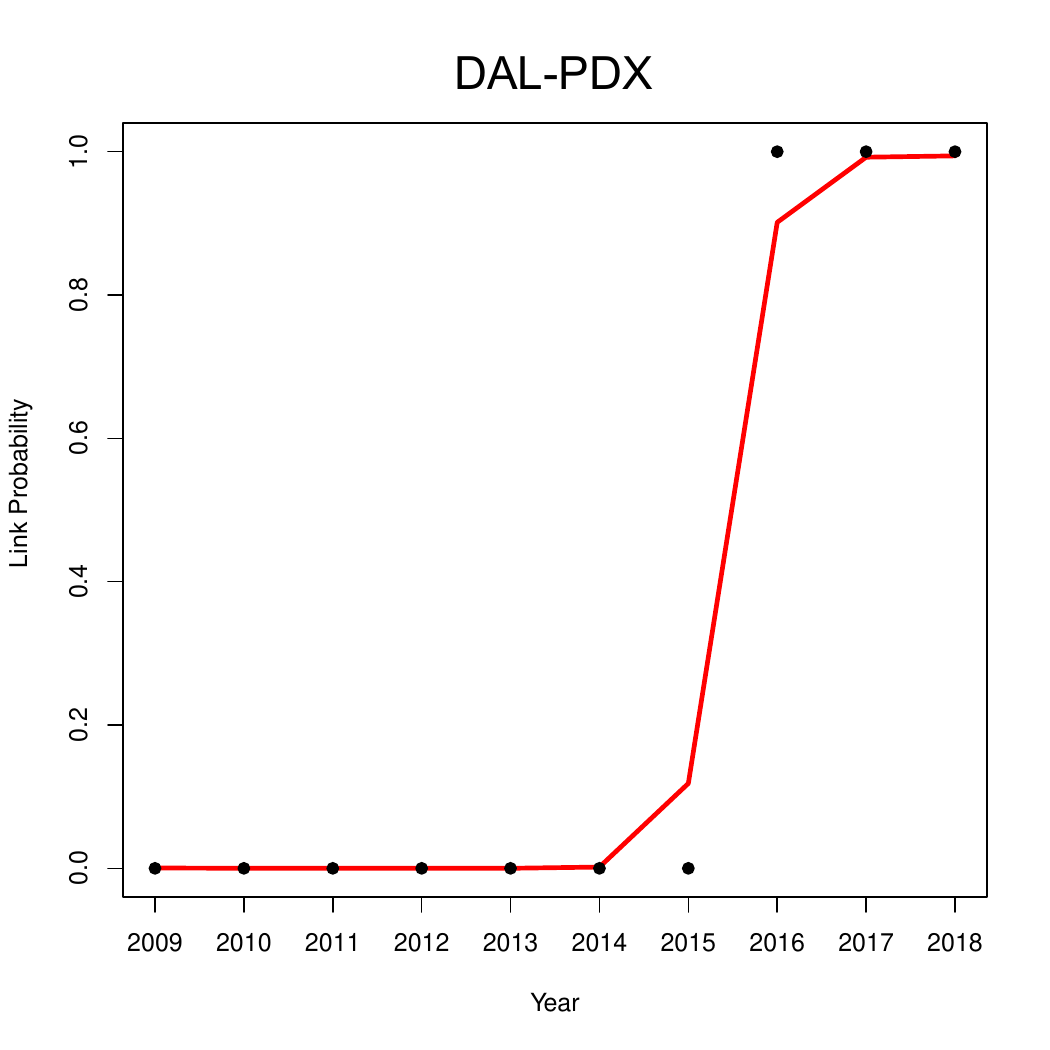} &
		\includegraphics[trim={20 30 20 30},width=.3\linewidth]{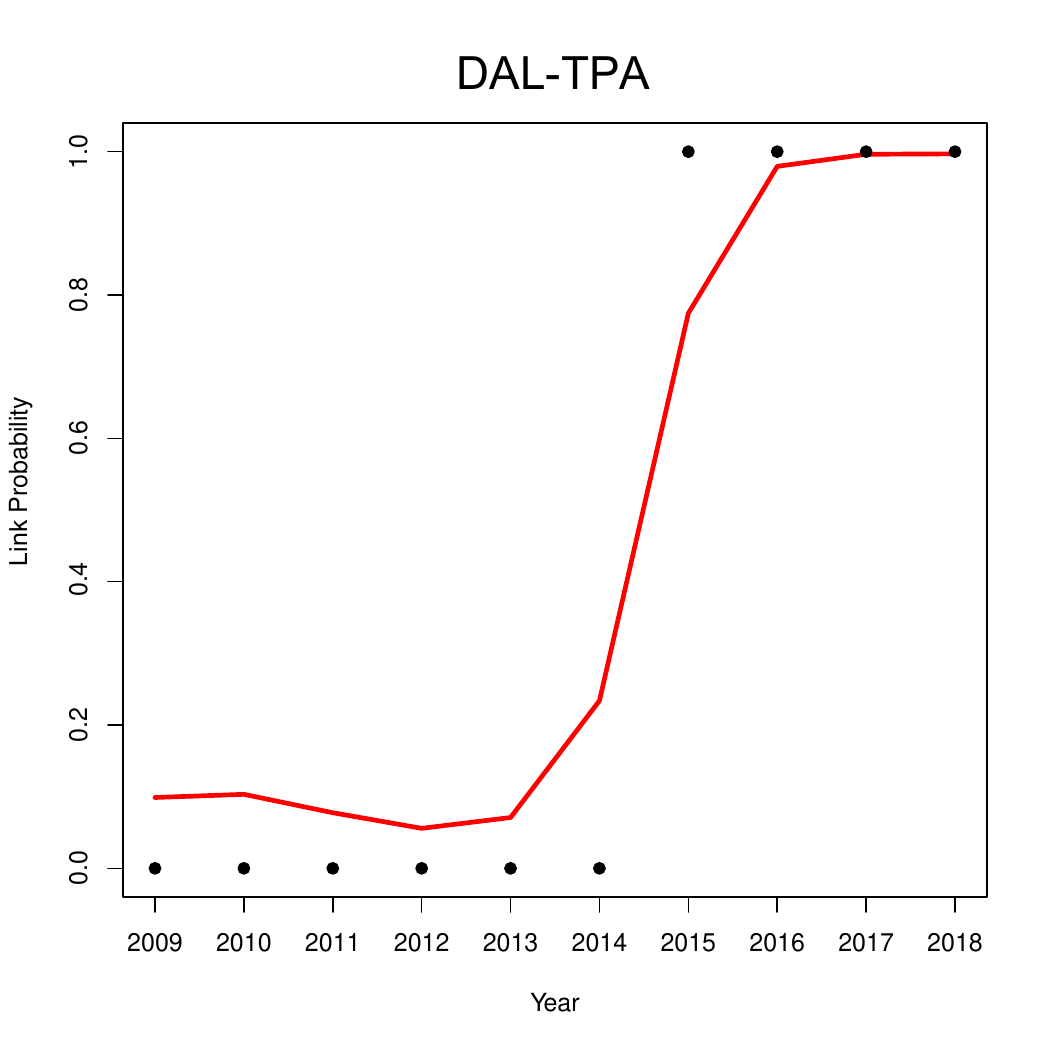}\\
		\includegraphics[trim={20 30 20 0},width=.3\linewidth]{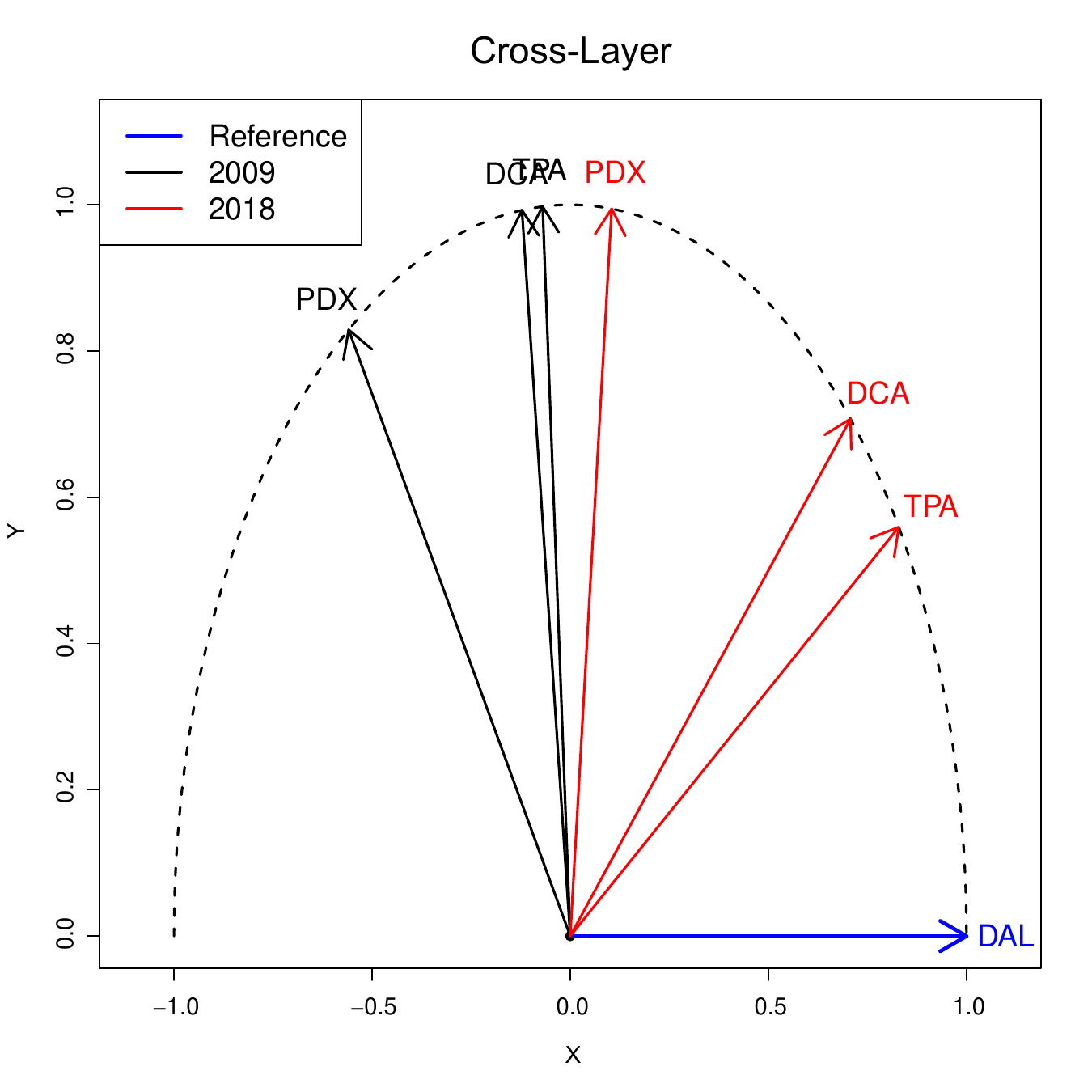} &   \includegraphics[trim={20 30 20 0},width=.3\linewidth]{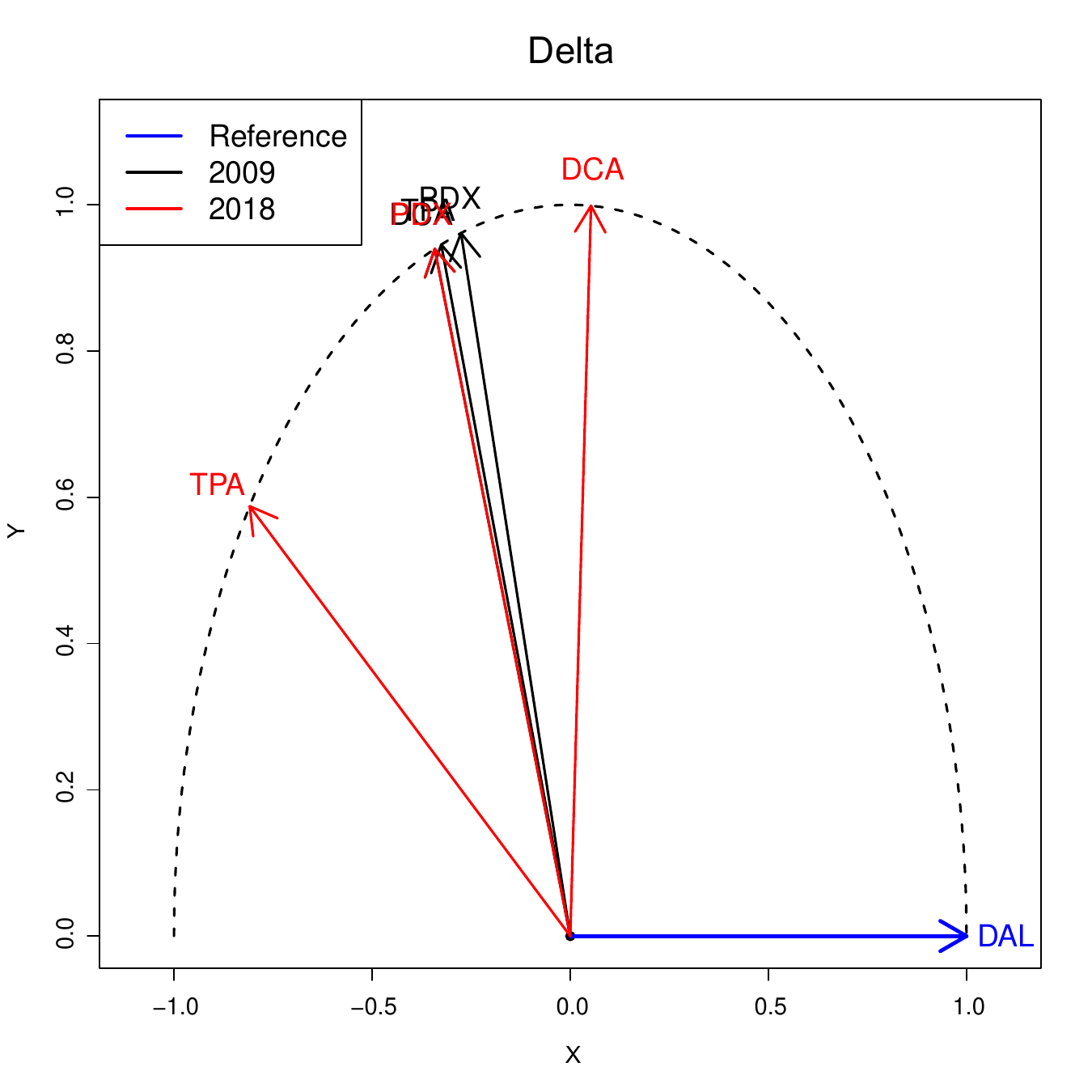} &
		\includegraphics[trim={20 30 20 0},width=.3\linewidth]{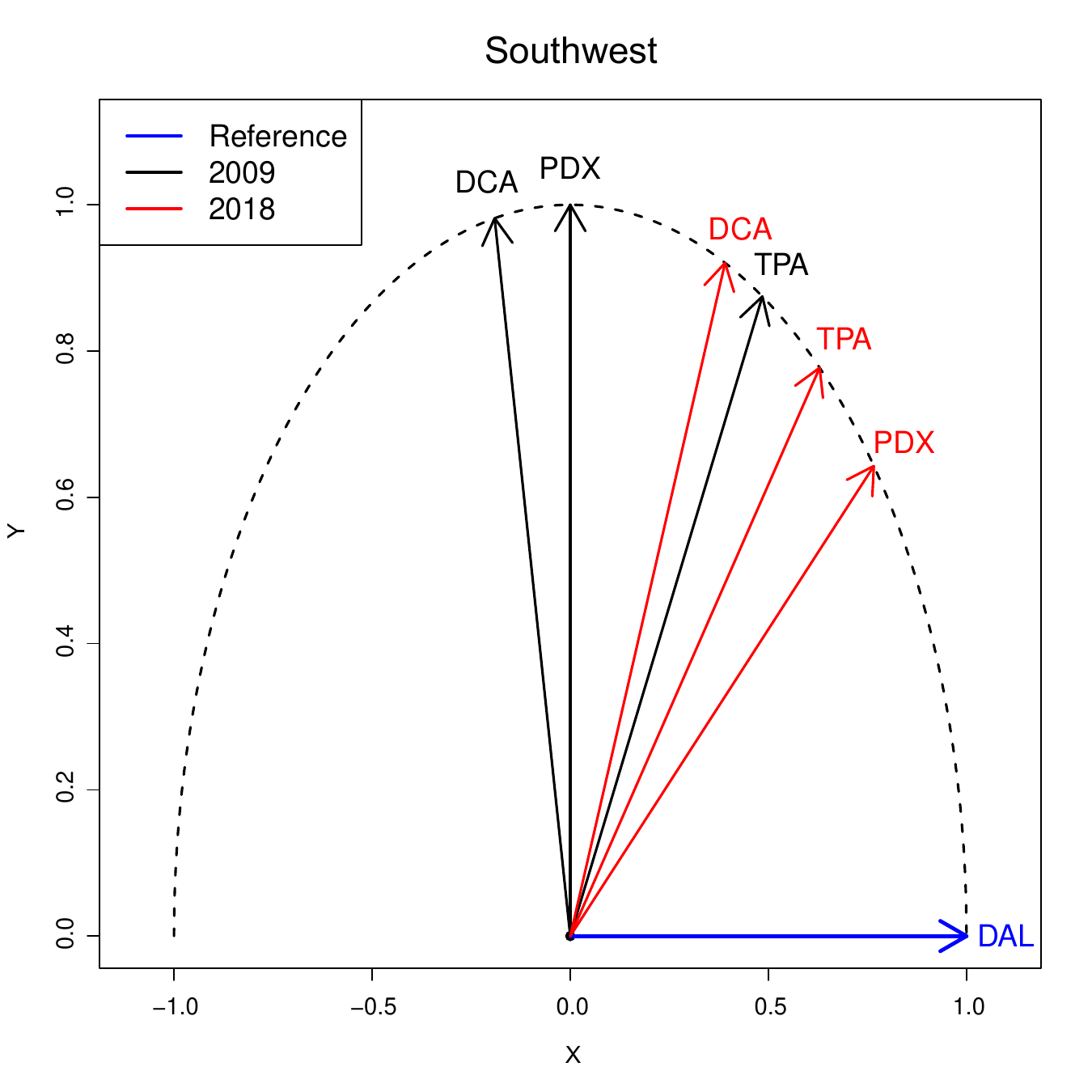}
	\end{tabular} 
	\caption{Top row: Estimated link probabilities for three new Southwest routes from Dallas Love-Field (DAL): Washington-Reagan (DCA - left), Portland (PDX - center) and Tampa (TPA - right). Black dots are the original data in the corresponding adjacency matrices. Bottom row: Angles between the estimated latent vectors of DAL, and the rest of  airports, 2009-2018. We use normalized vectors within the unit circle to represent angle differences between cross-layer (left) and within-layer (Delta and Southwest, center and right, respectively) coordinates.}
	\label{fig:cs_prob_dal}
\end{figure}

Table \ref{tab:cs_top10} presents airport rankings according to estimated vertex connectivity indices for the sample period. Overall, we get the expected actors at the top of the ranks, particularly at the airline (layer) level: Atlanta, Salt Lake City and Minneapolis lead Delta's network, whereas Chicago Midway, Denver, Las Vegas and Dallas-Love are the top performers for Southwest. In the cross-layer rankings, we notice the rising of Minneapolis and Atlanta between 2009-2018, which is probably due to their acquisition by Delta and Southwest after the mergers with Northwest (2008) and AirTran (2011), respectively. In 2014, Delta increased its presence at Seattle airport in order to open a transoceanic hub, therefore entering into direct competition in the domestic market with Alaska Airlines at their major base. This is reflected on Delta's within layer rankings with Seattle climbing from the $19$-th to the $8$-th position. From the Southwest's perspective, Dallas Love-Field appears as the airport with largest increase in within-layer connectivity. This is an expected result: after the repealing of 1979's Wright Amendment, which prohibited flights from Dallas Love-Field outside Texas and its neighboring states in order to protect Dallas Fort-Worth, Southwest greatly expanded at the former with 15 new routes in late 2014 \cite{CAPA2015}. In Figure \ref{fig:cs_prob_dal} (top row) we illustrate the model's ability to capture the network's dynamics during this expansion of Southwest at Love-Field, focusing on the new routes to/from Washington-Reagan, Portland and Tampa. Note how the underlying Gaussian processes that drive the logits adjust rapidly to changes in the network in all cases. Here the model's flexibility in latent space has clearly paid off, despite the lack of exogenous (e.g. passenger traffic) information outside the adjacency matrices. We have experimentally verified that the estimated link probabilities are not sensitive to reasonable changes in the degree of shrinkage of the latent coordinates (as controlled by the hyperparameters $a_1,a_2$ in the Gamma prior), which control the effective dimensionality of the latent space. Similarly, the estimates are also robust to variation in the size of the latent coordinates $R,H$, and the prior smoothness of the corresponding Gaussian processes. It is important to remark that the flexibility in latent space is due not only to the magnitude (e.g. connectivity scores) but also to the alignment (i.e. angles) of the vectors. For instance, the estimated cross-layer components can be written as $\hat{\bar{x}}_i(t)^T\hat{\bar{x}}_j(t)=\cos\theta_{ij}(t)\times\text{CL-VCS}_i(t)\times\text{CL-VCS}_j(t)$, where the angles $\theta_{ij}(t)$ between the latent airport coordinates are a function of time. We finally corroborate this for our example in Figure \ref{fig:cs_prob_dal} (bottom row): the cross-layer and layer-wise (Southwest) latent components of the new markets (2009-2018) are closer in angle to those of Dallas Love-Field.

\subsection{Dynamic multilayer forecasting of the US air transport network}

Despite its effectiveness, as we have just confirmed, fitting the original DMN model to very large graphs is not practical due to the over-parametrization and subsequent high estimation times. Here we test our proposed model extension by fitting and forecasting a large dynamic multilayer graph using all available airline data. The complete network features $N=80$ airports, $K=4$ airlines and $T=40$ time steps corresponding to quarters between 2009-2018. We compare both models in terms of classification accuracy and estimation times, and also investigate whether the block-wise structure of the DMBN model is able to reveal meaningful airport communities.

\begin{figure}
	\centering 
	\begin{tabular}{cccc}   
		\includegraphics[trim={20 0 20 -10},width=.38\linewidth]{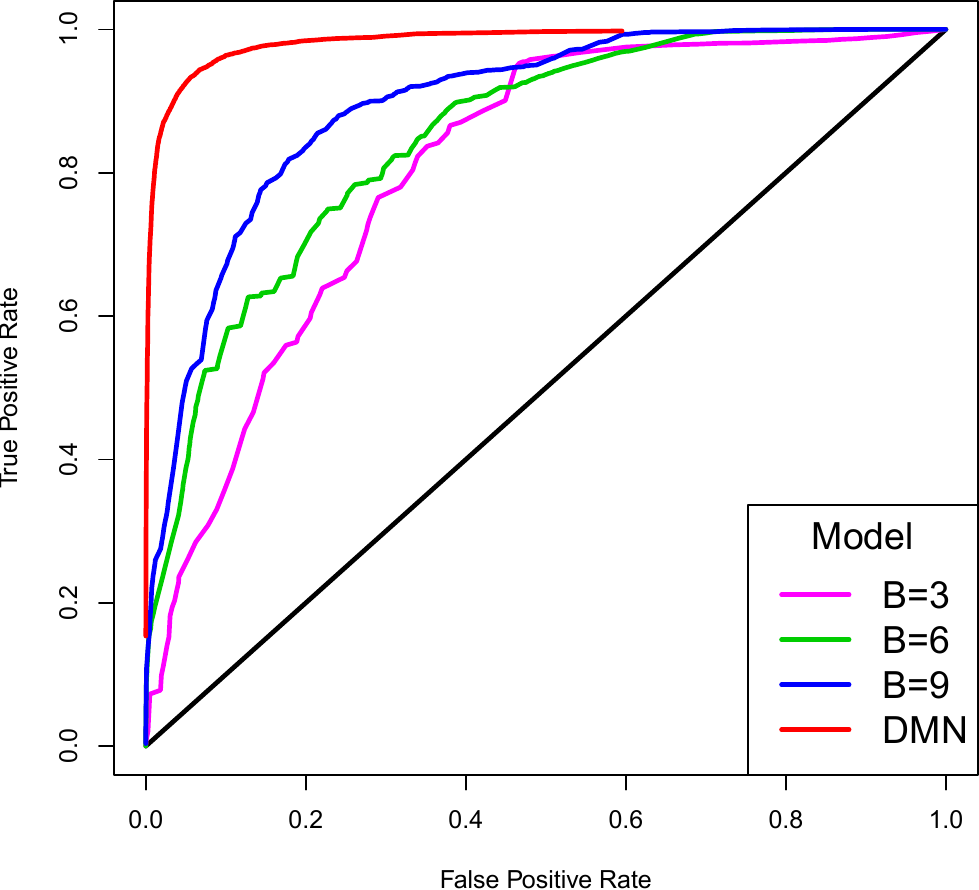} &   \includegraphics[trim={20 0 20 -10},width=.38\linewidth]{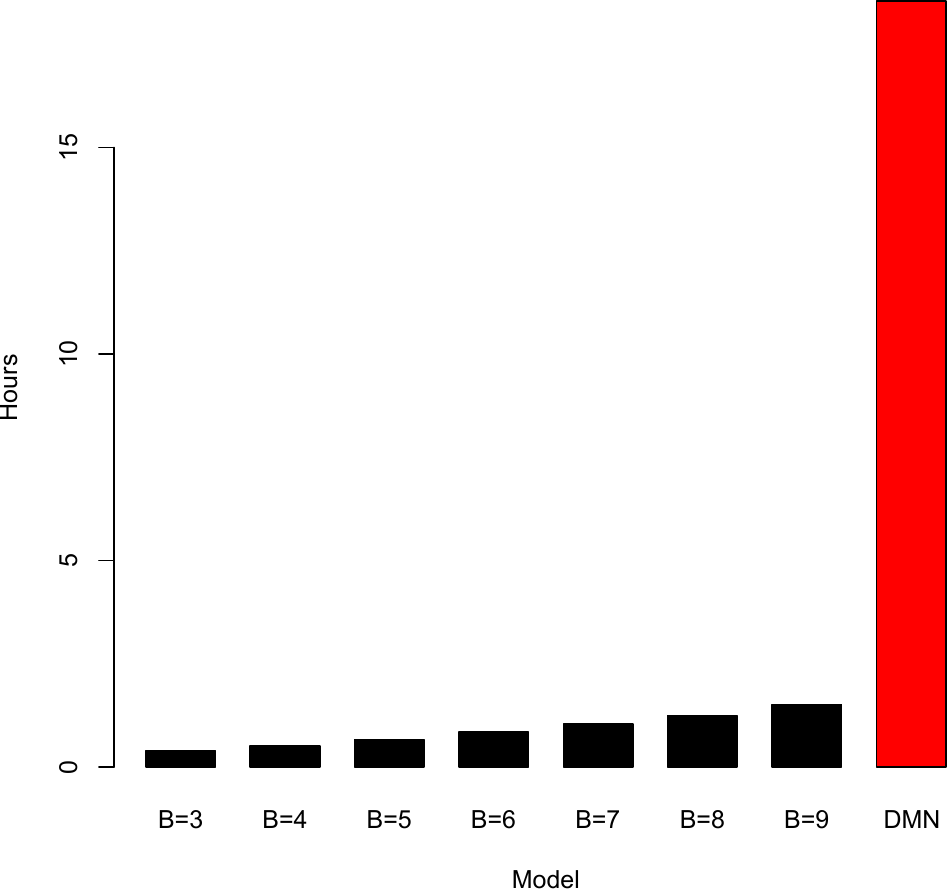} \\ \includegraphics[trim={20 0 20 -10},width=.4\linewidth]{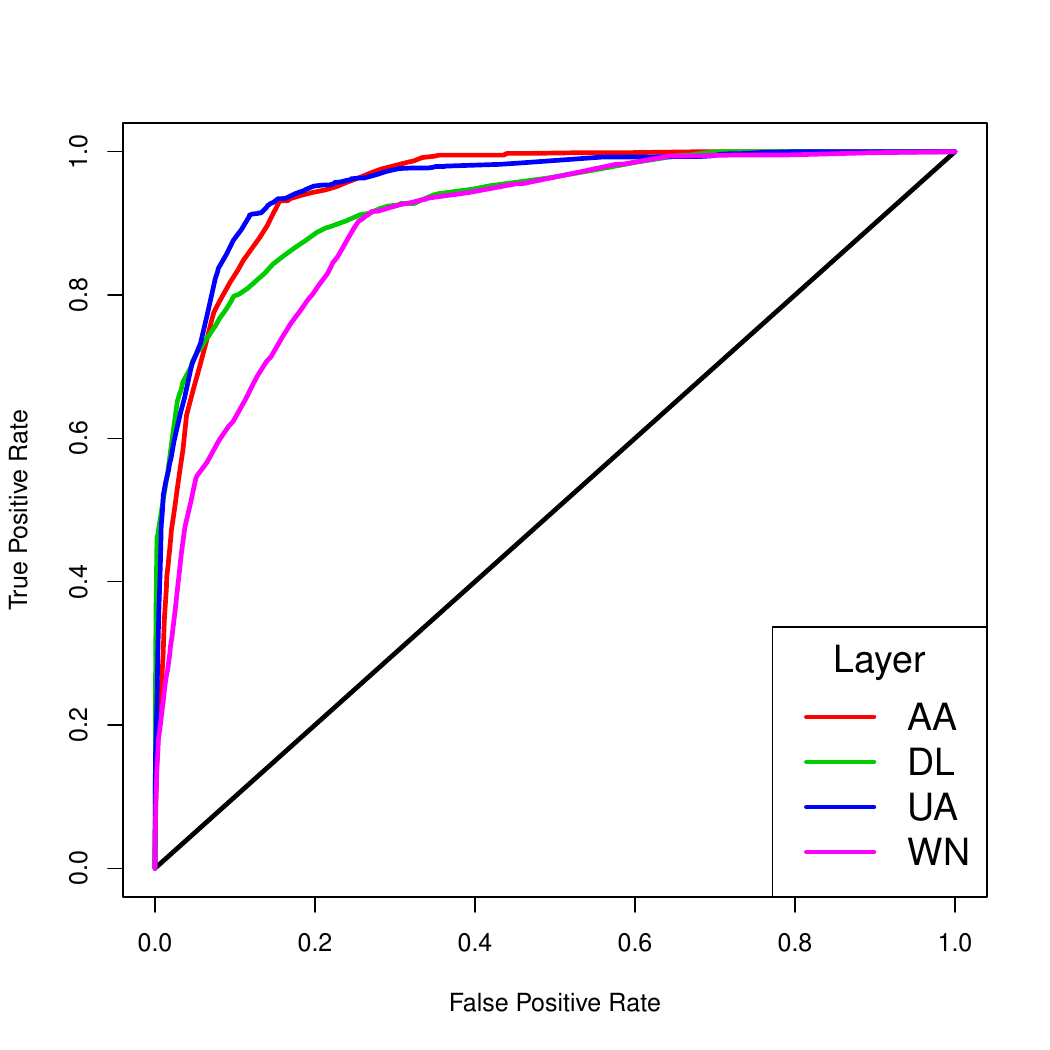} & \includegraphics[trim={20 0 20 -10},width=.4\linewidth]{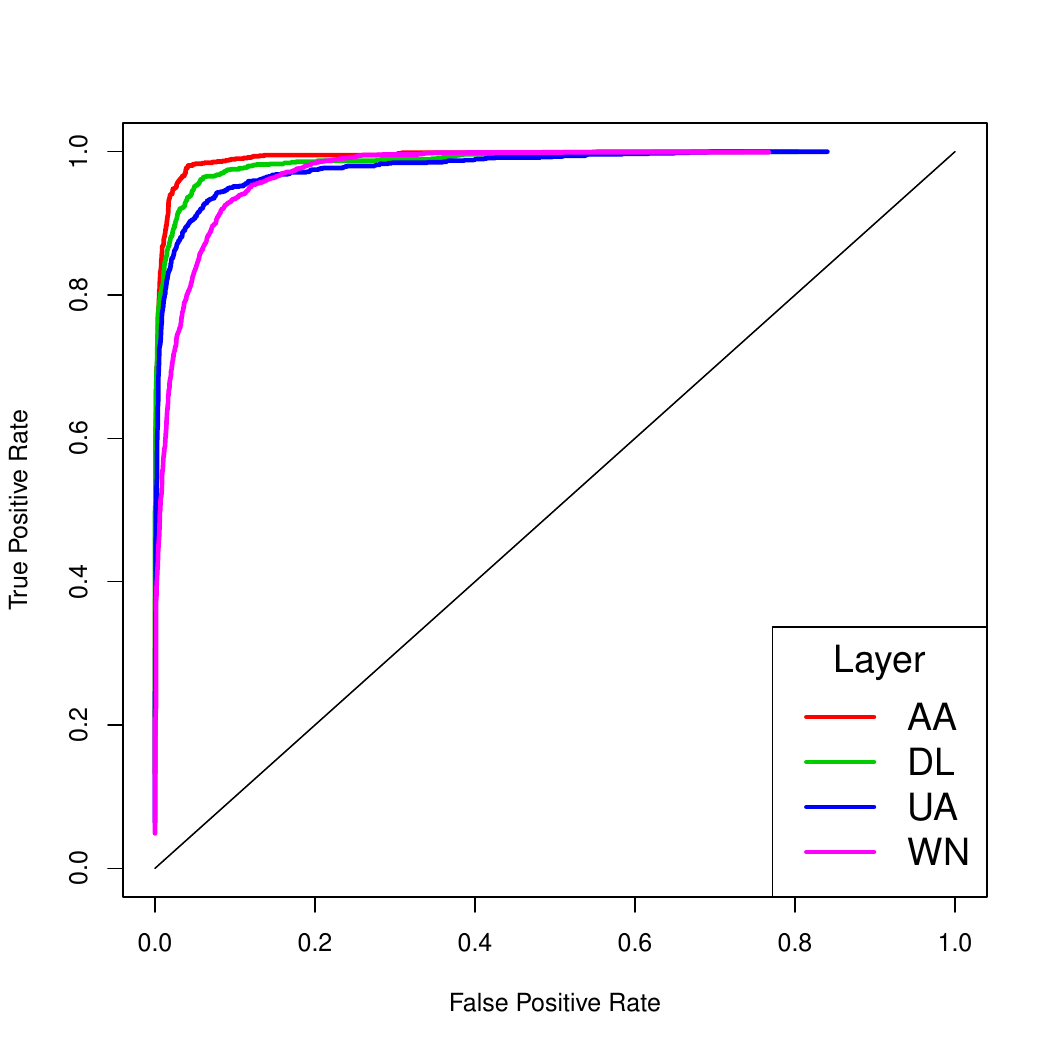}
	\end{tabular} 
	\caption{Top: ROC curves (left) and estimation times (right) for the proposed model with increasing number of blocks and the DMN  \citep{durante2017bayesian}. Bottom: layer-wise ROC curves from the proposed model with $B=9$ (left) and the DMN (right).}
	\label{fig:cs_roc}
\end{figure}

First we compare the performance of the DMN with the DMBN model. For both
models we choose a $R=H=2$, and a very smooth progression over time
$l_{\mu}=l_{\mu_p}=l_{\bar{x}}=l_{x}=5\times 10^{-5}$. We use the
first nine years of the sample (36 quarters) for training the model,
i.e. $t=\{t_1,\ldots,t_{36}\}$, and the last year for out-of-sample
testing. Both models were run for 5,000 MCMC iterations and 20\% burn-in,
with a random-scan for the DMBN. Computing the posterior predictive
distribution for the edge probabilities in the test sample $t^*=\{t_{37},\ldots,t_{40}\}$
is straightforward within the current Gibbs sampling framework, see
Step 11 in \ref{sec:gibbssampler}. Figure \ref{fig:cs_roc} (top-left) shows the ROC curves
over the test data for the DMBN with $B=\{3,6,9\}$ blocks and the
DMN. The DMN turns out to be a very accurate classifier,
but is computationally very costly as it estimates $N (N-1)/2K T=505,600$
logits, in contrast to the $B (B+1)/2 K T=14,400$ from the DMBN with
$B=9$ blocks. The performance of the DMBN model increases with $B$
and takes substantially less time to estimate; estimation times (Figure
\ref{fig:cs_roc}, top-right) for the DMBN range from 25 minutes ($B=3$) to 1.5
hours ($B=9$), and are at least one order of magnitude faster than
the DMN, which needs almost 19 hours to be estimated.
The layer-wise ROC curves presented below for the DMBN with $B=9$
blocks (left), and the DMN (right) also show how the least
structured airline network (Southwest) is the most difficult to predict. 

\begin{figure}
	\centering 
	\begin{tabular}{cccc}   
		\includegraphics[trim={60 65 60 65},width=.22\linewidth]{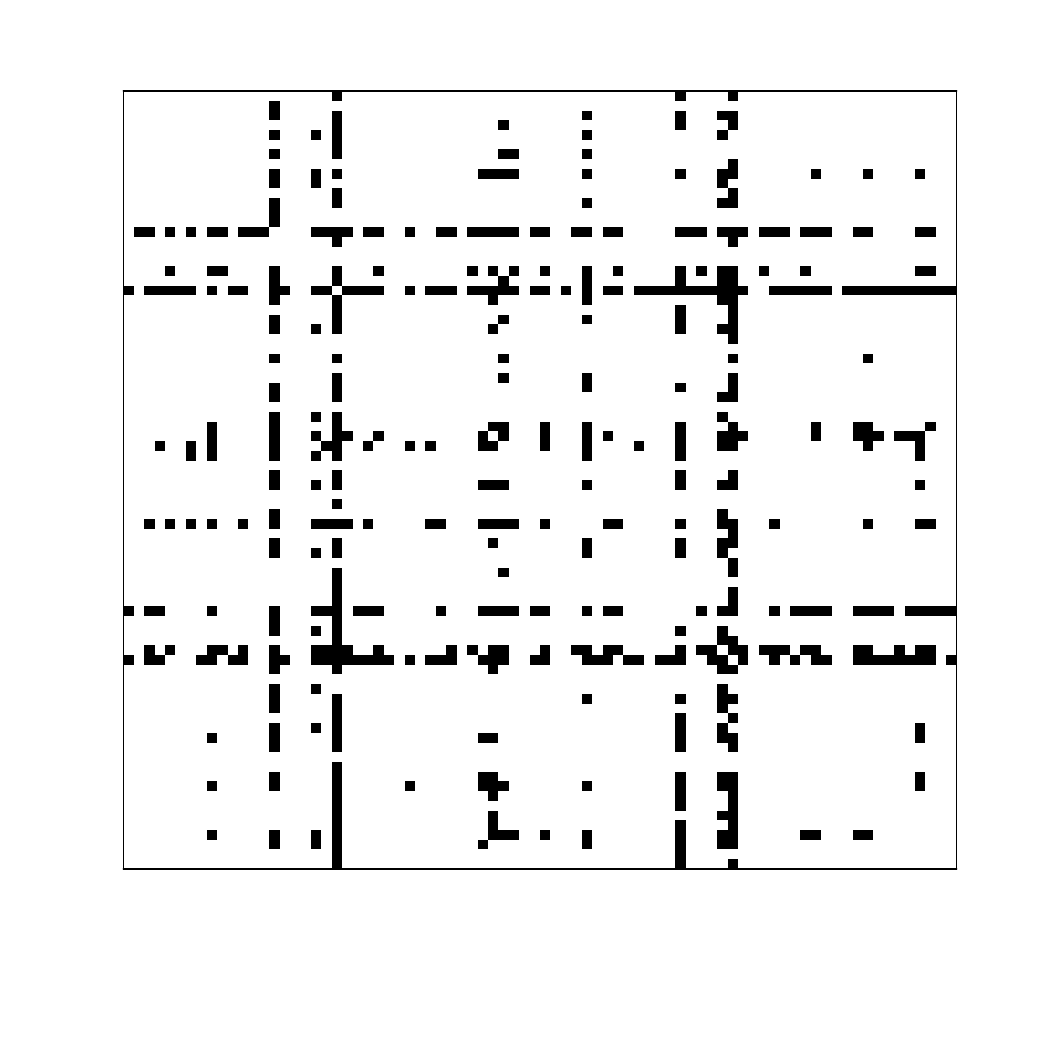} &   \includegraphics[trim={60 65 60 65},width=.22\linewidth]{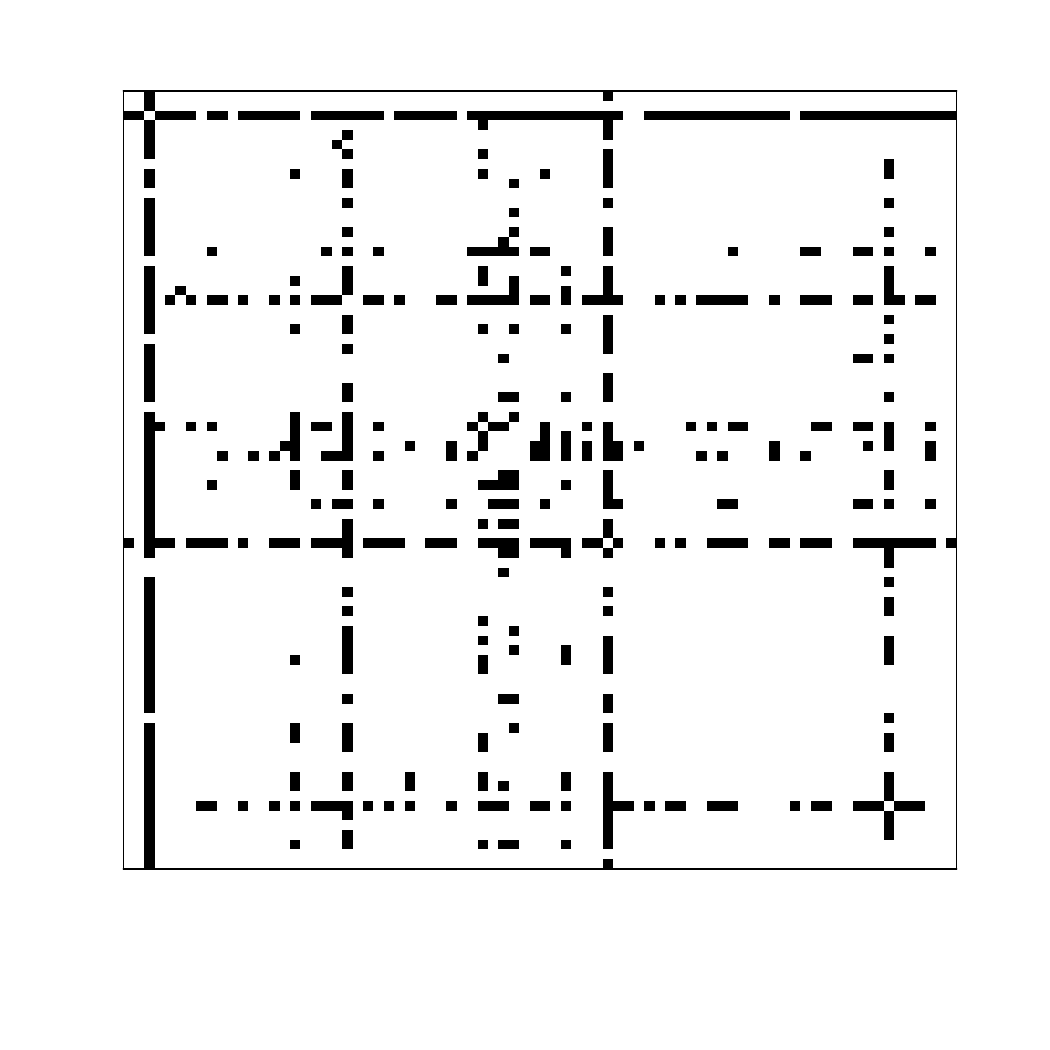} & \includegraphics[trim={60 65 60 65},width=.22\linewidth]{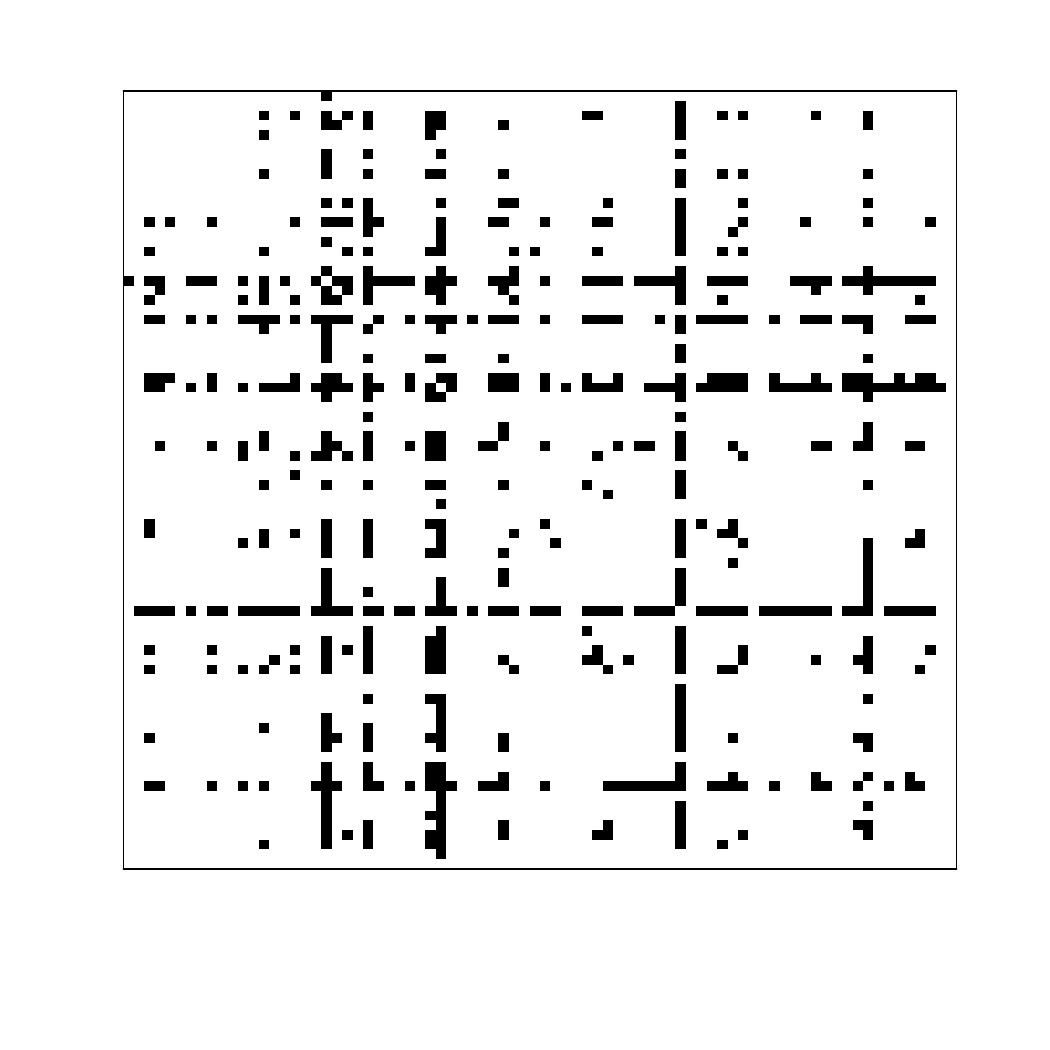} &  \includegraphics[trim={60 65 60 65},width=.22\linewidth]{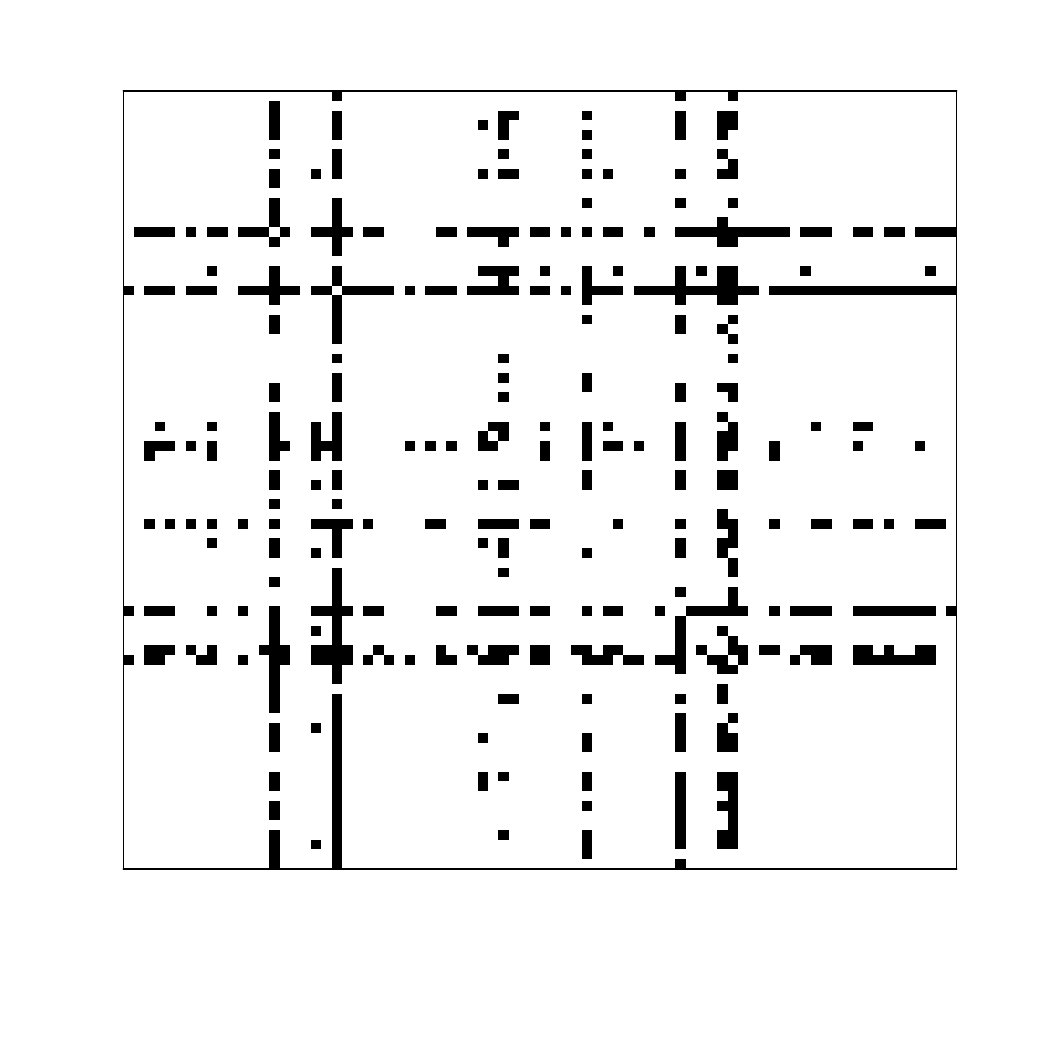} 
		\\
		\includegraphics[trim={60 65 60 65},width=.22\linewidth]{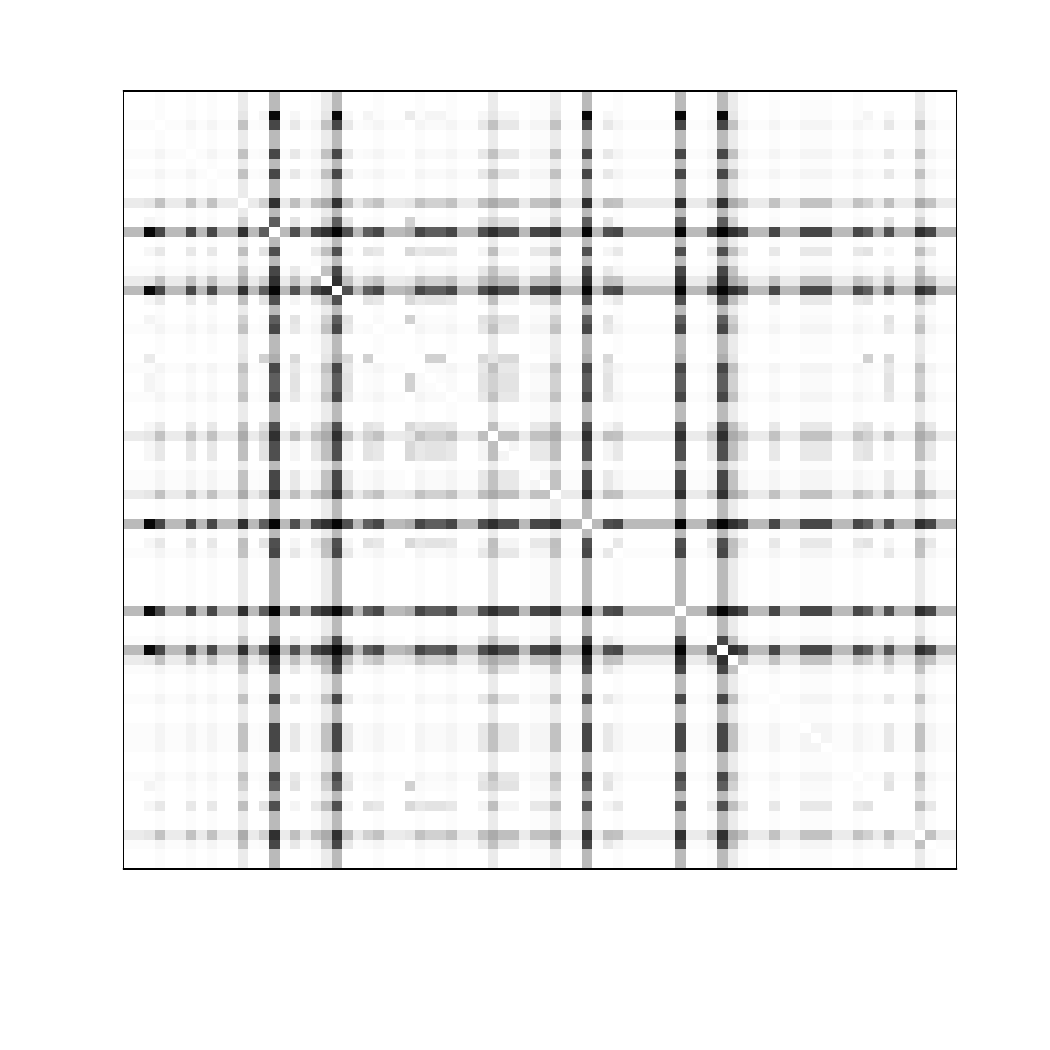} &   \includegraphics[trim={60 65 60 65},width=.22\linewidth]{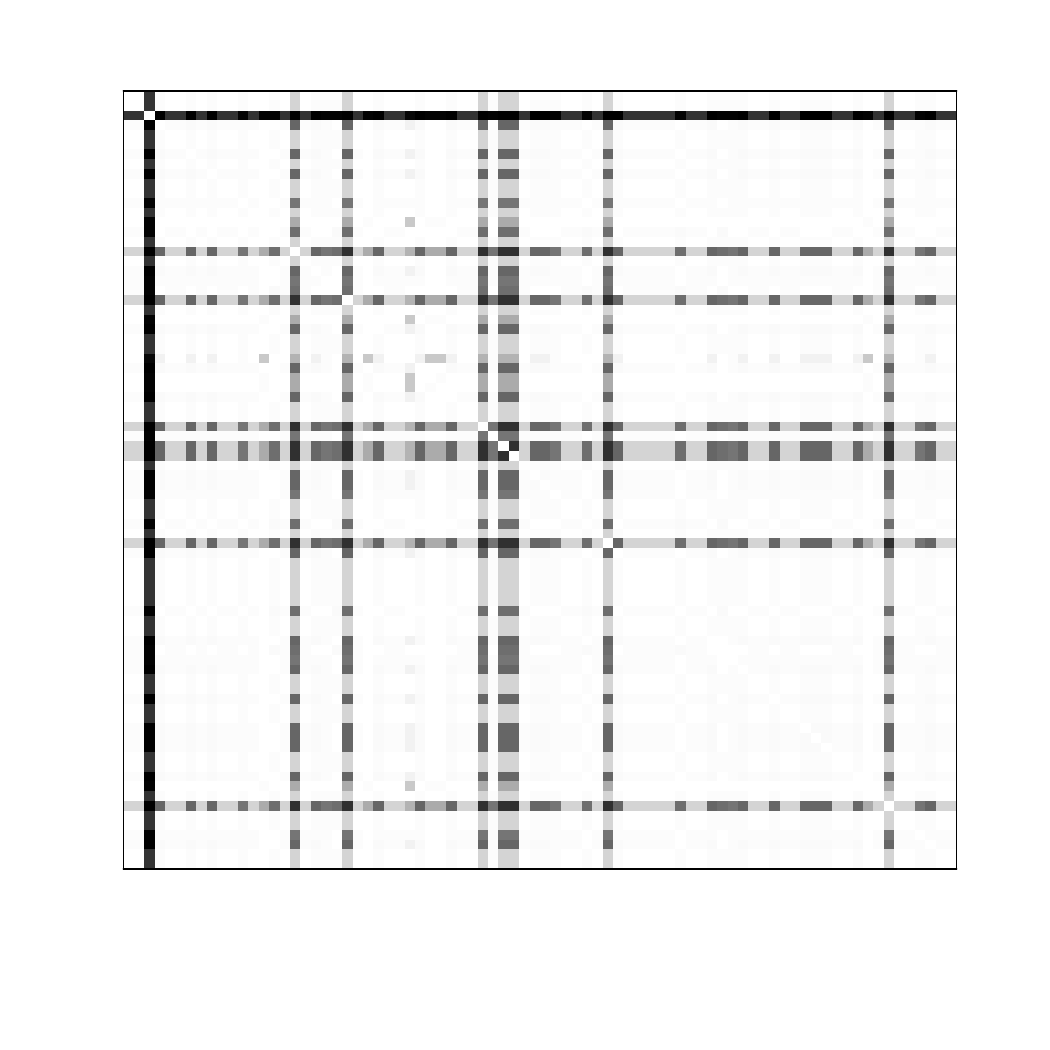} & \includegraphics[trim={60 65 60 65},width=.22\linewidth]{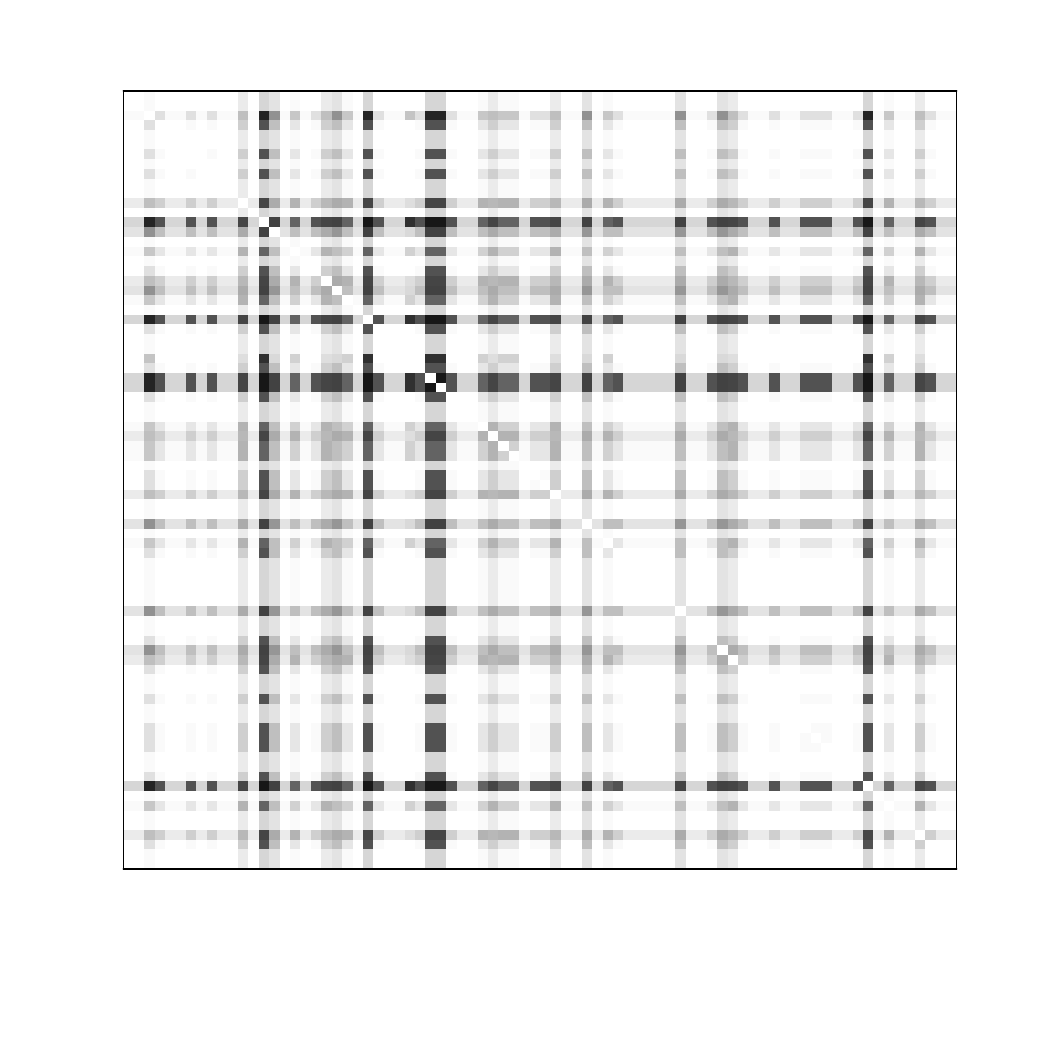} &  \includegraphics[trim={60 65 60 65},width=.22\linewidth]{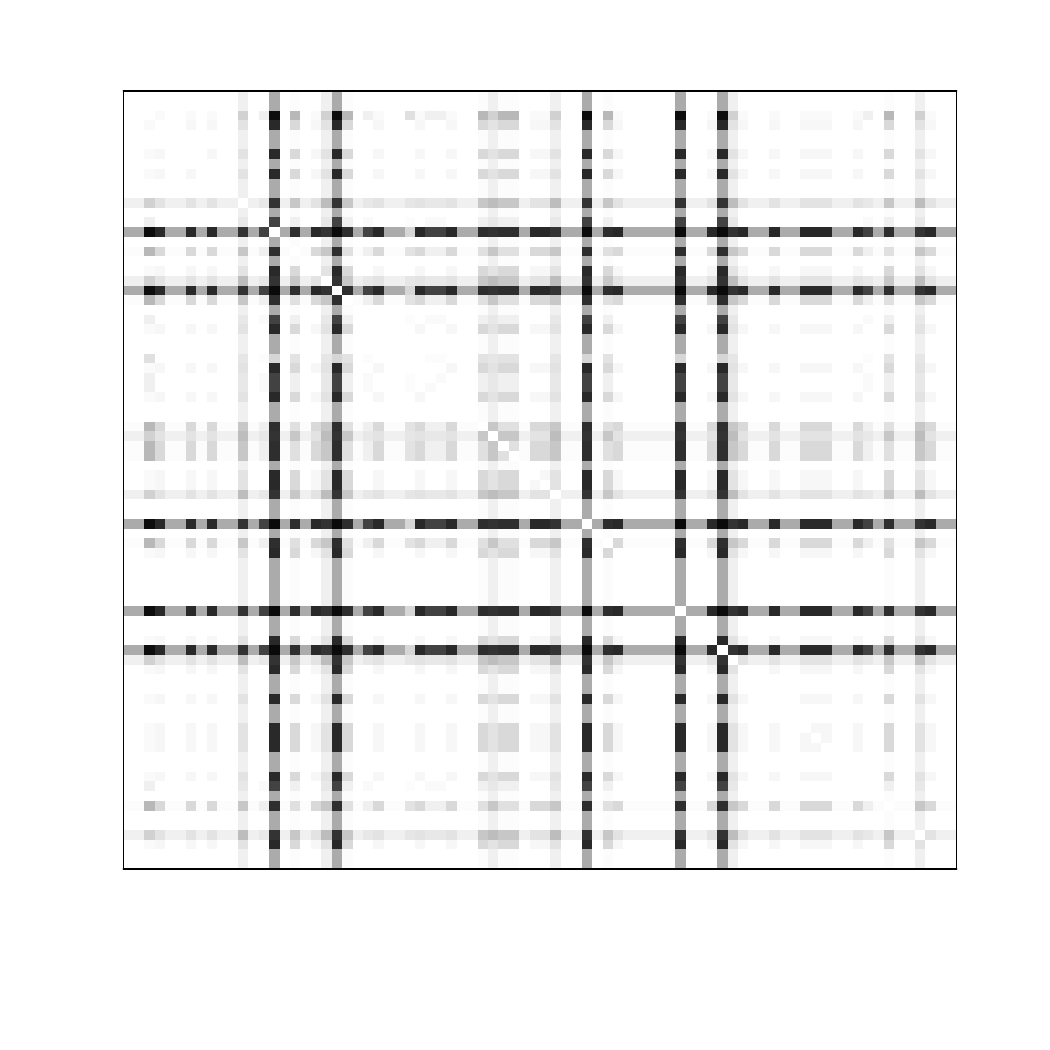}
		\\2009Q4 - AA&2012Q3 - DL&2015Q2 - UA&2018Q1 - AA
	\end{tabular} 
	\caption{Observed adjacency matrices (top row) and estimated edge probabilities (bottom row) on selected graphs from the multilayer airline network. The first three columns present estimates from graphs within the training set, whereas the last column show the one-quarter-ahead estimated probabilities.}
	\label{fig:cs_fitted_matrices}
\end{figure}

Figure \ref{fig:cs_fitted_matrices} presents some adjacency matrices from the multilayer network,
and their estimated edge probabilities calculated from the posterior
samples of the DMBN model with $B=9$ blocks. All matrices are $80\times 80$
in size, and their rows are ordered according to the airport's IATA
codes (see \ref{sec:iatacodes}). Most adjacency matrices (top row) present
a clear hub-and-spoke layout, with few dominant airports connected
to all other nodes at a given layer, which is captured well by the
model through the estimated probabilities (bottom row). Note that
probability matrices in Figure \ref{fig:cs_fitted_matrices} (bottom row) are not ordered according
to blocks but to airport codes, thus the block structure resulting
from the DMBN model is not clearly visible. The first column shows
the network of American Airlines at the last quarter of 2009, from
which we see how the model learned the connectivity patterns from
the adjacency matrix, assigning the highest probabilities to edges
connected to Charlotte, Dallas Fort-Worth, Miami, Chicago O'Hare and
Philadelphia. Having the latent block coordinates $\bar{x}_{z_i}(t)$
entering as a bilinear form \citep{hoff2005bilinear} is convenient
here to capture the cross-like patterns exhibited by these hubs, as
airports with larger magnitudes in their latent space will increase
their connectivity with respect to every other node in the network,
regardless of the block they belong to. The estimated edge probabilities
for Delta in the third quarter of 2009 (second column) seem to have
captured the most relevant patterns, corresponding to the connections
of Delta's major hubs, i.e. Atlanta, Detroit, Los Angeles, La Guardia,
Minneapolis and Salt Lake City. Similar results are obtained for United/Continental
in 2015 (third column) with Newark, Dulles, Houston International
and San Francisco, although some more noise is appreciated in the
estimated probabilities. The last column of Figure \ref{fig:cs_fitted_matrices} shows the one-quarter-ahead
predicted edge probabilities in the test set for American Airlines.\newline

\begin{figure}[h]
	\centering 
	\begin{tabular}{cc}   
		\includegraphics[trim={20 30 20 30},width=.43\linewidth]{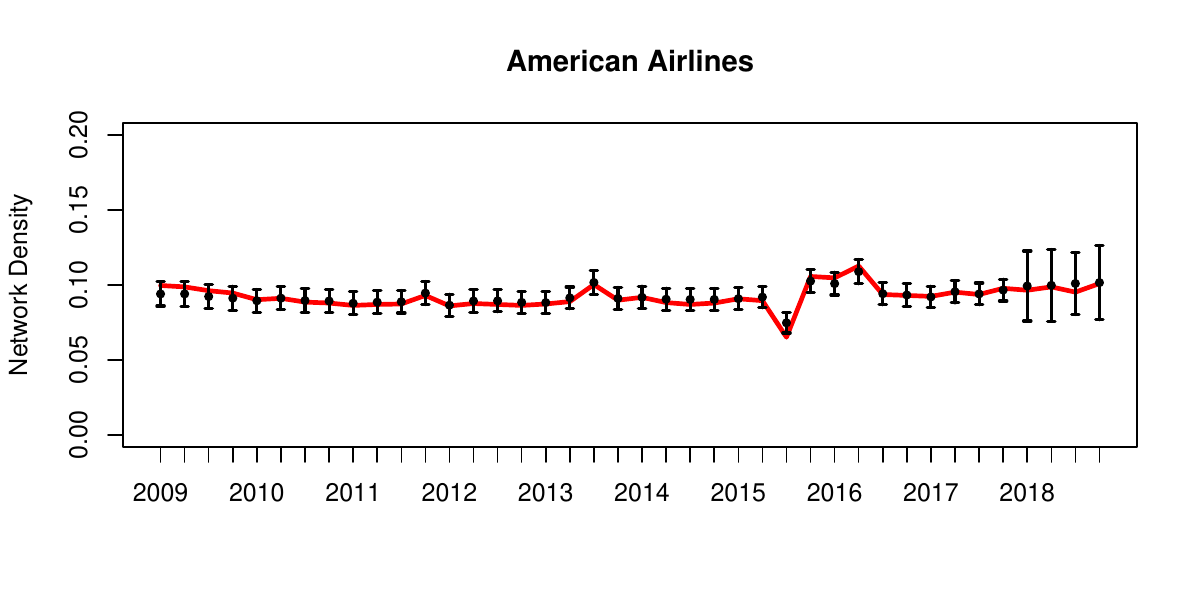} &   \includegraphics[trim={20 30 20 30},width=.43\linewidth]{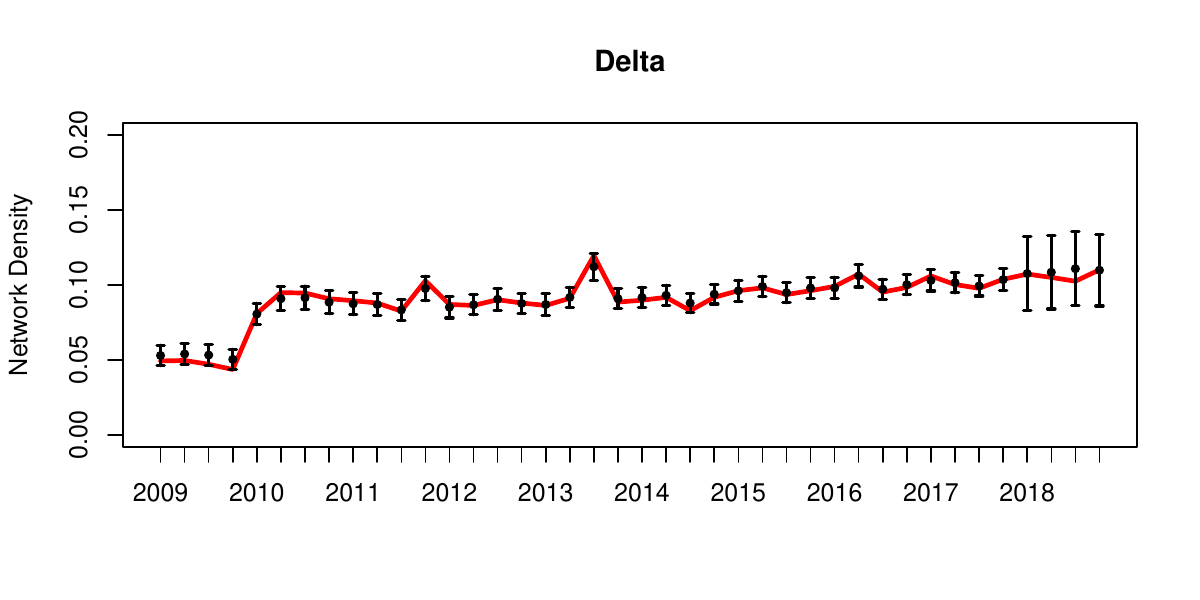}
		\\
		\includegraphics[trim={20 30 20 30},width=.43\linewidth]{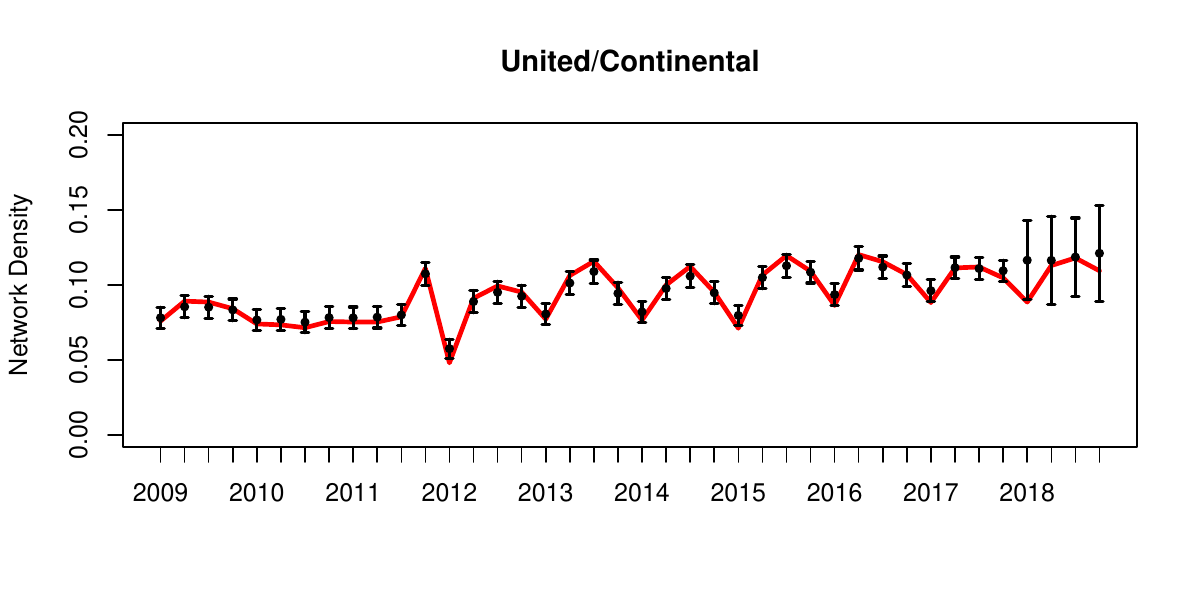} &   \includegraphics[trim={20 30 20 30},width=.43\linewidth]{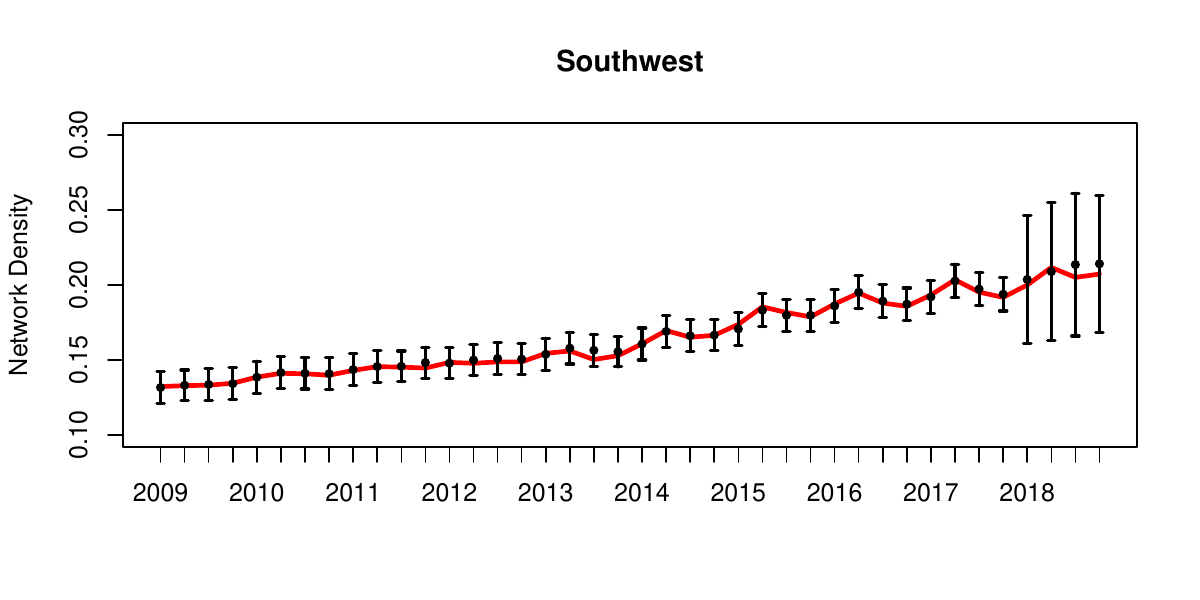} 
	\end{tabular} 
	\caption{Estimated (black) and observed (red) network densities.}
	\label{fig:cs_densities}
\end{figure}

Airline densities and airport degrees can be readily calculated from
the posterior edge probabilities (Eq. \ref{eq:Density},\ref{eq:Degree}), both for in-sample
and out-of-sample predictions. Figure \ref{fig:cs_densities} presents observed and estimated
network densities for all airlines during the sample period, which
range from 0.05 to 0.2. In all cases the estimated densities fit almost
perfectly the observed data, and forecasts lie within the 95\% posterior
intervals. Note that the density forecasts are not a mere projection
from a univariate time-series, as Figure \ref{fig:cs_densities} may suggest: here the
entire multilayer graph has been projected forward in time using the
smoothness from the Gaussian processes. Future densities and degree
distributions are then calculated from the predicted multilayer networks.
American Airlines appears as the airline with the most stable network
density, in a similar manner as Delta after it absorbed Northwest
in 2010. United/Continental shows a marked seasonal effect after the
merge of their former airlines in 2012, perhaps due to network restructuring.
Southwest is the only airline that grows steadily in density during
the sample period, with noticeable seasonality after 2013. 

\begin{subequations}\begin{align}
	D^k(t)&=\Bbb{E}\left[\sum_{i=2}^N\sum_{j=1}^{i-1}A_{ij}^k(t)/(N(N-1)/2)\right]=\sum_{i=2}^N\sum_{j=1}^{i-1}\pi_{z_iz_j}^k(t)/(N(N-1)/2)\label{eq:Density}\\
	d_i^k(t)&=\Bbb{E}\left[\sum_{j\ne i}A_{ij}^k(t)\right]=\sum_{j\ne i}\pi_{z_iz_j}^k(t)\label{eq:Degree}
	\end{align}\end{subequations}

\begin{figure}[h]
	\centering 
	\begin{tabular}{cc}   
		\includegraphics[trim={20 30 20 30},width=.43\linewidth]{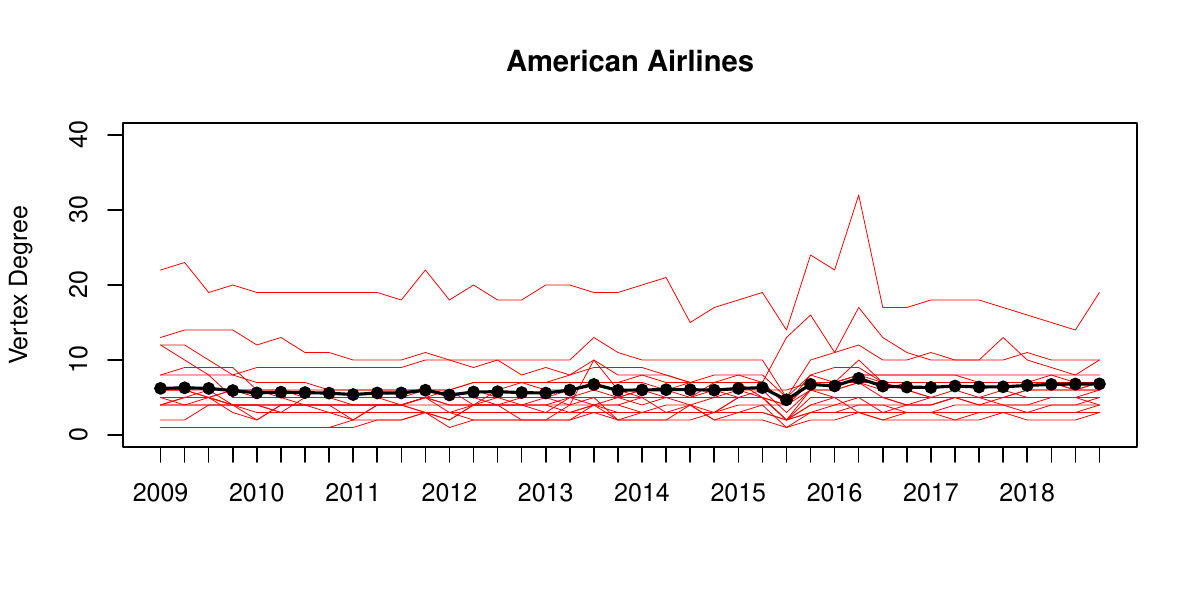} &   \includegraphics[trim={20 30 20 30},width=.43\linewidth]{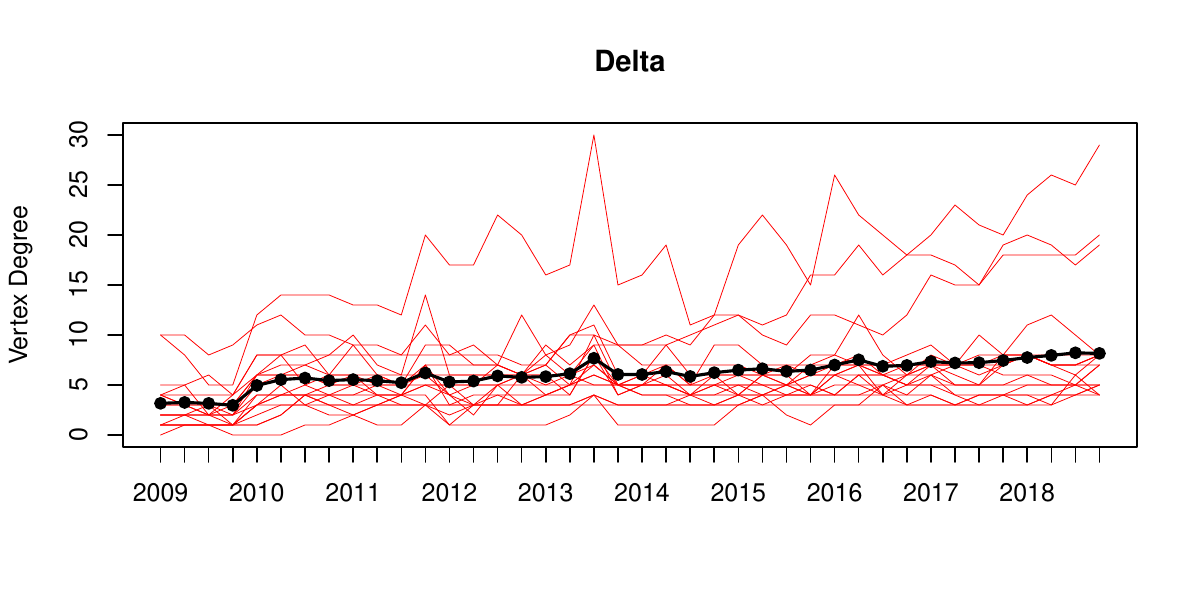}  
		\\
		\includegraphics[trim={20 30 20 30},width=.43\linewidth]{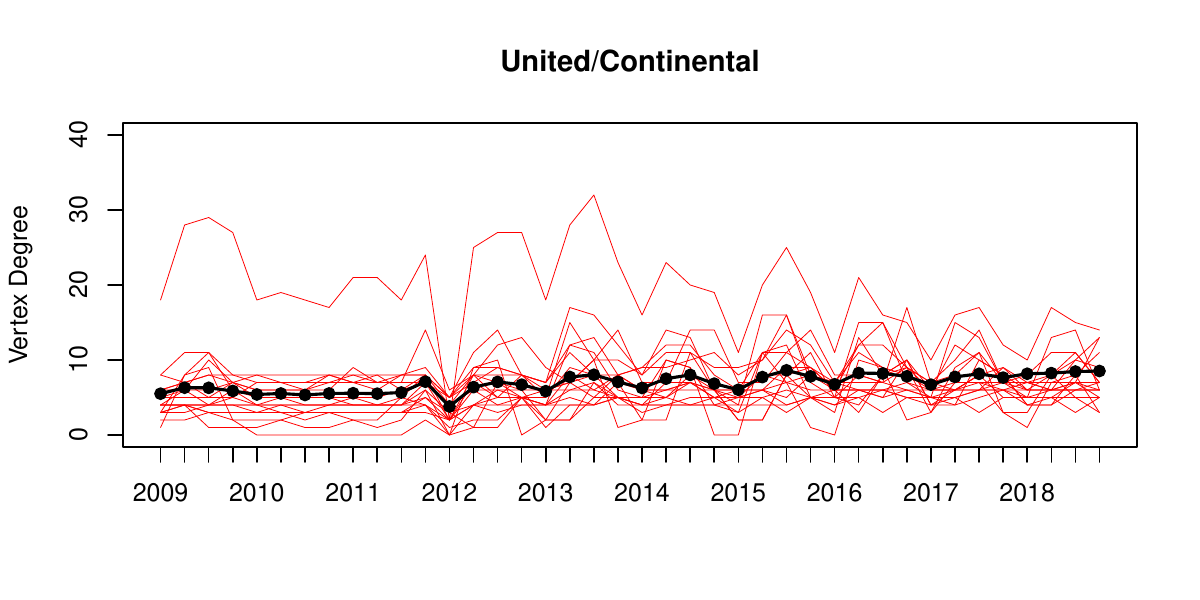} &   \includegraphics[trim={20 30 20 30},width=.43\linewidth]{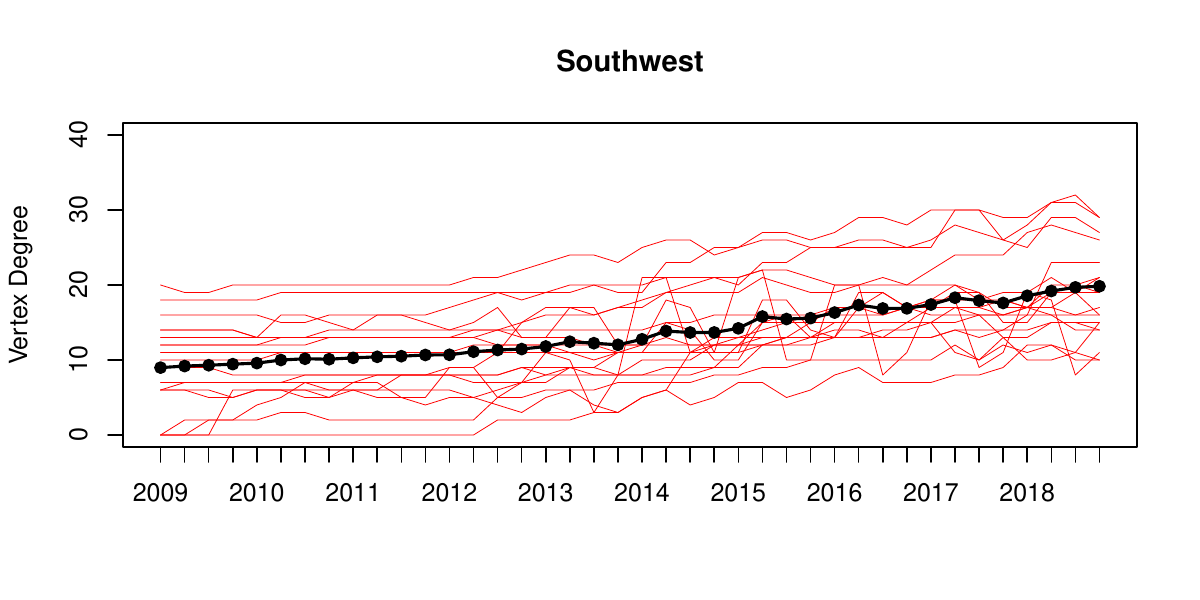} 
	\end{tabular} 
	\caption{Observed (red) and estimated (black) vertex degrees for block 2.}
	\label{fig:cs_vertexb2}
\end{figure}

\begin{figure}[h]
	\centering 
	\begin{tabular}{cc}   
		\includegraphics[trim={20 30 20 30},width=.43\linewidth]{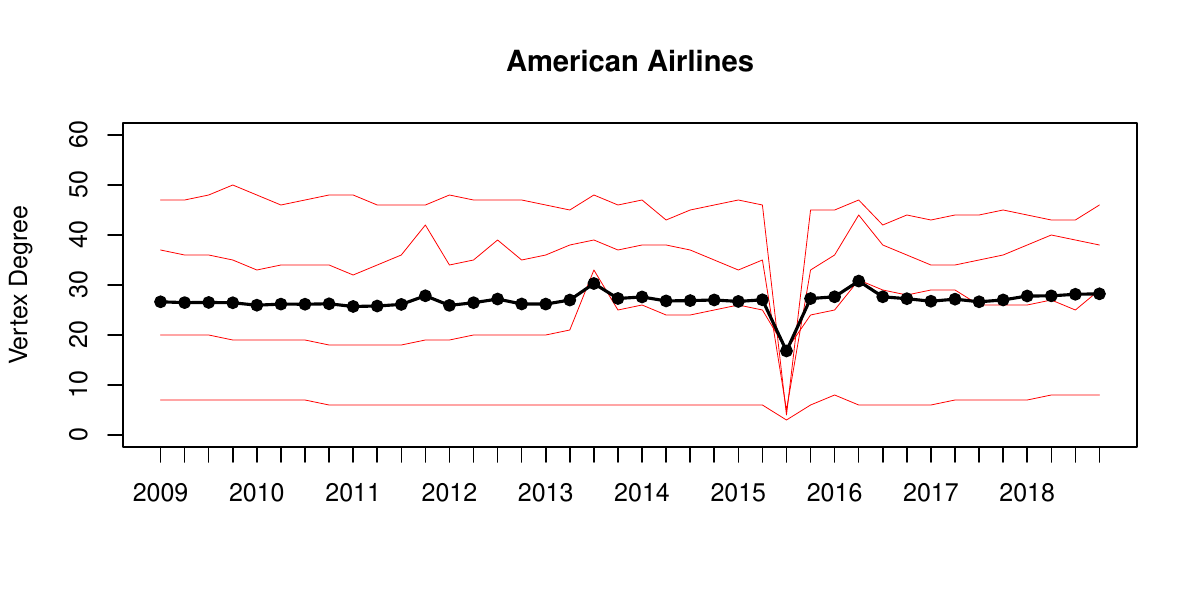} &   \includegraphics[trim={20 30 20 30},width=.43\linewidth]{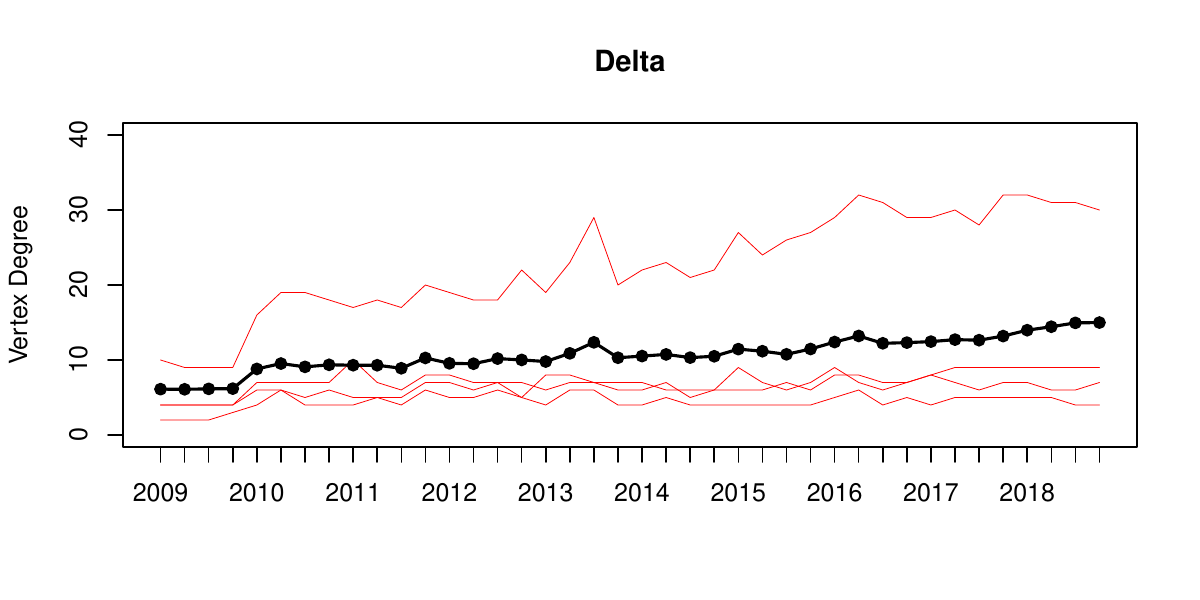}  
		\\
		\includegraphics[trim={20 30 20 30},width=.43\linewidth]{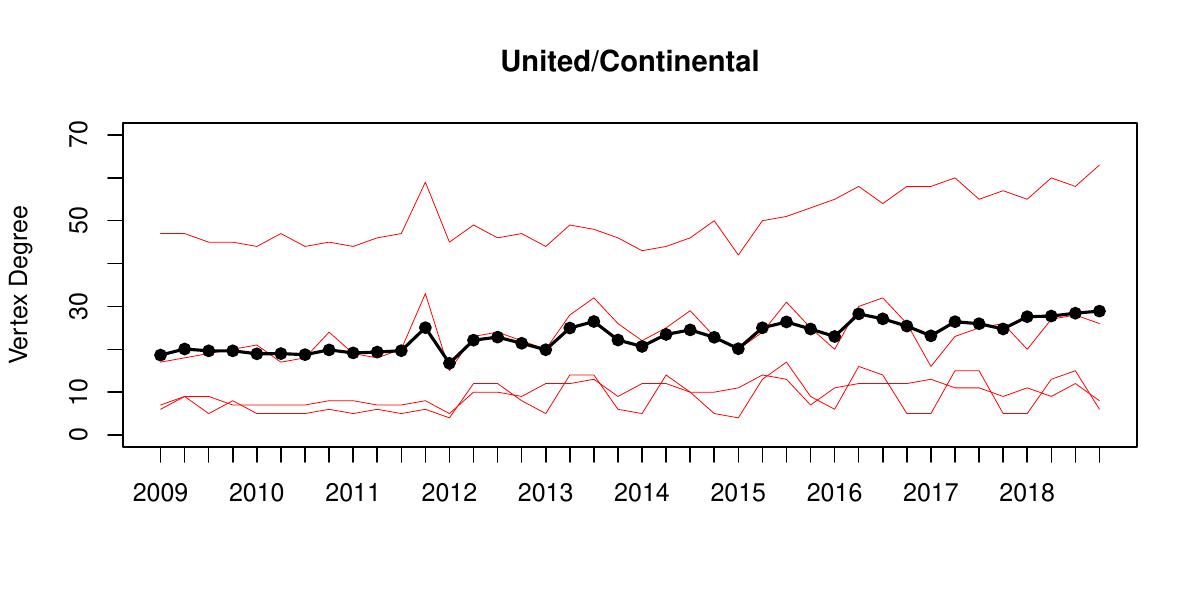} &   \includegraphics[trim={20 30 20 30},width=.43\linewidth]{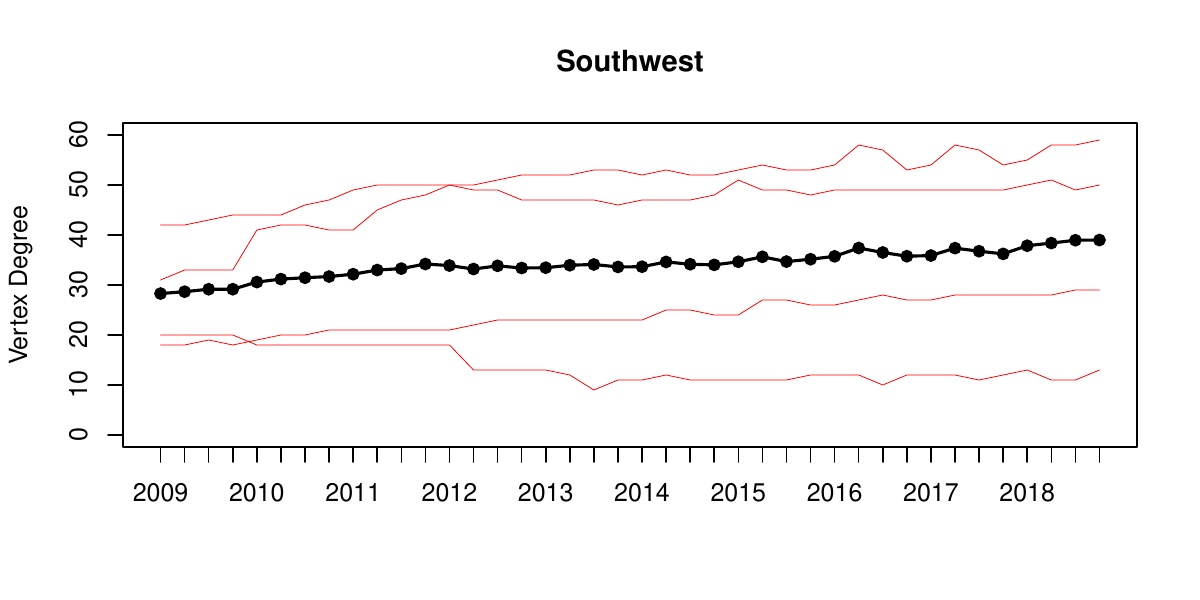} 
	\end{tabular} 
	\caption{Observed (red) and estimated (black) vertex degrees for block 8.}
	\label{fig:cs_vertexb8}
\end{figure}

Table \ref{tab:cs_clusters} lists the nine airport clusters found by the DMBN model. The
clustering structure becomes apparent after computing the matrix of
posterior probabilities that two nodes are in the same block, which
is invariant with respect to the block labels. Clusters have been
formed both layer-wise and also according to the connectivity dynamics
of each airport. The first cluster is the largest, and contains 33
airports that are mainly small and mid-sized Southwest airports, with
the exception of Honolulu, Cleveland and Memphis. Cluster 2 aggregates
airports with a rising number of Southwest connections, the outliers
here would be Boston, NY La Guardia, Seattle and Washington Reagan.
The third, fourth and sixth clusters represent the bulk of the major
hubs from American Airlines, Delta, and Southwest respectively, whereas
Chicago Midway stands alone in cluster 5 as the largest focus city
for Southwest, with a 96\% of market share \citep{BTS2019b}. The
three airports in cluster 7 are fast-growing Southwest bases, with
Dallas Love Field and Houston Hobby among the fastest growing airports
in the US in the last decade. Cluster 8 groups a number of large hubs
with little presence of Delta, with the exception of Los Angeles,
and cluster 9 features large United/Continental hubs. Figures \ref{fig:cs_vertexb2} and \ref{fig:cs_vertexb8} present the observed and estimated degrees (Eq.
\ref{eq:Degree}) for two clusters to assess the effect of the stochastic blockmodeling
on learning the dynamics of the multilayered network. Note how the
estimated Gaussian process for the edge probabilities in each block
captures the average dynamics from all airports belonging to that
block. 

\begin{table}[h]
	\centering 
	\begin{tabular}{cc}
		\hline 
		Cluster  & Airports\tabularnewline
		\hline 
		$\hat{p}=1$ & ABQ, ALB, BDL, BHM, BOI, BUF, BUR, CHS, CMH, CVG, ELP\tabularnewline
		& GEG, GRR, HNL, ISP, JAX, LIT, MEM, MHT, OGG, OKC, OMA\tabularnewline
		& ONT, ORF, PBI, PNS, PVD, RIC, RNO, SAV, SDF, TUL, TUS\tabularnewline
		\hline 
		$\hat{p}=2$ & AUS, BOS, CLE, DCA, IND, LGA, MCI, MKE, MSY, PDX\tabularnewline
		& PIT, RDU, RSW, SAN, SAT, SEA, SJC, SMF, SNA\tabularnewline
		\hline 
		$\hat{p}=3$ & CLT, DFW, JFK, MIA\tabularnewline
		\hline 
		$\hat{p}=4$ & ATL, DTW, MSP, SLC\tabularnewline
		\hline 
		$\hat{p}=5$ & MDW\tabularnewline
		\hline 
		$\hat{p}=6$ & BNA, BWI, FLL, LAS, MCO, STL, TPA\tabularnewline
		\hline 
		$\hat{p}=7$ & DAL, HOU, OAK\tabularnewline
		\hline 
		$\hat{p}=8$ & DEN, LAX, PHL, PHX\tabularnewline
		\hline 
		$\hat{p}=9$ & EWR, IAD, IAH, ORD, SFO\tabularnewline
		\hline 
	\end{tabular}\caption{Estimated airport clusters. IATA airport codes in Appendix D. }
	\label{tab:cs_clusters}
\end{table}

\begin{figure}[h]
	\centering\label{fig:auc}
	\includegraphics[height=8cm]{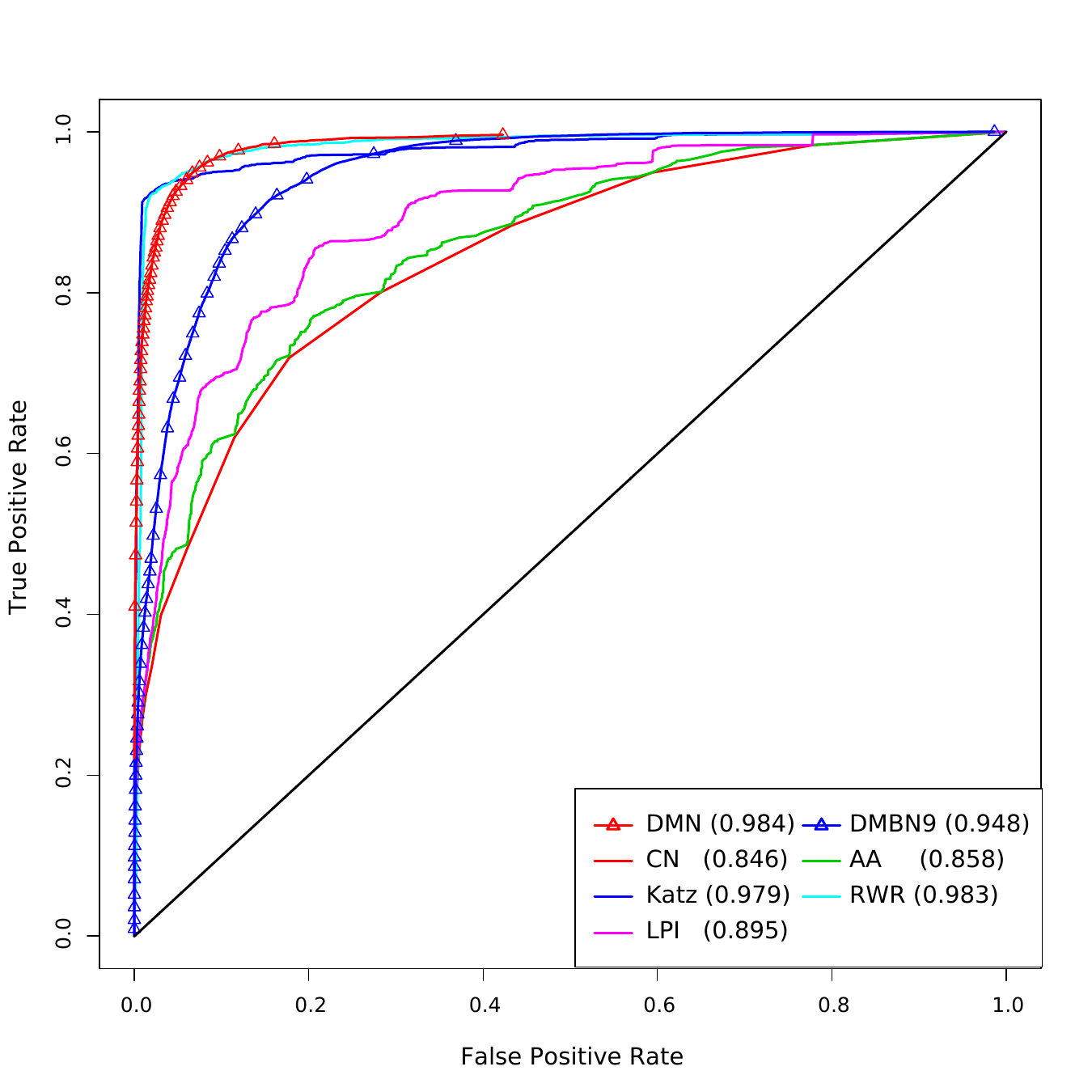}
	\caption{ROC curves for probabilistic and similarity-based link prediction methods. The numbers in parenthesis are the areas under the curve (AUC).}
	\label{fig:cs_bench_auc}
\end{figure}

We finally benchmark the DMN and DMBN forecasting performance against several popular, non-probabilistic, algorithms. Link prediction is a common task in network science, where a large proportion of methods are based on similarity metrics \citep{lu2011link,martinez2016survey}. Similarity-based algorithms define a function, usually based on the topology of the graph, to assign a similarity score between every pair of nodes in the network. These scores can be used as input for a binary classifier that predicts the existence or non-existence of links between nodes. We choose five similarity-based algorithms: i) Common Neighbors (CN)-\citep{newman2001clustering}, ii) Adamic-Adar Index (AA)-\citep{adamic2003friends}, iii) Katz Index (Katz)-\citep{katz1953new}, iv) Random walk with restart (RWR), based on PageRank algorithm \citep{brin1998anatomy}, and v) Local Path Index (LPI)-\citep{zhou2009predicting}. The first two indexes use the local topology to calculate scores whereas Katz and RWR are global methods that use all available information in the graph; LPI is based on a mixed strategy. These algorithms can be readily implemented in e.g. R for link prediction \citep{kolaczyk2014statistical,Bojanowski2018link}. The selected methods are originally defined for single-layered non-dynamic graphs. We will  therefore apply the algorithms layerwise, but extending them to a dynamic setting, as follows. Let $\mathrm{score}_{ij}^k(t)$ be the similarity score between nodes $i,j$ at layer $k$ and time $t$. A straightforward way to account for time dynamics is to use a simple exponential smoother,
\begin{align}
S_{ij}^k(t)&=\alpha\ \mathrm{score}_{ij}^k(t)+(1-\alpha)\ S_{ij}^k(t-1),\ t>1,\\
S_{ij}^k(1)&={\mathrm{score}}_{ij}^k(1)
\end{align}
where $\alpha\in[0,1]$ is the smoothing parameter. The recursion above can be applied on each layer until we obtain $S^k(t=t_{36})$, a matrix of scores that summarizes the information contained in the training data on that layer, and that can be used to classify/predict the links on the test graphs $A^k(t=t_{37},\ldots,t_{40})$. In Figure \ref{fig:cs_bench_auc} we present the ROC curves and the area under the curve (AUC) for the DMN and DMBN with 9 blocks (DMBN9), and for the similarity-based prediction algorithms with $\alpha=1$, i.e. a random-walk. We see that the classifiers based on the global indicators (Katz and RWR) are almost as good as the DMN. Local and quasi-local models (CN, AA, and LPI) are faster alternatives comparatively, but with a reduced predictive performance. Table \ref{tab:cs_bench_top} presents evaluation metrics for all algorithms at two different levels of $\alpha$. The classification threshold for each model has been chosen to maximize the F1 measure. Overall, the DMN and the Katz index are the best probabilistic and similarity-based classifiers, respectively. Probabilistic models yield better results in term of precision/recall and the combined F1 measure. The classification performance of similarity-based algorithms improves as their scores take more information from the recent graphs in the series due to the smoothing. Still, while some of the similarity-based methods are competitive as pure link-prediction devices, their outcome (i.e. the score matrices) can be difficult to interpret, and the insights from such models are quite limited. As we have shown in the two case studies, probabilistic multilayer dynamic models have the important advantage of offering, among other possibilities, probabilistic forecasts, temporal network analysis from inferred latent coordinates, and community detection.

\begin{table}[h]
	\centering
	\begin{tabular}{ |l|c|c|c|c| } 
		\hline
		Model ($\alpha=1$) & Precision & Recall & $F_1$ & AUC \\ 
		\hline
		DMN & (0.933) & \textbf{0.950} & \textbf{0.941} & \textbf{0.984}\\ 
		DMBN - 9 Blocks & 0.855 & (0.914) & 0.884 & 0.948 \\ 
		Common Neighbors & 0.441 & 0.620 & 0.515 & 0.846 \\ 
		Adamic-Adar & 0.526 & 0.590 & 0.557 & 0.858 \\ 
		Katz Index & \textbf{0.936} & 0.913 & (0.924) & 0.979 \\ 
		Random Walk Restart & 0.908 & 0.905 & 0.907 & (0.983) \\
		Local Path Index & 0.566 & 0.678 & 0.617 & 0.895 \\  
		\hline
		\hline
		Model ($\alpha=0.4$) & Precision & Recall & $F_1$ & AUC \\ 
		\hline
		DMN & (0.933) & \textbf{0.950} & \textbf{0.941} & 0.984 \\ 
		DMBN - 9 Blocks & 0.855 & 0.914 & 0.884 & 0.948 \\ 
		Common Neighbors & 0.511 & 0.577 & 0.542 & 0.855 \\ 
		Adamic-Adar & 0.519 & 0.606 & 0.559 & 0.863 \\ 
		Katz Index & \textbf{0.945} & 0.915 & (0.930) & \textbf{0.992} \\ 
		Random Walk Restart & 0.895 & (0.922) & 0.908 & (0.988) \\
		Local Path Index & 0.564 & 0.673 & 0.613 & 0.897 \\  
		\hline			
	\end{tabular}
	\caption{Classification metrics for probabilistic and similarity-based link prediction methods, with $\alpha$ as temporal smoothing parameter. Best model is indicated in bold font, and runner-up in parenthesis.}	
	\label{tab:cs_bench_top}
\end{table}

\subsection{Some practical implications in air transportation}
The network predictions provided by the DMN and DMBN models can have a variety of applications. From a regulatory perspective, we can mention merger screening. In a post-covid setting with further concentration predicted for the airline sector \citep{budd2020european}, major consolidations in the US had often involved thousands of city-pair markets that must be screened in order to identify those where the merger would lead to an excessive increase in monopolistic power. These models can predict the system-wide evolution of market shares and route concentration post-merger to pinpoint the main routes of concern for competition authorities, who might then formulate a series of remedial divestitures to mitigate the anticompetitive threats or reject the merger altogether.  From a network planning perspective, our model can aid in the identification of new or potentially underserved routes, a process that would require the addition of demand and supply factors into the prediction. These would include, among others, price levels, catchment areas, substitute travel modes, and even the evolving nature of aircraft technology as it affects fuel efficiency and break-even load factors. 

We have obtained promising results both in terms of model expressiveness and prediction power by focusing on three full-service-carriers plus Southwest within the US market, but these may not generalize to a different case study. The application to different markets would require to carefully set up the model and experiments for that particular case. A bigger coverage of secondary cities, with the corresponding sacrifice of computational time, would be needed to properly characterize the networks of low-cost carriers in the US, with the added challenge, in terms of accuracy, of training the model to account for the operation of narrow-body aircraft by LCCs like Spirit or Frontier on relatively thinner coast-to-coast routes \citep{soyk2018revenue}. An additional level of complexity would be present if applying the model to the less concentrated European airline market, where most big carriers operate single or dual-hub strategies and there is more integration with high-speed rail, bringing both competition and collaboration opportunities to air carriers and thus playing a key role in shaping the multilayer network dynamics. 

\section{Conclusion}
\label{sec:conc}
We present dynamic multilayer network methods with potential applications to transportation networks due to their potential to model and forecast time series of complex graphs. Flexible time series analysis is obtained by modeling the probability of edges between vertices through latent Gaussian processes. The block-based extension is natural for many real networks, such as transportation networks, where community structure naturally arises, and makes it possible to substantially improve the scaling of Bayesian inference algorithms to larger networks. Specifically, the models have the potential to enhance the analysis of transportation networks due to their ability to: i) capture the dynamic, multi-layered nature of most transport networks, ii) model both endogenous and exogenous effects underlying such dynamics, iii) perform out-of-sample network forecasting, and iv) scale to reasonably large problems. The models and Bayesian inference methodology are illustrated on a sample of 10-year quarterly data from four major US airlines: American, Delta, United/Continental and Southwest.

We take advantage from the fact that important network restructuring and merges within the US air transportation system takes place during the sample period (2009-2018), and assess the ability of the models in reflect those changes. Results show how the estimated latent parameters from the models are related to the airline's connectivity dynamics, reflecting e.g. the entrance of Southwest into Atlanta after its merge with AirTran (2011), or its expansion at Dallas Love-Field in 2014. We show also how the extended model is able to capture the hub-and-spoke nature of the air transport network, and to project the entire multilayer graph into the future for out-of-sample full network forecasts, which differs from the current practice of visual analysis of static topological indicators. The stochastic blockmodeling allows for a time-series clustering of the airports' connectivity dynamics, and the identification of relevant communities, while keeping estimation times within reasonable limits. 

Several interesting extensions of the model are possible. For example,
explicit modeling of three-way dependencies, which are common in e.g.
air transportation. Extending the model to accommodate exogenous network
covariates or layer-wise stochastic blocks is straightforward. Better
methods for handling label-switching \citep{celeux2018computational}
in multi-layered networks would make it easier to interpret some aspects
of the extended model's results. For very large network problems, variational
approximations within the P{\'o}lya-Gamma framework \citep{zhou2012lognormal}
may be a good strategy to reduce estimation times. 

\section*{Acknowledgments}
This work was partially supported by the Wallenberg AI, Autonomous Systems and Software Program (WASP) funded by the Knut and Alice Wallenberg Foundation.

\setlength{\bibsep}{0.0pt}

\newpage

\begin{appendix}

\section{Normal Approximation of the P{\'o}lya-Gamma variables}
\label{sec:pgapprox}
The moment generating function (\citealp{polson2013bayesian}) is
defined as

\begin{equation}
m_{\omega}(t)=\int_0^\infty e^{t\omega}p(\omega\lvert b,c)d\omega = \mathrm{cosh}^b\left(\frac{c}{2}\right)\mathrm{cosh}^{-b}\left(\sqrt{\frac{c^2/2-t}{2}}\right)
\end{equation}

Mean and variance can be readily obtained through the first two moments

\begin{equation}
m_{\omega}'(t) = \frac{b}{2\sqrt{2}}\mathrm{cosh}^b\left(\frac{c}{2}\right)\mathrm{cosh}^{-b-1}\left(\sqrt{\frac{c^2/2-t}{2}}\right)\mathrm{sinh}\left(\sqrt{\frac{c^2/2-t}{2}}\right)\left(\frac{c^2-2t}{2}\right)^{-1/2}
\end{equation}

\begin{equation}
m_{\omega}'(0)= \frac{b}{2c}\mathrm{cosh}^{-1}\left(\frac{c}{2}\right)\mathrm{sinh}\left(\frac{c}{2}\right)
\end{equation}

\begin{equation}\label{eq:Eomega}
\Bbb{E}[\omega]=m_{\omega}'(0)=\frac{b}{2c}\alpha
\end{equation}

\begin{equation}
\alpha=\mathrm{tanh}\left(\frac{c}{2}\right)= \frac{e^c-1}{e^c+1}
\end{equation}

\begin{subequations}\begin{align}
	m_{\omega}''(t) &= \left[\frac{b}{2\sqrt{2}}\mathrm{cosh}^b\left(\frac{c}{2}\right)\right] \Biggl[(-b-1)\mathrm{cosh}^{-b-2}\left(\sqrt{\frac{c^2/2-t}{2}}\right)\mathrm{sinh}^2\left(\sqrt{\frac{c^2/2-t}{2}}\right) \\ 
	&\times \frac{-1}{2\sqrt{2}}\left(\frac{c^2-2t}{2}\right)^{-1}+\frac{-1}{2\sqrt{2}}\left(\frac{c^2-2t}{2}\right)^{-1}\mathrm{cosh}^{-b}\left(\sqrt{\frac{c^2/2-t}{2}}\right) \\
	&+\frac{1}{2}\left(\frac{c^2-2t}{2}\right)^{-3/2}\mathrm{cosh}^{-b-1}\left(\sqrt{\frac{c^2/2-t}{2}}\right)\mathrm{sinh}\left(\sqrt{\frac{c^2/2-t}{2}}\right)\Biggr]
	\end{align}\end{subequations}

\begin{subequations}\begin{align}
	m_{\omega}''(0) &= \frac{b}{2\sqrt{2}}\Biggl[\frac{b+1}{2\sqrt{2}}\left(\frac{c^2}{2}\right)^{-1}\underbrace{\mathrm{cosh}^{-2}\left(\frac{c}{2}\right)\mathrm{sinh}^{2}\left(\frac{c}{2}\right)}_{\alpha^2}+\frac{-1}{2\sqrt{2}}\left(\frac{c^2}{2}\right)^{-1}    \\
	&+ \frac{1}{2}\left(\frac{c^2}{2}\right)^{-3/2}\underbrace{\mathrm{cosh}^{-1}\left(\frac{c}{2}\right)\mathrm{sinh}\left(\frac{c}{2}\right)}_{\alpha}\Biggr]     
	\end{align}\end{subequations}

\begin{equation}
\Bbb{E}[\omega^2]=m_{\omega}''(0)=\frac{(b^2+b)\alpha^2-b}{4c^2}+\frac{b\alpha}{2c^3}
\end{equation}

\begin{equation}\label{eq:Vomega}
\mathrm{Var}[\omega]=\Bbb{E}[\omega^2]-\Bbb{E}^2[\omega]=\frac{b(\alpha^2-1)}{4c^2}+\frac{b\alpha}{2c^3}
\end{equation}

\begin{figure}[H]
	\centering  
	\begin{tabular}{cc}   
		\includegraphics[width=55mm]{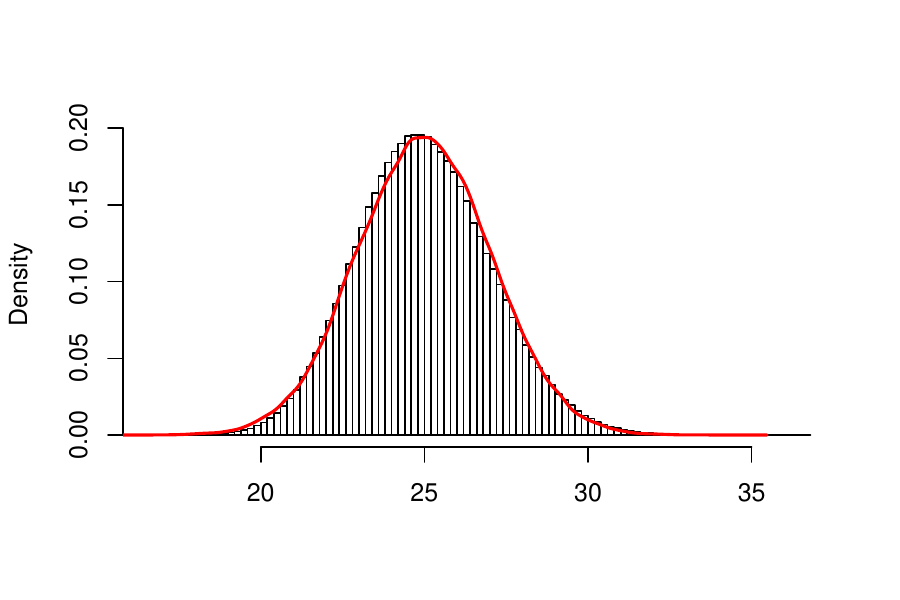} &   \includegraphics[width=55mm]{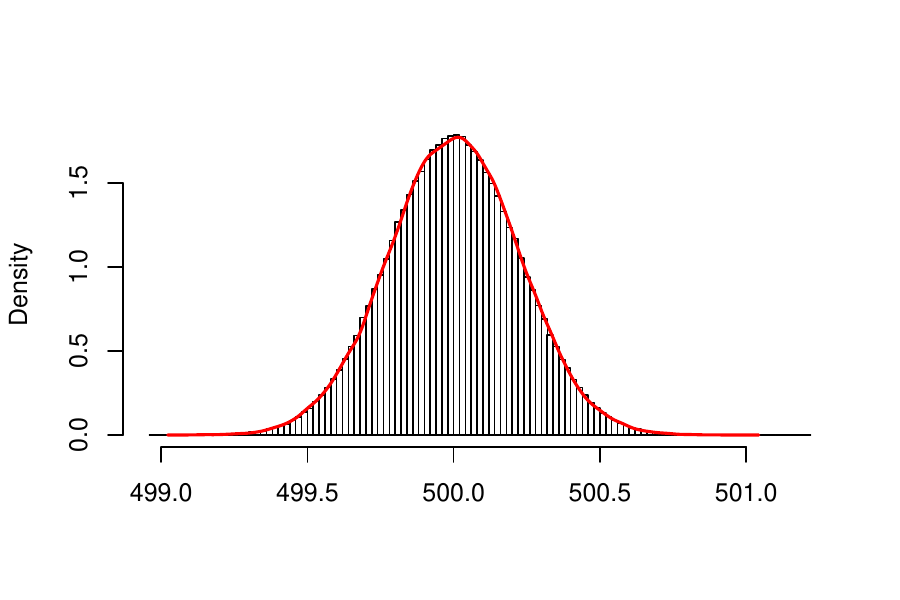} \\ 
		(a) $PG(b=10^2,c=0.001)$ & (b) $PG(b=10^3,c=10)$ \\[6pt]  \includegraphics[width=55mm]{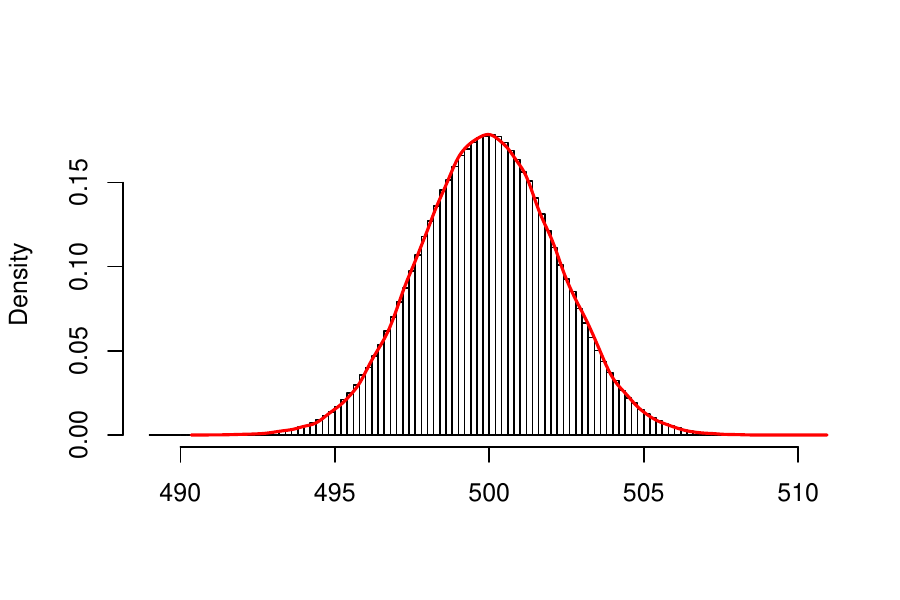} &   \includegraphics[width=55mm]{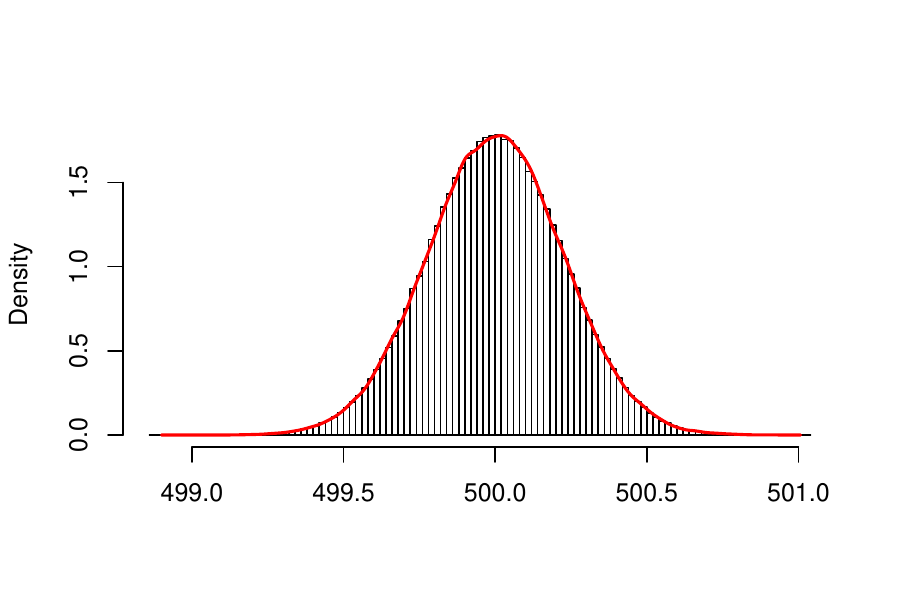} \\ (c) $PG(b=10^4,c=-10)$ & (d) $PG(b=10^5,c=100)$ \\[6pt] 
	\end{tabular} 
	\caption{Four P{\'o}lya-Gamma distributions (histograms), and their Normal approximations (red lines). Sample size is $10^6$.}\label{fig:pgapprox}
\end{figure}

Figure \ref{fig:pgapprox} present several P{\'o}lya-Gamma distributions with large shape
parameter $b$, and their moment-matching Normal approximation using
(\ref{eq:Eomega}, \ref{eq:Vomega}). For very large $b$ and/or $c$ computations should be
performed in logarithmic scale to avoid numerical overflow.

\section{Gibbs Sampler}

\subsection{P{\'o}lya-Gamma data augmentation}
\label{sec:pgtrick}
The proposed Gibbs sampler involves the P{\'o}lya-Gamma data augmentation
trick in \citet{polson2013bayesian}. A random variable $\omega \in \Bbb{R^+}$
is said to have a P{\'o}lya-Gamma distribution, $\omega \sim PG(b,c)$
with parameters $b\in\Bbb{R^{+}}$ and $c \in \Bbb{R}$, if
\begin{equation}
\omega\stackrel{D}{=}\frac{1}{2\pi^{2}}\sum_{k=1}^{\infty}\frac{g_{k}}{(k-1/2)^{2}+c^{2}/(4\pi^{2})},\quad g_{k}\stackrel{\mathrm{iid}}{\sim}\mathrm{Gamma}(b,1),
\end{equation}
with density
\begin{equation}
p(\omega\lvert b,c)=\mathrm{cosh}^{b}(c/2)\frac{2^{b-1}}{\Gamma(b)}\sum_{n=0}^{\infty}{(-1)^{n}}\frac{\Gamma(n+b)}{\Gamma(n+1)}\frac{2n+b}{\sqrt{2\pi\omega^{3}}}e^{-\frac{(2n+b)^{2}}{8\omega}-\frac{c^{2}\omega}{2}}.
\end{equation}

The relation between a Binomial likelihood and the P{\'o}lya-Gamma distribution
is given the following two key results in \citet{polson2013bayesian}
\begin{equation}\label{eq:polsonkey1}
\frac{{(e^{\psi})}^{a}}{{(1+e^{\psi})}^{b}}=2^{-b}e^{\kappa\psi}\int_{0}^{\infty}{e^{-\omega\psi^{2}/2}p(\omega\lvert b,0)d\omega},
\end{equation}
\begin{equation}\label{eq:polsonkey2}
p(\omega\lvert\psi)=\frac{e^{-\omega\psi^{2}/2}p(\omega\lvert b,0)}{\int_{0}^{\infty}{e^{-\omega\psi^{2}/2}p(\omega\lvert b,0)d\omega}}\sim PG(b,\psi).
\end{equation}
To see the connection to the binomial model, note that the contribution
to the likelihood of each observation in the model $y_{i}\sim \mathrm{Binomial}(n_{i},\pi_{i})$
with $\ensuremath{\psi_{i}=\mathrm{logit}(\pi_{i})=\boldsymbol{x}_{i}^{T}\boldsymbol{\beta}}$
is
\begin{equation}
p(y_{i}\lvert\boldsymbol{\beta},x_{i})=\frac{\exp(\boldsymbol{x}_{i}^{T}\boldsymbol{\beta})^{y_{i}}}{\left[1+\exp(\boldsymbol{x}_{i}^{T}\boldsymbol{\beta})\right]^{n_{i}}}.
\end{equation}
Using Eq. (\ref{eq:polsonkey1}) this can be expressed as
\begin{equation}
p(y_{i}\lvert\boldsymbol{\beta},x_{i})\propto\exp(\kappa_{i}\boldsymbol{x}_{i}^{T}\boldsymbol{\beta})\int_{0}^{\infty}{\exp(-\omega_{i}({\boldsymbol{x}_{i}^{T}\boldsymbol{\beta}})^{2}/2)p(\omega_{i}\lvert n_{i},0)}
\end{equation}
where $\ensuremath{\kappa_{i}=y_{i}-n_{i}/2}$ and $\omega_{i}\sim PG(n_{i},0)$.
By conditioning on the P{\'o}lya-Gamma variables $\boldsymbol{\omega}={\omega_{1},\ldots,\omega_{N}}$
using Eq. (\ref{eq:polsonkey2}), direct application of Bayes' theorem yield the posterior
for $\boldsymbol{\beta}$,
\begin{align}
p(\boldsymbol{\beta}\lvert\boldsymbol{\omega},\boldsymbol{y},X) & \propto p(\boldsymbol{\beta})\prod_{i=1}^{N}p(y_{i}\lvert\boldsymbol{\beta},\omega_{i},x_{i})=p(\boldsymbol{\beta})\prod_{i=1}^{N}\exp(\kappa_{i}\boldsymbol{x}_{i}^{T}\boldsymbol{\beta}-\omega_{i}(\boldsymbol{x}_{i}^{T}\boldsymbol{\beta})^{2}/2)\\
& \propto p(\boldsymbol{\beta})\exp(-\frac{1}{2}(\boldsymbol{z}-X\boldsymbol{\beta})^{T}\Omega(\boldsymbol{z}-X\boldsymbol{\beta})),
\end{align}
which is the posterior for a Gaussian linear regression with response
$\boldsymbol{z}$, mean $X\boldsymbol{\beta}$ and known covariance
matrix $\Omega^{-1}$, where $\boldsymbol{z}=(\kappa_{1}/\omega_{1},\ldots,\kappa_{N}/\omega_{N})$
and $\Omega=\mathrm{diag}(\boldsymbol{\omega})$. Using the prior
$\boldsymbol{\beta}\sim\mathcal{N}(\boldsymbol{\mu}_{0},\Sigma_{0})$,
the P{\'o}lya-Gamma method for the Binomial model therefore results in
a two-step Gibbs sampler that alternates between \begin{subequations}\begin{align} \omega_i\lvert\boldsymbol{\beta},X & \sim PG(n_i,\boldsymbol{x}_i^T\boldsymbol{\beta}) \\
	\boldsymbol{\beta}\lvert X,\boldsymbol{y},\boldsymbol{\omega} & \sim \mathcal{N}(\boldsymbol{\mu}_\omega,\Sigma_\omega)
	\end{align}\end{subequations}where \begin{subequations}\begin{align} \boldsymbol{\mu}_\omega & = \Sigma_\omega(X^T\kappa+\Sigma_0^{-1}\boldsymbol{\mu_0}) \\
	\Sigma_\omega & = (X^T\Omega X+\Sigma_0^{-1})^{-1}.
	\end{align}\end{subequations}

\subsection{Updating steps}
\label{sec:gibbssampler}
\subsection*{1. Update the block probabilities}

\begin{algorithmic} 
	\State Compute the clustering quantities given the current assignments $z$
	\State $n_p=\sum_{i=1}^N\mathbb{I}(z_i=p)$, for all $p = 1,\ldots,B$
	\State $n_{pq}^k(t)=n_pn_q-n_p\mathbb{I}(p=q)$, for all $\{p,q\}\in\{1,\ldots,B\}$, $t=t_1,\ldots,t_T$, $k=1,\ldots,K$  
	\State $y_{pq}^k(t)=\underset{\{i,j\}:z_i=p,z_j=q}{\sum\sum A_{ij}^k(t)}$, for all $\{p,q\}\in\{1,\ldots,B\}$, $t=t_1,\ldots,t_T$, $k=1,\ldots,K$  
	\State Sample the vector of block probabilities $\eta_1,\ldots,\eta_B$ 
	\State $\eta\lvert - \sim \mathrm{Dirichlet}(\alpha_1+n_1,\ldots,\alpha_B+n_B)$
\end{algorithmic}

\subsection*{2. Generate the P{\'o}lya-Gamma variables}

\begin{algorithmic} 
	\State Sample the augmented data $\omega_{pq}^k(t)$ 
	\For {each time $t=t_1,\ldots,t_T$, layer $k=1,\ldots,K$ and block pair $\{p,q\}\in\{1,\ldots,B\}$}
	\State \textbf{if} {$p=q$} \textbf{then} {$c=\mu_p^k(t)+\sum_{r=1}^R\bar{x}_{pr}(t)$} \textbf{else} {$c=\mu(t)+\bar{x}_p^{\intercal}(t)\bar{x}_q(t)+x^{k \intercal }_p(t)x^k_q(t)$}
	\State Let $b=n_{pq}^k(t)$ and $\alpha = \mathrm{tanh}(0.5c)$
	\State \textbf{if} {$b<100$} \textbf{then} {$\omega_{pq}^k(t)\lvert - \sim PG(b,c)$} \textbf{else} {$\omega_{pq}^k(t)\lvert - \sim \mathcal{N}\left(\frac{b}{2c}\alpha,\frac{b(\alpha^2-1)}{4c^2}+\frac{b\alpha}{2c^3}\right)$}
	\EndFor
\end{algorithmic}

\subsection*{3. Update the between-block dynamic mean}

\begin{algorithmic} 
	\State Sample the vector $\mu(t)=[\mu(t_1),\ldots,\mu(t_T)]^\intercal$ from 
	\State $\mu\lvert - \sim \mathcal{N}_T(\mu_\mu,\Sigma_\mu)$
	\State $\Sigma_\mu=\left[\mathrm{diag}\left\{\sum_{k=1}^{K}\sum_{p=2}^{B}\sum_{q=1}^{p-1}{\omega_{pq}^k(t_1)},\ldots,\sum_{k=1}^{K}\sum_{p=2}^{B}\sum_{q=1}^{p-1}{\omega_{pq}^k(t_T)}\right\} +K^{-1}_\mu\right]$
	\State $\mu_\mu=\Sigma_{\mu}^{-1}\begin{bmatrix}   
	\sum_{k=1}^{K}\sum_{p=2}^{B}\sum_{q=1}^{p-1}\{y_{pq}^k(t_1)-n_{pq}^k(t_1)/2-\omega_{pq}^k(t_1)[\bar{x}_p^\intercal(t_1)\bar{x}_q(t_1)+x^{k\intercal}_p(t_1)x^k_q(t_1)]\} \\   
	\vdots  \\
	\sum_{k=1}^{K}\sum_{p=2}^{B}\sum_{q=1}^{p-1}\{y_{pq}^k(t_T)-n_{pq}^k(t_T)/2-\omega_{pq}^k(t_T)[\bar{x}_p^\intercal(t_T)\bar{x}_q(t_T)+x^{k\intercal}_p(t_T)x^k_q(t_T)]\}
	\end{bmatrix}$
\end{algorithmic}

\subsection*{4. Update the cross-layer block coordinates}

\begin{algorithmic} 
	\State Sample the vectors $\bar{x}_p(t_1),\ldots,\bar{x}_p(t_T)$ for every block and layer 
	\For {each block $p = 1,\ldots,B$}
	\State Block-sample $\{\bar{x}_p(t_1),\ldots,\bar{x}_p(t_T)\}$ conditioned on $\{\bar{x}_q(t):p\neq q,t=t_1,\ldots,t_T\}$
	\State Let $\bar{x}_p=\{\bar{x}_{p1}(t_1),\ldots,\bar{x}_{p1}(t_T),\ldots,\bar{x}_{pR}(t_1),\ldots,\bar{x}_{pR}(t_T)\}$ 
	\State Let $x_p^k=\{x_{p1}^k(t_1),\ldots,x_{p1}^k(t_T),\ldots,x_{pH}^k(t_1),\ldots,x_{pH}^k(t_T)\}$ 
	\For {each layer $k=1,\ldots,K$ and $q<p$}
	\State Define a Bayesian logistic regression with $\bar{x}_p$ as coefficient vector like
	\State $y_p^k \sim \mathrm{Binomial}(n_p^k,\pi_p^k),\quad \text{logit}(\pi_p^k)=\tilde{\mu}+\bar{X}_{-p}\bar{x}_p+X_{-p}^kx_p^k$
	\State $y_p^k = [\bigcup\limits_{p\neq q}\{y^k_{pq}(t_1),\ldots,y^k_{pq}(t_T)\},\{y^k_{pp}(t_1),\ldots,y^k_{pp}(t_T)\}]^\intercal$
	\State $n_p^k = [\bigcup\limits_{p\neq q}\{n^k_{pq}(t_1),\ldots,n^k_{pq}(t_T)\},\{n^k_{pp}(t_1),\ldots,n^k_{pp}(t_T)\}]^\intercal$
	\State $\pi_p^k = [\bigcup\limits_{p\neq q}\{\pi^k_{pq}(t_1),\ldots,\pi^k_{pq}(t_T)\},\{\pi^k_{pp}(t_1),\ldots,\pi^k_{pp}(t_T)\}]^\intercal$
	\State $\tilde{\mu}=[\mathbf{1}_{B-1}\otimes \mu,\mu_p^k]^\intercal$
	\State The prior is $\bar{x}_p \sim \mathcal{N}_{T\times R}(0,\mathrm{diag}(\tau_1^{-1},\ldots,\tau_R^{-1})\otimes K_{\bar{x}})$. Design matrices {$\bar{X}_{-p}$ and\State $\bar{X}_{-p}^k$ 
		 contain regressors chosen respectively} from $\bar{x}_p$ and $x_p^k$ to fulfill Eq. (\ref{eq:OurModel}). 
	\State $\Omega_p^k$ is a diagonal matrix with the  corresponding P{\'o}lya-Gamma variables.
	\EndFor
	\State Using the above specification the update of $\bar{x}_p$ becomes
	\State $\bar{x}_p\lvert - \sim \mathcal{N}_{T\times R}(\mu_{\bar{x}_p},\Sigma_{\bar{x}_p})$
	\State $\Sigma_{\bar{x}_p}=\left[\bar{X}_{-p}^\intercal(\sum_{k=1}^{K}\Omega_p^k)\bar{X}_{-p} +\mathrm{diag}(\tau_1,\ldots,\tau_R)\otimes K^{-1}_{\bar{x}}\right]$
	\State $\mu_{\bar{x}_p}=\Sigma_{\bar{x}_p}^{-1}\left[\bar{X}_{-p}^\intercal(\sum_{k=1}^{K}\{y_p^k-n_p^k(t)/2-\Omega_p^k[\tilde{\mu}+X_{-p}^kx_p^k]\})\right]$
	\EndFor
\end{algorithmic}

\subsection*{5. Update the within-layer coordinates}

\begin{algorithmic} 
	\State Sample the vectors $x_p^k(t_1),\ldots,x_p^k(t_T)$ for every block and layer 
	\For {each layer $k = 1,\ldots,K$}
	\For {each block $p = 1,\ldots,B$}
	\State Block-sample $\{x^k_p(t_1),\ldots,x^k_p(t_T)\}$ conditioned on $\{x^k_q(t):p\neq q,t=t_1,\ldots,t_T\}$
	\State Adapting Step 4 to fulfill Eq. (\ref{eq:OurModel}) yields the following update for $x^k_p$
	\State $x^k_p\lvert - \sim \mathcal{N}_{T\times H}(\mu_{x^k_p},\Sigma_{x^k_p})$
	\State $\Sigma_{x^k_p}=\left[X_{-p}^{k\intercal}\Omega_p^kX_{-p}^k +\mathrm{diag}(\tau_1^k,\ldots,\tau_H^k)\otimes K^{-1}_{x}\right]$
	\State $\mu_{x^k_p}=\Sigma_{x^k_p}^{-1}\left[X_{-p}^{k\intercal}(y_p^k-n_p^k(t)/2-\Omega_p^k[\mathbf{1}_{B-1}\otimes\mu+\bar{X}_{-p}\bar{x}_p])\right]$
	\EndFor
	\EndFor
\end{algorithmic}

\subsection*{6. Update the cross-layer shrinkage parameters}

\begin{algorithmic}
	\State Sample the gamma quantities that define the shrinkage parameters $\tau_1,\ldots,\tau_R$
	\State $\delta_1\lvert -\sim \mathrm{Gamma}\left(a_1+\frac{B\times T\times R}{2},1+0.5\sum\limits_{m=1}^{R}\theta_m^{(-1)}\sum\limits_{p=1}^{B}\bar{x}_{pm}^\intercal K_{\bar{x}}^{-1}\bar{x}_{pm}\right)$
	\State $\delta_{r\ge 2}\lvert -\sim \mathrm{Gamma}\left(a_2+\frac{B\times T\times (R-r+1)}{2},1+0.5\sum\limits_{m=r}^{R}\theta_m^{(-r)}\sum\limits_{p=1}^{B}\bar{x}_{pm}^\intercal K_{\bar{x}}^{-1}\bar{x}_{pm}\right)$
	\State where $\theta_m^{(-r)}=\prod\limits_{f=1,f\neq r}^m\delta_f$ for $r=1,\ldots,R$ and $\bar{x}_{pm}=\{\bar{x}_{pm}(t_1),\ldots,\bar{x}_{pm}(t_T)\}^\intercal$
	
\end{algorithmic}

\subsection*{7. Update the within-layer shrinkage parameters}

\begin{algorithmic}
	\For {each layer $k=1,\ldots,K$}
	\State Sample the gamma quantities that define the within-layer $\tau^k_1,\ldots,\tau^k_H$
	\State $\delta_1^k\lvert -\sim \mathrm{Gamma}\left(a_1+\frac{B\times T\times H}{2},1+0.5\sum\limits_{l=1}^{H}\theta_l^{(-1)}\sum\limits_{p=1}^{B}{x}_{pl}^{k\intercal}K_{x}^{-1}x^k_{pl}\right)$
	\State $\delta_{h\ge 2}^k\lvert -\sim \mathrm{Gamma}\left(a_2+\frac{B\times T\times (H-h+1)}{2},1+0.5\sum\limits_{l=1}^{H}\theta_l^{(-h)}\sum\limits_{p=1}^{B}{x}_{pl}^{k\intercal}K_{x}^{-1}x^k_{pl}\right)$
	\State where $\theta_l^{(-h)}=\prod\limits_{f=1,f\neq h}^l\delta_f^k$ for $h=1,\ldots,H$ and $x^k_{pl}=\{x^k_{pl}(t_1),\ldots,x^k_{pl}(t_T)\}^\intercal$
	\EndFor
\end{algorithmic}

\subsection*{8. Update the within-block dynamic mean}

\begin{algorithmic} 
	\State Sample the vector $\mu_p^k(t)=[\mu_p^k(t_1),\ldots,\mu_p^k(t_T)]^\intercal$ for every block and layer 
	\For {each block $p=1,\ldots,B$ and layer $k=1,\ldots,K$}
	\State $\mu_p^k\lvert - \sim \mathcal{N}_T(\mu_{\mu_p},\Sigma_{\mu_p})$
	\State $\Sigma_{\mu_p}=\left[\mathrm{diag}\left\{\omega_{pp}^k(t_1),\ldots,\omega_{pp}^k(t_T)\right\}+K^{-1}_{\mu_p}\right]$
	\State $\mu_{\mu_p}=\Sigma_{\mu_{\mu_p}}^{-1}\begin{bmatrix}   
	y_{pp}^k(t_1)-n_{pp}^k(t_1)/2-\omega_{pp}^k(t_1)\sum_{r=1}^R\bar{x}_{pr}(t) \\   
	\vdots  \\
	y_{pp}^k(t_T)-n_{pp}^k(t_T)/2-\omega_{pp}^k(t_T)\sum_{r=1}^R\bar{x}_{pr}(t)
	\end{bmatrix}$
	\EndFor
\end{algorithmic}

\subsection*{9. Compute posterior block probabilities}

\begin{algorithmic}
	\State Obtain the posterior block probabilities $\pi_{pq}^k(t)$
	\For {each time $t=t_1,\ldots,t_T$, layer $k=1,\ldots,K$ and block pair $\{p,q\}\in\{1,\ldots,B\}$}
	\If {$p \neq q$} 
	\State $\pi_{pq}^k(t)=\left[1+\exp\{-\mu(t)-\bar{x}_{p}^\intercal (t)\bar{x}_{q}(t)-x^{k\intercal}_p(t)x^k_q(t)\}\right]^{-1}$
	\Else 
	\State $\pi_{pp}^k(t)=\left[1+\exp\{-\mu_p^k(t)-\sum_{r=1}^R\bar{x}_{pr}(t)\}\right]^{-1}$
	\EndIf
	\EndFor
\end{algorithmic}

\subsection*{10. Update the block assignments}

\begin{algorithmic}
	\State Sample the latent block assignments $z$ sequentially.
	\State Denote $z^*_i$ if the assignment of vertex $i$ has been already updated, and $z_i$ otherwise.
	\For {each vertex $i=1,\ldots,N$}
	\State Let $\tilde{z}=[z^*_1,\ldots,z^*_{i-1},z_{i+1},\ldots,z_N]$
	\For {each block $p=1,\ldots,B$}
	\State $\gamma_{ip}=p(z_i=p\lvert -)\propto\eta_p\prod\limits_{t=t_1}^{t_T}\prod\limits_{k=1}^{K}\prod\limits_{q=1}^{B}[\pi_{pq}^k(t)]^{\sum\limits_{j\neq i:\tilde{z}_j=q} A_{ij}^k(t)}[1-\pi_{pq}^k(t)]^{\sum\limits_{j\neq i:\tilde{z}_j=q}1-A_{ij}^k(t)}$
	\EndFor
	\State $z_i\lvert -\sim \mathrm{Categorical}(\gamma_i)$
	\EndFor
	\State \textbf{Note}: The pseudocode above describes a complete update of all $z_i$'s though we strongly recommend random-scan Gibbs sampling to alleviate the computational burden. In a random-scan the outer loop will iterate only over the $z_i$'s randomly selected at a given MCMC step, and the vector $\tilde{z}$ will not have a sequential structure but need to be defined appropriately. All other calculations remain the same. 
\end{algorithmic}

\subsection*{11. {[}Optional{]} Edge prediction/imputation}

\begin{algorithmic} 
	\State Sample the unobserved edges from $\pi_{pq}^k(t^*)$.
	\State Let $t^*\subset t$ be the unobserved time intervals, and denote the unobserved
	\State part of the data as $A_u=A_{ij}^k(t^*)$.
	\For {each time $t^* \in \{t^*_1,\ldots,t^*_{T^*}\}$, layer $k=1,\ldots,K$ and vertex pair $\{i,j\}\in\{1,\ldots,N\}$}
	\State Impute the unobserved edges from $A_u\lvert - \sim \mathrm{Bernoulli}(\pi_{z_iz_j}^k(t^*))$
	\EndFor
\end{algorithmic}

\section{Notation}
\label{sec:notation}
\begin{tabular}{|c|l|}
    \hline
    \textbf{Symbol} & \textbf{Meaning}\tabularnewline
	\hline 
	$N\in\Bbb{N}$ & Number of network nodes/vertices\tabularnewline
	\hline 
	$\{i,j\}=1,\ldots,N$ & Indices for the network nodes\tabularnewline
	\hline 
	$T\in\Bbb{N}$ & Number of time intervals\tabularnewline
	\hline 
	$t=t_1,\ldots,t_T$ & Index for the time intervals\tabularnewline
	\hline 
	$K\in\Bbb{N}$ & Number of network layers\tabularnewline
	\hline 
	$k=1,\ldots,K$ & Index for the network layers\tabularnewline
	\hline 
	$B\in\Bbb{N}$ & Number of network blocks/clusters\tabularnewline
	\hline 
	$\{p,q\}=1,\ldots,B$ & Indices for the network blocks\tabularnewline
	\hline 
	$A_{ij}^k(t)\in\{0,1\}$ & Adjacency matrix at time $t$ and layer $k$ \tabularnewline
	\hline 
	$\theta_{ij}^k(t)\in[0,1]$ & Matrix of edge probabilities at time $t$ and layer $k$\tabularnewline
	\hline 
	$\pi_{pq}^k(t)\in[0,1]$ & Matrix of block probabilities at time $t$ and layer $k$\tabularnewline
	\hline 
	$\psi_{pq}^k(t)\in\Bbb{R^+}$ & Logit of the block probabilities at time $t$ and layer $k$\tabularnewline
	\hline 
	$\bar{x}_{pr}(t)\in\Bbb{R}$ & Latent between-layer coordinate $r$, for block $p$ at time $t$\tabularnewline
	\hline 
	$x^k_{ph}(t)\in\Bbb{R}$ & Latent within-layer coord. $h$, for block $p$, at time $t$
	and layer $k$\tabularnewline
	\hline 
	$\mu(t)\in\Bbb{R}$ & Latent between-block intercept at time $t$\tabularnewline
	\hline 
	$\mu_p^k(t)\in\Bbb{R}$ & Latent within-block intercept for block $p$, at time $t$ and layer
	$k$ \tabularnewline
	\hline 
	$\tau_r\in\Bbb{R^+}$ & Shrinkage parameter for the latent between-layer coordinate $r$\tabularnewline
	\hline 
	$\tau_h^k\in\Bbb{R^+}$ & Shrinkage for the latent within-layer coordinate $h$ at
	layer $k$\tabularnewline
	\hline 
	$\delta_r\in\Bbb{R^+}$ & Gamma for shrinkage parameter $\tau_r$\tabularnewline
	\hline 
	$\delta_h^k\in\Bbb{R^+}$ & Gamma for shrinkage parameter $\tau_h^k$\tabularnewline
	\hline 
	$\{a_1,a_2\}\in\Bbb{R^+}$ & Shape hyperparameters for the shrinkage Gammas\tabularnewline
	\hline 
	$z_i\in\{0,\ldots,B\}$ & Block/cluster assignment vector\tabularnewline
	\hline 
	$\eta_p\in[0,1]$ & Prior probability that a node belongs to block $p$\tabularnewline
	\hline 
	$\boldsymbol{\alpha}=\{\alpha_1,\ldots,\alpha_B\}\in[0,1]$ & Concentration hyperparameter vector for $\eta$\tabularnewline
	\hline 
	$\gamma_{ip}=p(z_i=p)\in[0,1]$ & Posterior probablity that node $i$ belongs to block $p$\tabularnewline
	\hline 
	$n_p \in\Bbb{N}$ & Number of vertices in block $p$, such that $\sum_{p=1}^Bn_p=N$\tabularnewline
	\hline 
	$n_{pq}^k(t)\in\Bbb{N}$ & Matrix of potential edges b/w blocks $\{p,q\}$, at time $t$, layer $k$\tabularnewline
	\hline 
	$y_{pq}^k(t)\in\Bbb{N}$ & Matrix of actual edges b/w blocks $\{p,q\}$, at time $t$, layer $k$\tabularnewline
	\hline 
	$\omega_{pq}^k(t)\in\Bbb{R^+}$ & P{\'o}lya-Gamma variable for block pair $\{p,q\}$, at time $t$, layer $k$\tabularnewline
	\hline 
	$\Omega=\mathrm{diag}(\omega_{pq}^k(t))\in\Bbb{R^+}$ & Matrix with diagonal P{\'o}lya-Gamma variables\tabularnewline
	\hline 
	$k_f(t,t')\in\Bbb{R^+}$ & Kernel function for latent variable $f$\tabularnewline
	\hline 
	$l\in\Bbb{R^+}$ & Lengthscale for a Radial-Basis-Function kernel\tabularnewline
	\hline 
	$K_f\in\Bbb{R^+}$ & Gramian matrix from the kernel $k_f(t,t')$\tabularnewline
	\hline 
	$T^*\in\Bbb{N}$ & Number of unobserved time intervals, $T^*<T$\tabularnewline
	\hline 
	$t^*=t^*_1,\ldots,t^*_{T^*}$ & Index for the unobserved time intervals, $t^*\subset t$\tabularnewline
	\hline 
	$A_u=A_{ij}^k(t^*)$ & Unobserved adjacency matrices\tabularnewline
	\hline 
	$D^k(t)$ & Network density at time $t$ and layer $k$ \tabularnewline
	\hline 
	$d_i^k(t)$ & Degree of the vertex $i$ at time $t$ and layer $k$ \tabularnewline
	\hline
	$S_{ij}^k(t)$ & Matrix of smoothed similarity scores at time $t$ and layer $k$ \tabularnewline
	\hline

\end{tabular}
\newpage

\section{IATA Airport Codes}
\label{sec:iatacodes}
\small
\begin{tabular}{|c|c|c|c|}
	\hline 
	\textbf{IATA} & \textbf{Airport} & \textbf{IATA} & \textbf{Airport}\tabularnewline
	\hline 
	\hline 
	ABQ & Albuquerque Intl.  & MCO & Orlando Intl.\tabularnewline
	\hline 
	ALB & Albany Intl. & MDW & Chicago Midway Intl.\tabularnewline
	\hline 
	ATL & Hartsfield-Jackson Atlanta Intl. & MEM & Memphis Intl.\tabularnewline
	\hline 
	AUS & Austin Bergstrom Intl. & MHT & Manchester-Boston Regional\tabularnewline
	\hline 
	BDL & Hartford Bradley Intl. & MIA & Miami Intl.\tabularnewline
	\hline 
	BHM & Birmingham-Shuttlesworth Intl. & MKE & Milwaukee General Mitchell Intl.\tabularnewline
	\hline 
	BNA & Nashville Intl. & MSP & Minneapolis-St Paul Intl.\tabularnewline
	\hline 
	BOI & Boise Air Terminal & MSY & Louis Armstrong New Orleans Intl.\tabularnewline
	\hline 
	BOS & Boston Logan Intl. & OAK & Metropolitan Oakland Intl.\tabularnewline
	\hline 
	BUF & Buffalo Niagara Intl. & OGG & Kahului Airport\tabularnewline
	\hline 
	BUR & Burbank Bob Hope & OKC & Oklahoma City \tabularnewline
	\hline 
	BWI & Baltimore/Washington Intl. & OMA & Omaha Eppley Airfield\tabularnewline
	\hline 
	CHS & Charleston AFB/Intl. & ONT & Ontario Intl.\tabularnewline
	\hline 
	CLE & Cleveland-Hopkins Intl. & ORD & Chicago O'Hare Intl.\tabularnewline
	\hline 
	CLT & Charlotte Douglas Intl. & ORF & Norfolk Intl.\tabularnewline
	\hline 
	CMH & John Glenn Columbus Intl. & PBI & Palm Beach Intl.\tabularnewline
	\hline 
	CVG & Cincinnati/Northern Kentucky Intl. & PDX & Portland Intl.\tabularnewline
	\hline 
	DAL & Dallas Love Field & PHL & Philadelphia Intl.\tabularnewline
	\hline 
	DCA & Ronald Reagan Washington National & PHX & Phoenix Sky Harbor Intl.\tabularnewline
	\hline 
	DEN & Denver Intl. & PIT & Pittsburgh Intl.\tabularnewline
	\hline 
	DFW & Dallas/Fort Worth Intl. & PNS & Pensacola Intl.\tabularnewline
	\hline 
	DTW & Detroit Metro Wayne County & PVD & Providence Theodore Francis Green\tabularnewline
	\hline 
	ELP & El Paso Intl. & RDU & Raleigh-Durham Intl.\tabularnewline
	\hline 
	EWR & Newark Liberty Intl. & RIC & Richmond, VA: Richmond Intl.\tabularnewline
	\hline 
	FLL & Fort Lauderdale-Hollywood Intl. & RNO & Reno/Tahoe Intl.\tabularnewline
	\hline 
	GEG & Spokane Intl. & RSW & Fort Myers Southwest Florida Intl.\tabularnewline
	\hline 
	GRR & Grand Rapids Gerald R. Ford Intl. & SAN & San Diego Intl.\tabularnewline
	\hline 
	HNL & Honolulu Intl. & SAT & San Antonio Intl.\tabularnewline
	\hline 
	HOU & Houston William P Hobby & SAV & Savannah/Hilton Head Intl.\tabularnewline
	\hline 
	IAD & Washington Dulles Intl. & SDF & Louisville Intl.-Standiford Field\tabularnewline
	\hline 
	IAH & Houston George Bush Intl. & SEA & Seattle/Tacoma Intl.\tabularnewline
	\hline 
	IND & Indianapolis Intl. & SFO & San Francisco Intl.\tabularnewline
	\hline 
	ISP & Long Island MacArthur & SJC & Norman Y. Mineta San Jose Intl.\tabularnewline
	\hline 
	JAX & Jacksonville Intl. & SLC & Salt Lake City Intl.\tabularnewline
	\hline 
	JFK & New York John F. Kennedy Intl. & SMF & Sacramento Intl.\tabularnewline
	\hline 
	LAS & Las Vegas McCarran Intl. & SNA & John Wayne Orange County\tabularnewline
	\hline 
	LAX & Los Angeles Intl. & STL & St Louis Lambert Intl.\tabularnewline
	\hline 
	LGA & New York La Guardia & TPA & Tampa Intl.\tabularnewline
	\hline 
	LIT & Little Rock Clinton Nat. Adams Field & TUL & Tulsa Intl.\tabularnewline
	\hline 
	MCI & Kansas City Intl. & TUS & Tucson Intl.\tabularnewline
	\hline 
\end{tabular}

\end{appendix}

\end{document}